\documentclass[times, review, 10pt]{elsarticle}
\usepackage{amsmath,amssymb,amsfonts}
\usepackage{cases}
\usepackage{color,xcolor}
\usepackage{graphicx}
\usepackage[normalem]{ulem}
\usepackage{subfigure}
\usepackage{subcaption}
\usepackage{algorithm}
\usepackage{algorithmic}
\usepackage{epstopdf}
\usepackage{multirow}
\usepackage{makecell}
\usepackage{rotating}
\usepackage{booktabs}
\usepackage{subfigure}
\usepackage{tabularray}
\usepackage{setspace}
 \usepackage{multirow}
 \useunder{\uline}{\ul}{}
\usepackage[colorlinks,
linkcolor=blue,
anchorcolor=blue,
citecolor=blue
]{hyperref}
\usepackage{amssymb}
\usepackage{amsmath}

\journal{Pattern Recognition}

\begin{document}

\begin{frontmatter}



\title{STAR-Net: An Interpretable Model-Aided Network for Remote Sensing Image Denoising} 


\author[1,2]{Jingjing Liu} 
\ead{jjliu@shu.edu.cn} 
\author[1]{Jiashun Jin}
\ead{jsjin@shu.edu.cn} 
\author[3]{Xianchao Xiu\corref{cor1}}
\ead{xcxiu@shu.edu.cn} 
\author[1]{Jianhua Zhang}
\ead{jhzhang@shu.edu.cn} 

\author[4]{Wanquan Liu}
\ead{liuwq63@mail.sysu.edu.cn} 
\affiliation[1]{organization={School of Microelectronics, Shanghai Key Laboratory of Chips and Systems for Intelligent Connected Vehicle},
            addressline={Shanghai University}, 
            city={Shanghai 200444}, 
            country={China}}
\affiliation[2]{organization={State Key Laboratory of Integrated Chips and Systems},
            addressline={Fudan University}, 
            city={Shanghai 201203}, 
            country={China}}             
\affiliation[3]{organization={School of Mechatronic Engineering and Automation},
            addressline={Shanghai University}, 
            city={Shanghai 200444}, 
            country={China}}    
\affiliation[4]{organization={School of Intelligent Systems Engineering},
            addressline={Sun Yat-Sen University}, 
            city={Shenzhen 510275}, 
            country={China}}  
\cortext[cor1]{Corresponding author}       
\begin{abstract}
Remote sensing image (RSI) denoising is an important topic in the field of remote sensing.
Despite the impressive denoising performance of RSI denoising methods, most current deep learning-based approaches function as black boxes and lack integration with physical information models, leading to limited interpretability.
Additionally, many methods may struggle with insufficient attention to non-local self-similarity in RSI and require tedious tuning of regularization
parameters to achieve optimal performance, particularly in conventional iterative optimization approaches.
In this paper,  we first propose a novel RSI denoising method named sparse tensor-aided representation network (STAR-Net), which leverages a low-rank prior to effectively capture the non-local self-similarity within RSI. 
Furthermore, we extend STAR-Net to a sparse variant called STAR-Net-S to deal with the interference caused by non-Gaussian noise in original RSI for the purpose of improving robustness. 
Different from conventional iterative optimization, we develop an alternating direction method of multipliers (ADMM)-guided deep unrolling network, in which all regularization parameters can be automatically learned, thus inheriting the advantages of both model-based and deep learning-based approaches and successfully addressing the above-mentioned shortcomings. 
Comprehensive experiments on synthetic and real-world datasets demonstrate that STAR-Net and STAR-Net-S outperform state-of-the-art RSI denoising methods. 
The code will be available at \href{https://github.com/Jason011212/STAR-Net}{https://github.com/Jason011212/STAR-Net}.
\end{abstract}


\end{frontmatter}



\section{Introduction}
\label{sec1}
With the rapid development of remote sensing technology, remote sensing image (RSI) has found extensive applications in anomaly detection \cite{WANG2023109795}, denoising \cite{10144690}, classification \cite{WANG2025110878}, and unmixing \cite{chen2023integration}. By capturing detailed spectral information, RSI enables precise analysis of ground features.
Nevertheless, during the acquisition process, RSI is inevitably affected by noise, which significantly impacts subsequent data analysis. 
Therefore, recovering clean RSI from noisy ones has become a critical challenge. Gernerally speaking, the existing RSI denoising methods can be categorized into two primary groups: model-based and deep learning-based.

 Model-based methods, whose core is to create a connection between clean and noisy RSI through physical priors related to natural statistics or image formation for denoising \cite{9920675}. A widely used method is block matching 3D (BM3D) \cite{4271520}, which exploits the similarity between non-local blocks to achieve the purpose of denoising. To consider the information across bands, Maggioni \textit{et al.} \cite{6253256} proposed a 4D collaborative filtering paradigm for volumetric data denoising called block matching 4D (BM4D). 
 Another successful method is low-rank matrix factorization \cite{peng2021low}. Zhang \textit{et al.} \cite{6648433} was the first to introduce it to the field of RSI denoising. After that, Xu \textit{et al.} \cite{7932170} constructed robust principal component analysis with it to accelerate the iterative optimization process. Nevertheless, matrix factorization-based methods require 3D RSI to be converted into 2D data first, which may result in some loss of image information. Since RSI is essentially 3D image, leveraging tensor representation is very suitable for RSI denoising. In recent years, there has been considerable advancement in low-rank tensor factorization. For example, Chang \textit{et al.} \cite{8100108} proposed a hyper-Laplacian regularized unidirectional low-rank tensor recovery (LLRT) method and showed its advantages over matrix-based models. Wang \textit{et al.} \cite{8233403} incorporated the total variation prior into low-rank tensor decomposition (LRTDTV), thus achieving excellent RSI denoising performance. Recently, He \textit{et al.} \cite{8954159} introduced non-local self-similarity and developed a promising denoiser named non-local meets global (NGMeet), which can fully exploit both global and local geometric structures. 
Zha \textit{et al.} \cite{10106506} introduced a nonlocal structured sparsity regularization (NLSSR) method, which simultaneously exploits global spectral characteristics and nonlocal structured sparsity priors to improve the denoising of RSI.
It is worth noting that although the above model-based methods provide strong interpretability and theoretical guarantees, they often require tedious tuning of regularization parameters to achieve optimal performance.

Deep learning-based methods, which can dynamically adjust the trainable parameters based on external data and labels, have received extensive attention in RSI denoising \cite{PAN2023109832}. 
For instance, Yuan \textit{et al.} \cite{8454887} developed a spatial-spectral convolutional neural network (CNN) that exploits the nonlinear relationship between noisy and clean RSI. Wei \textit{et al.} \cite{9046853} developed a 3D quasi-recurrent neural network by employing 3D convolution to simultaneously extract RSI features.  Maffei \textit{et al.} \cite{8913713} enhanced denoising efficiency by introducing a reversible downsampling operator and utilizing a noise level map to flexibly handle different levels of noise. This method is abbreviated as hyperspectral image (HSI) single denoising CNN (HSI-SDeCNN). 
Zhuang \textit{et al.} \cite{10475370} decomposed remote sensing image features and combined them with CNN to improve RSI denoising performance.
In fact, when there is enough training data, deep learning-based methods often outperform model-based ones. However, since neural networks are black boxes, deep learning-based methods have poor interpretability, which makes it difficult to understand the underlying denoising mechanism.

Recently, researchers have tried to combine the advantages of deep learning with the high interpretability of model-based methods, and some excellent works have emerged \cite{xie2020mhf}. 
Zhuang \textit{et al.} \cite{9552462} integrated a deep learning network named fast and flexible denoising network (FFDNet) \cite{8365806} into the subspace representation framework and proposed a fast and parameter free HSI mixed noise removal method (FastHyMix).
Xiong \textit{et al.} \cite{9631264} extended the low-rank matrix model with spatial deep priors for RSI denoising. To handle multidimensional structures, Xiong \textit{et al.} \cite{9855427} constructed a subspace-based multidimensional sparse tensor model and then unrolled it into a neural network (SMDS-Net). The non-local self-similarity prior captures the repetitive patterns of textures and structures across distant regions within an image, enabling effective preservation of edges and fine details. Unfortunately, SMDS-Net overlooks the non-local self-similarity of RSI. Very recently, Peng \textit{et al.} \cite{10413643} suggested a framework based on the representative coefficient image (RCI) with spectral low-rank tensor decomposition (RCILD). However, in terms of interpretability, this method retrains denoising CNN (DnCNN) \cite{7839189} as the RCI denoiser within the iterative optimization process. 
The above methods either fail to fully unroll the iterative process of the entire physical model into the network or overlook non-local self-similarity, which may result in inaccurate learned geometric structures.
\textit{Therefore, a natural question arises whether it is possible to propose a novel end-to-end network that can address ``lack of non-local self-similarity, sensitivity to parameter tuning, and interpretability of neural networks"?}

\begin{figure*}[t]
	\centering
	\includegraphics[width=4.9 in]{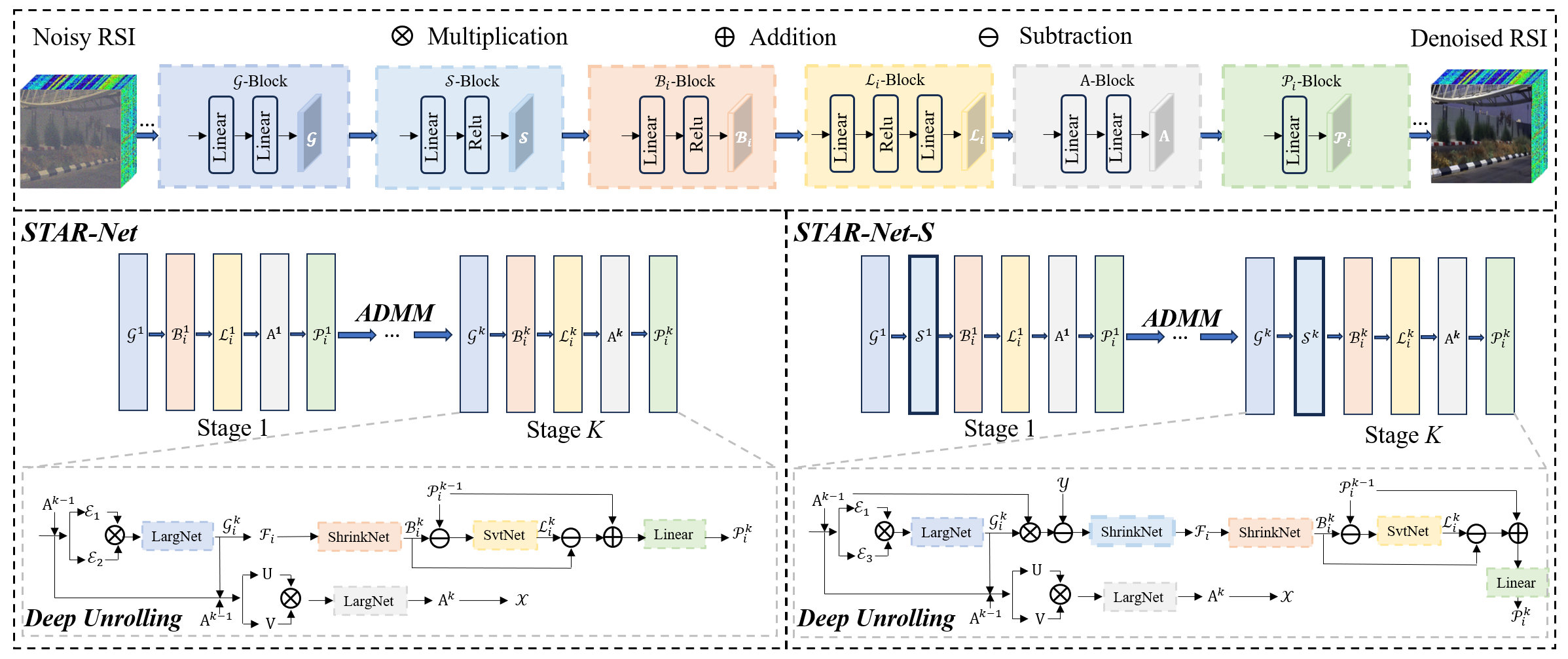}
	\caption{Illustration of the proposed STAR-Net and STAR-Net-S with $K$ stages. The denoised RSI is obtained through $K$ iterations after the noisy RSI is fed into the network. STAR-Net and STAR-Net-S employ the ADMM framework, unrolling each iteration into a network structure.}
	\label{proposed}
\end{figure*}

Inspired by the above observations, this paper first introduces a low-rank prior to characterize the non-local self-similarity, which is often ignored in the literature. 
Meanwhile, matrix-based methods reshape the data from 3D to 2D, which often disrupts the inherent structural self-similarity of the original RSI \cite{9359537}.
Unlike matrix representation, tensor representation can fully exploit prior knowledge and more effectively preserve the complete spatial-spectral structure of RSI, thereby better capturing the inherent non-local self-similarity.
Therefore, a tensor representation model is established to improve the interpretability. 
Finally, a deep unrolling strategy \cite{monga2021algorithm} is adopted to treat the regularization parameters in iterative optimization as learnable parameters in neural networks, thereby avoiding the tedious tuning of regularization parameters. 
Different from SMDS-Net and RCILD, our method leverages the alternating direction method of multipliers (ADMM) to iteratively optimize the proposed model and fully unrolls the iterative process into an end-to-end network, where all regularization parameters are treated as learnable components.
For convenience, we call it sparse tensor-aided representation network (STAR-Net). 
In addition, we modify STAR-Net by introducing an additional prior to handle the sparse noise often present in original RSI, which is called STAR-Net-S.
Figure \ref{proposed} shows the network framework, and more explanation will be provided in Section \ref{sec3}. 

Compared with the existing work, the primary contributions of this paper are summarized as follows.
\begin{itemize}
\item We propose a novel RSI denoising model named STAR-Net, which incorporates a low-rank prior to effectively exploit the non-local self-similarity property of RSI, thereby improving the edge and detail preservation ability and denoising performance.
\item We further introduce a sparse prior to extend STAR-Net into a sparse variant called STAR-Net-S, thereby enhancing robustness to the non-Gaussian noise commonly present in real-world RSI and enabling better preservation of critical spatial and spectral information.
\item We develop an ADMM-guided deep unrolling network that enables end-to-end learning of all regularization parameters, combining the interpretability of model-based methods with the flexibility and efficiency of deep learning-based methods, thus avoiding tedious manual parameter tuning.
\end{itemize}

The remainder of this paper is structured as follows. 
Section \ref{sec2} introduces the notations and related work. 
Section \ref{sec3} describes the proposed models, algorithms, and deep unrolling networks.
Section \ref{sec4} provides the denoising experiments and discussions. 
Finally, Section \ref{sec5} provides a conclusion to this paper.

\section{Preliminaries}\label{sec2}

\subsection{Tensor Representation Model}

For RSI, tensor representation facilitates simultaneous smoothing of the spectral and spatial dimensions, ensuring that both spectral consistency and spatial structure are preserved while effectively removing noise \cite{10068806}. However, directly processing high-dimensional tensors usually takes a lot of time. Fortunately, this challenge can be addressed by adopting subspace representation \cite{4556647}.
By projecting the data into a low-dimensional space, subspace representation retains essential spectral information and improves computational efficiency.
\begin{figure*}[t]
	\centering
	\includegraphics[width=4 in]{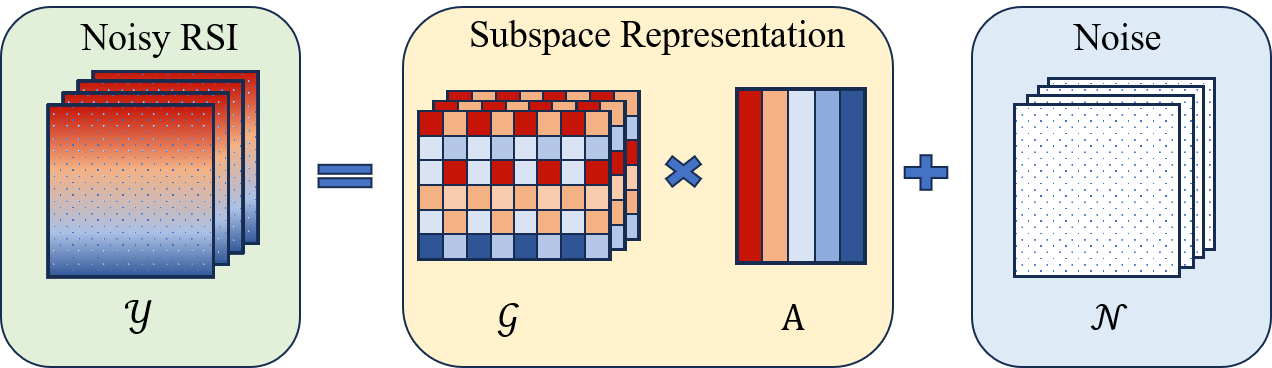}
	\caption{Illustration of tensor low-rank decomposition.  $\mathcal{Y}$ is the noisy RSI, $\mathcal{G}$ is the RCI, $\mathrm{A}$ is an orthogonal basis matrix, and $\mathcal{N}$ is the contaminated noise.}
	\label{subspace}
\end{figure*}
 
Assume that the RSI tensor is $\mathcal{Y} \in \mathbb{R}^{n_{1} \times n_{2}\times n_{3}}$. Then it can be decomposed into
\begin{equation} \label{model}
	\begin{aligned} 
	   \mathcal{Y} = \mathcal{X} +	\mathcal{N},
	\end{aligned}
\end{equation}
where $\mathcal{X}$ is the clean RSI, and $\mathcal{N}$ is the contaminated noise. 

Given the spectral low-rank nature of RSI, the subspace representation of a clean RSI can effectively capture its inherent spectral redundancy \cite{10035509}. As a result, the clean RSI can be approximately expressed through the following decomposition
\begin{equation} \label{model111}
	\begin{aligned} 
	   \mathcal{X} = \mathcal{G}\times_3 \mathrm{A} ,
	\end{aligned}
\end{equation}
where $\mathcal{G} \in \mathbb{R}^{n_{1} \times n_{2}\times n_{4}}$ is the RCI with $n_{4} \ll n_{3}$, $\mathrm{A}$ is an orthogonal basis matrix. An illustration of this process is provided in Figure \ref{subspace}.

Considering the prior information of $\mathcal{G}$, SMDS-Net in \cite{9855427} is formulated as
 \begin{equation} \label{model-smds}
	\begin{aligned} 
		\min_{\mathcal{G}, \mathcal{B}_i, \mathrm{A}}~&\frac{1}{2}\|\mathcal{Y}-\mathcal{G}\times_3 \mathrm{A}\|_\textrm{F}^2+\lambda \sum_i(\phi(\mathcal{G}, \mathcal{B}_i)+ \gamma_1 \|\mathcal{B}_i\|_1)\\
		\rm{s.t.}~~~&\mathrm{A}^\top \mathrm{A}=\mathrm{I},
	\end{aligned}
\end{equation}
of which $\|\cdot\|_F$ is the Frobenius norm, $\|\cdot\|_1$ is the $\ell_{1}$-norm, $\lambda, \gamma_1 > 0$ are the trade-off parameters to balance $\phi(\mathcal{G}, \mathcal{B}_i)$ and $\|\mathcal{B}_i\|_1$. Here,
\begin{equation}
\phi(\mathcal{G}, \mathcal{B}_i)=\frac{1}{2}\|\mathcal{R}_i\mathcal{G}-\mathcal{B}_i\times_1 \mathrm{D}_1 \times_2 \mathrm{D}_2 \times_3 \mathrm{D}_3\|_\textrm{F}^2,
\end{equation}
where $\mathrm{D}_j~(j=1, 2, 3)$ is the dictionary matrix along the $j$-th mode. Obviously, the size of dictionaries will affect performance, as discussed in Section \ref{dic}.
Note that $\mathcal{R}_i$ denotes the operator that extracts $\mathcal{G}_i$ from $\mathcal{G}$, and $i$ is the number of extracted tensors.

\subsection{Deep Unrolling}

Deep unrolling is a new technique that closely integrates iterative optimization algorithms with deep learning \cite{10136197}. The core involves progressively unrolling the traditional iterative optimization process, treating each iteration step as a neural network layer.
Many classical optimization algorithms, such as the ADMM, rely on a sequence of iterative steps to seek the optimal solution. These algorithms typically operate in a fixed manner and may struggle to adapt to complex data distributions. Through unrolling these iterative steps layer-by-layer, deep unrolling techniques introduce learnable parameters to replace the regularization parameters, allowing the model to be automatically optimized via deep learning. Yang \textit{et al.} \cite{8550778} proposed a model-based compressed sensing (CS) method optimized using ADMM, which was unrolled into a network, resulting in the ADMM-CSNet. Li \textit{et al.} \cite{LI2024110412} proposed interpretable sampler and implicit regularization learning network (ISP-IRLNet), where the regularized reconstruction model is optimized using ADMM and unrolled into a deep network with learnable implicit regularization subnetworks.

By merging the expressive power of deep learning with the theoretical strengths of traditional optimization methods, deep unrolling not only preserves the interpretability of the model but also eliminates the need for tedious parameter tuning typically required by traditional algorithms.
Inspired by this, we will propose RSI denoising methods by unrolling the ADMM-based models into deep networks.  

\section{The Proposed Method}\label{sec3}

This section presents the mathematical model, optimization algorithm, and deep unrolling network for STAR-Net and STAR-Net-S.

\subsection{STAR-Net} 

In fact, SMDS-Net overlooks the non-local self-similarity of RSI and fails to account for the impact of non-Gaussian noise present in the original RSI, both of which are critical for effective RSI denoising.
In this section, we first integrate a non-local prior to account for the non-local self-similarity of RSI and propose the STAR-Net. 
Different from the previous RSI denoising methods, we adopt ADMM for iterative optimization of the proposed model and gradually unroll it into an end-to-end network.
\begin{equation} \label{model-star}
	\begin{aligned} 
		\min_{\mathcal{G}, \mathcal{B}_i, \mathrm{A}}~&\frac{1}{2}\|\mathcal{Y}-\mathcal{G}\times_3 \mathrm{A}\|_\textrm{F}^2+\lambda \sum_i(\phi(\mathcal{G}, \mathcal{B}_i)+\gamma_1 \|\mathcal{B}_i\|_1+ \gamma_2 \|\mathcal{B}_i\|_*)\\
		\rm{s.t.}~~~&\mathrm{A}^\top \mathrm{A}=\mathrm{I},
	\end{aligned}
\end{equation}
where $\|\cdot\|_*$ is the tensor nuclear norm, $ \gamma_2>0$ is the regularization parameter. Obviously, if $ \gamma_2\to 0$,  \eqref{model-star} will reduce to  \eqref{model-smds}.

For computational ease, we here introduce an auxiliary variable $\mathcal{L}_i = \mathcal{B}_i$  and reformulate \eqref{model-star} as 
\begin{equation} \label{model-star-1}
	\begin{aligned} 
		\min_{\mathcal{G}, \mathcal{B}_i, \mathcal{L}_i, \mathrm{A}}~&\frac{1}{2}\|\mathcal{Y}-\mathcal{G}\times_3 \mathrm{A}\|_\textrm{F}^2+\lambda \sum_i(\phi(\mathcal{G}, \mathcal{B}_i) +\gamma_1 \|\mathcal{B}_i\|_1+ \gamma_2 \|\mathcal{L}_i\|_*)\\
		\rm{s.t.}~~~~&\mathrm{A}^\top \mathrm{A}=\mathrm{I},~\mathcal{L}_i = \mathcal{B}_i.
	\end{aligned}
\end{equation}
The augmented Lagrangian function of \eqref{model-star-1} is
\begin{equation}\label{Lagrange}
	\begin{aligned}
		&L_{\beta}(\mathcal{G}, \mathcal{B}_{i}, \mathcal{L}_{i}, \mathrm{A}, \mathcal{P}_{i})\\
&=\frac{1}{2}\|\mathcal{Y}-\mathcal{G}\times_3 \mathrm{A}\|_\textrm{F}^2+\lambda \sum_i(\phi(\mathcal{G}, \mathcal{B}_i)+\gamma_1 \|\mathcal{B}_i\|_1+ \gamma_2 \|\mathcal{L}_i\|_*)\\
&+\langle\mathcal{P}_{i}, \mathcal{L}_{i}-\mathcal{B}_{i}\rangle + \frac{\beta }{2}  \| \mathcal{L}_{i}-\mathcal{B}_{i} \|_\textrm{F}^{2},
	\end{aligned}
\end{equation}
where $\mathcal{P}_{i}$ is the Lagrangian multiplier, $\beta$ is the penalty parameter. 
By using the ADMM, it can be solved by
\begin{subnumcases}{}
\mathcal{G}^{k+1}=\mathop{\arg\min}\limits_{\mathcal{G}}~L_{\beta}(\mathcal{G}, \mathcal{B}_{i}^k, \mathcal{L}_{i}^k, \mathrm{A}^k, \mathcal{P}_{i}^k),\label{eq:g} \\
\mathcal{B}_{i}^{k+1}=\mathop{\arg\min}\limits_{\mathcal{B}_{i}}~L_{\beta}(\mathcal{G}^{k+1}, \mathcal{B}_{i}, \mathcal{L}_{i}^k, \mathrm{A}^k, \mathcal{P}_{i}^k),\label{eq:b} \\
\mathcal{L}_{i}^{k+1}=\mathop{\arg\min}\limits_{\mathcal{L}_{i}}~L_{\beta}(\mathcal{G}^{k+1}, \mathcal{B}_{i}^{k+1}, \mathcal{L}_{i}, \mathrm{A}^k, \mathcal{P}_{i}^k),\label{eq:l} \\
\mathrm{A}^{k+1}=\mathop{\arg\min}\limits_{\mathrm{A}}~L_{\beta}(\mathcal{G}^{k+1}, \mathcal{B}_{i}^{k+1}, \mathcal{L}_{i}^{k+1}, \mathrm{A}, \mathcal{P}_{i}^k),\label{eq:a} \\
\mathcal{P}_{i}^{k+1}=\mathcal{P}_{i}^{k}+\beta(\mathcal{L}_{i}^{k+1}-\mathcal{B}_{i}^{k+1}).\label{eq:p}
\end{subnumcases}
Now, we will show how to solve these subproblems by deep unrolling techniques.

\subsubsection{$\mathcal{G}$-Block} The  $\mathcal{G}$-subproblem in \eqref{eq:g} can be simplified to the following optimization problem
\begin{equation}\label{GG}
	\begin{aligned}
		\min_{\mathcal{G}}~& \frac{1}{2}\|\mathcal{Y}-\mathcal{G}\times_3 \mathrm{A}^k\|_\textrm{F}^2 + \frac{\lambda}{2}\sum_i \|\mathcal{R}_i\mathcal{G}-\mathcal{B}_i^k\times_1 \mathrm{D}_1 \times_2 \mathrm{D}_2 \times_3 \mathrm{D}_3\|_\textrm{F}^2.\\ 
	\end{aligned}
\end{equation}
Taking the derivative of \eqref{GG}, it derives
\begin{equation}\label{GGG}
	\begin{aligned}
		&\mathcal{G}^{k+1} = (\mathcal{I} + \lambda \sum_i \mathcal{R}_i^\top\mathcal{R}_i )^{-1}(\lambda \sum_i \mathcal{R}_i^\top\mathcal{B}_i^k\times_1 \mathrm{D}_1 \times_2 \mathrm{D}_2 \times_3 \mathrm{D}_3 
+ \mathcal{Y} \times_3\mathrm{A}^{k\top}).
	\end{aligned}
\end{equation}
Denote
$\mathcal{E}_1 = (\mathcal{I} + \lambda \sum_i \mathcal{R}_i^\top\mathcal{R}_i )^{-1}$,
$\mathcal{E}_2 = \lambda \sum_i \mathcal{R}_i^\top\mathcal{B}_i^k\times_1 \mathrm{D}_1 \times_2 \mathrm{D}_2 \times_3 \mathrm{D}_3  + \mathcal{Y} \times_3\mathrm{A}^{k\top}$. Then, $\mathcal{G}^{k+1}$ can be updated by the following network
\begin{equation}
	\begin{aligned} \label{Gnet}
        \mathcal{G}^{k+1} = \textbf{LargNet}(\mathcal{E}_1, \mathcal{E}_2),
	\end{aligned}
\end{equation}
 where \textbf{LargNet}  can be seen as two distinct linear layers. Note that 
$\mathcal{E}_1$ only needs to be calculated once.

\subsubsection{$\mathcal{B}_{i}$-Block} The $\mathcal{B}_{i}$-subproblem in \eqref{eq:b} can be solved by
\begin{equation}\label{BB}
	\begin{aligned}
		\min_{\mathcal{B}_{i}}~&\frac{\lambda}{2}\|\mathcal{R}_i\mathcal{G}^{k+1}-\mathcal{B}_i\times_1 \mathrm{D}_1 \times_2 \mathrm{D}_2 \times_3 \mathrm{D}_3\|_\textrm{F}^2 \\
& + \frac{\beta }{2}  \| \mathcal{L}_{i}^k-\mathcal{B}_{i} + \mathcal{P}^k_{i}/\beta \|_\textrm{F}^{2} + \lambda \gamma_1 \|\mathcal{B}_i\|_1,
	\end{aligned}
\end{equation}
which can be further simplified to
 \begin{equation}\label{BBBB}
	\begin{aligned}
		\min_{\mathcal{B}_{i}}~&\frac{1}{2}\|(\beta\mathcal{I} + \lambda \mathcal{I}\times_1 \mathrm{D}_1 \times_2 \mathrm{D}_2 \times_3 \mathrm{D}_3)\mathcal{B}_i\\
& -(\lambda \mathcal{R}_i\mathcal{G}^{k+1}+\beta \mathcal{L}_i^k + \mathcal{P}_i^k) \|_\textrm{F}^2   + \lambda  \gamma_1 \|\mathcal{B}_i\|_1 .
	\end{aligned}
\end{equation}

Denote 
 \begin{equation}
 \mathcal{F}_i = \mathcal{B}_i+\frac{1}{l} \mathcal{H} ^\top (\lambda \mathcal{R}_i\mathcal{G}^{k+1}
 +\beta\mathcal{L}_i^k + \mathcal{P}_i^k-\mathcal{H} \mathcal{B}_i),
\end{equation}
 where $\mathcal{H} = \beta\mathcal{I} + \lambda \mathcal{I}\times_1 \mathrm{D}_1 \times_2 \mathrm{D}_2 \times_3 \mathrm{D}_3$, and $l>0$ is the  Lipschitz constant. 
Recall that the  iterative shrinkage thresholding algorithm (ISTA) \cite{8578294}, the solution of \eqref{BBBB} can be obtained by the following solution
\begin{equation}
	\begin{aligned}\label{BBBBB}
		\mathcal{B}_{i}^{k+1} &= \mathcal{M}_{{\lambda \gamma_1 } /{l}}(\mathcal{F}_i),
	\end{aligned}
\end{equation}
where $\mathcal{M}_{{\lambda \gamma_1 }/l } = \verb"sgn"(x)\{ |x|-{\lambda \gamma_1 }/l  \}_{+}   $ is the  soft-thresholding operator, where \verb"sgn" is the sign function and $\{ \cdot \}_{+}  = \max (0, x)$.

Considering that the rectified linear unit (ReLU) is similar to $\{ \cdot \}_{+}  = \max (0, x)$. Consequently, the deep unrolling of \eqref{BBBBB} can be written as

\begin{equation}
	\begin{aligned}\label{Bnet}
		\mathcal{B}_{i}^{k+1} &=  \textbf{ShrinkNet}(\mathcal{F}_i,  \lambda\gamma_1/l)\\
        & =\verb"sgn"({\mathcal{F}_i})\circ \verb"ReLU"(|\mathcal{F}_i| -  \lambda\gamma_1/l\mathcal{I} ),
	\end{aligned}
\end{equation}
where $\circ$ denotes the Hadamard product.

\subsubsection{$\mathcal{L}_{i}$-Block} The $\mathcal{L}_{i}$-subproblem in \eqref{eq:l} can be equivalently transformed to
\begin{equation}\label{LL}
	\begin{aligned}
		\min_{\mathcal{L}_{i}}~& \frac{\beta }{2}  \| \mathcal{L}_{i}-\mathcal{B}_{i}^{k+1} + \mathcal{P}_{i}^k/\beta\|_{\textrm{F}}^{2} + \lambda \gamma_2 \|\mathcal{L}_i\|_*.
	\end{aligned}
\end{equation}
Denote $\mathcal{B}_{i}^{k+1} - \mathcal{P}_{i}^k/\beta=\mathcal{U}_{i}\ast \mathcal{W}_{i}\ast \mathcal{V}_{i}^\top$. Then, according to the singular value thresholding (SVT) \cite{cai2010singular}, it has
\begin{equation} \label{LLLL}
	\begin{aligned}
		\mathcal{L}_{i}^{k+1} =  \mathcal{U}_{i}\ast \verb"diag"( \{ \mathcal{W}_{i}  - \lambda\gamma_2/\beta \mathcal{I} \}_{+}  ) \ast \mathcal{V}_{i}^\top.
	\end{aligned}
\end{equation}
Following the similar idea as \eqref{Bnet}, the above solution can also be described by the following network
\begin{equation} \label{Lnet}
	\begin{aligned}
	\mathcal{L}_{i}^{k+1} &=  \textbf{SvtNet}(\mathcal{B}_{i}^{k+1} - \mathcal{P}_{i}^k/\beta,  \lambda \gamma_2/\beta)\\
     &=  \mathcal{U}_i\ast \verb"ReLU"(\verb"diag" (\mathcal{W}_{i} - \lambda \gamma_2/\beta \mathcal{I}  ) )\ast \mathcal{V}_i^\top.
	\end{aligned}
\end{equation}
 
\subsubsection{$\mathrm{A}$-Block} The  $\mathrm{A}$-subproblem in \eqref{eq:a} can be rewritten as the form of
\begin{equation}\label{AA}
	\begin{aligned}
		\min_{\mathrm{A}}~&\frac{1}{2}\|\mathcal{Y}-\mathcal{G}^{k+1}\times_3 \mathrm{A}\|_\textrm{F}^2\\
         \rm{s.t.}~&\mathrm{A}^\top \mathrm{A}=\mathrm{I}.
	\end{aligned}
\end{equation}
In fact, it is a reduced rank Procrustes rotation problem \cite{Zou2006SparsePC}. 
Denote $\verb"unfold"(\mathcal{Y}, 3)\\
\verb"unfold"(\mathcal{G}^{k+1}, 3)^\top=\mathrm{U}\Sigma\mathrm{V}^\top$, where $\verb"unfold"$ means unfolding the tensor along the $i$-th dimension. Then, it has the solution given by
\begin{equation}
\mathrm{A}^{k+1} = \mathrm{U}\mathrm{V}^\top,
\end{equation}
which can be obtained in the following network
\begin{equation} \label{Anet}
	\begin{aligned}
		\mathrm{A}^{k+1} = \textbf{LargNet}(\mathrm{U}, \mathrm{V}^\top).
	\end{aligned}
\end{equation}

\subsubsection{$\mathcal{P}_{i}$-Block} The $\mathcal{P}_{i}$-subproblem in \eqref{eq:p} is
\begin{equation}
\mathcal{P}_{i}^{k+1}=\mathcal{P}_{i}^{k}+\beta(\mathcal{L}_{i}^{k+1}-\mathcal{B}_{i}^{k+1}),
\end{equation}
which can be solved by the following network
\begin{equation}\label{Pnet}
	\begin{aligned}
		\mathcal{P}_{i}^{k+1} = \textbf{Linear}(\Theta_i),
	\end{aligned}
\end{equation}
where \textbf{Linear} represents the linear layer, $\Theta_i$ can be calculated by $\mathcal{P}_{i}^k + \beta  (\mathcal{L}_{i}^{k+1}  + \mathcal{B}_{i}^{k+1} )$ with $\beta$ being a learnable parameter.

To sum up, the whole deep unrolling framework for STAR-Net is presented in Algorithm \ref{ADMM}. 

\begin{algorithm}[t]
	\caption{Deep unrolling for STAR-Net} \label{ADMM}
	\textbf{Input:} Noisy RSI data $\mathcal{Y}$, parameters $\lambda, \gamma_1, \gamma_2, \beta$\\
	\textbf{Initialize:} $(\mathcal{G}^0,\mathcal{B}_i^0,\mathcal{L}_i^0, \mathrm{A}^0, \mathcal{P}_{i}^0)$\\
	\textbf{While} not converged \textbf{do}	
	\begin{algorithmic}[1]
		\STATE  Update $\mathcal{G}$-Block by
		\[\mathcal{G}^{k+1} = \textbf{LargNet}(\mathcal{E}_1, \mathcal{E}_2)\]
		\STATE  Update  $\mathcal{B}_i$-Block by
		\[\mathcal{B}_i^{k+1}=\textbf{ShrinkNet}(\mathcal{F}_i,  \lambda\gamma_1/l)\]
		\STATE  Update  $\mathcal{L}_i$-Block by
		\[\mathcal{L}_i^{k+1}=\textbf{SvtNet}(\mathcal{B}_{i}^{k+1} - \mathcal{P}_{i}^k/\beta,  \lambda \gamma_2/\beta)\]
		\STATE  Update  $\mathrm{A}$-Block by
		\[\mathrm{A}^{k+1} = \textbf{LargNet}(\mathrm{U}, \mathrm{V}^\top)\]
		\STATE  Update  $\mathcal{P}_{i}$-Block by
		\[\mathcal{P}_{i}^{k+1} = \textbf{Linear}(\Theta_i)\]
	\end{algorithmic}
	\textbf{End While}\\
	\textbf{Output:} Denoised RSI data $\mathcal{X} = \mathcal{G}^{k+1} \times_3 \mathrm{A}^{k+1}$
\end{algorithm}

\subsection{STAR-Net-S} 

Besides Gaussian noise, the original RSI is commonly influenced by non-Gaussian noise as well.
We can further reformulate \eqref{model-star} as
\begin{equation}\label{model-stars}
	\begin{aligned}
		\min_{\mathcal{G}, \mathcal{S}, \mathcal{B}_i, \mathrm{A}}~&\frac{1}{2}\|\mathcal{Y}-\mathcal{G}\times_3 \mathrm{A}-\mathcal{S}\|_\textrm{F}^2+\mu\|\mathcal{S}\|_1+\lambda \sum_i(\phi(\mathcal{G}, \mathcal{B}_i) +\gamma_1 \|\mathcal{B}_i\|_1+ \gamma_2 \|\mathcal{B}_i\|_*) \\
		\rm{s.t.}~~~~&\mathrm{A}^\top \mathrm{A}=\mathrm{I},
	\end{aligned}
\end{equation}
where $\mathcal{S}$ denotes the sparse noise, and $\mu >0$ is the regularization parameter to control the sparaity. In this paper, we refer to \eqref{model-stars} as STAR-Net-S.

Similar to STAR-Net, \eqref{model-stars} can be represented in the following equivalent form
\begin{equation}\label{model-stars2}
	\begin{aligned}
		\min_{\mathcal{G}, \mathcal{S}, \mathcal{B}_i, \mathcal{L}_i, \mathrm{A}}~&\frac{1}{2}\|\mathcal{Y}-\mathcal{G}\times_3 \mathrm{A}-\mathcal{S}\|_\textrm{F}^2+ \mu\|\mathcal{S}\|_1+\lambda \sum_i(\phi(\mathcal{G}, \mathcal{B}_i) +\gamma_1 \|\mathcal{B}_i\|_1+ \gamma_2 \|\mathcal{L}_i\|_*) \\
		\rm{s.t.}~~~~&\mathrm{A}^\top \mathrm{A}=\mathrm{I}, ~\mathcal{L}_i = \mathcal{B}_i,
	\end{aligned}
\end{equation}
and the augmented Lagrangian function is
\begin{equation}\label{Lagrange2}
	\begin{aligned}
		&L_{\beta } (\mathcal{G}, \mathcal{S}, \mathcal{B}_i, \mathcal{L}_i, \mathrm{A},\mathcal{P}_{i})\\
&=\frac{1}{2}\|\mathcal{Y}-\mathcal{G}\times_3 \mathrm{A} - \mathcal{S}\|_\textrm{F}^2 + \mu\|\mathcal{S}\|_1 +\lambda \sum_i(\phi(\mathcal{G}, \mathcal{B}_i)+\gamma_1 \|\mathcal{B}_i\|_1+ \gamma_2 \|\mathcal{L}_i\|_*)\\
&+\langle\mathcal{P}_{i}, \mathcal{L}_{i}-\mathcal{B}_{i}\rangle + \frac{\beta }{2}  \| \mathcal{L}_{i}-\mathcal{B}_{i} \|_\textrm{F}^{2}.
	\end{aligned}
\end{equation}
It is found that compared with \eqref{Lagrange}, only the following two blocks are different. 

\begin{algorithm}[t]
	\caption{Deep unrolling for STAR-Net-S} \label{stars}
	\textbf{Input:} Noisy RSI data $\mathcal{Y}$, parameters $\lambda, \mu, \gamma_1, \gamma_2, \beta$\\
	\textbf{Initialize:} $(\mathcal{G}^0,\mathcal{S}^0, \mathcal{B}_i^0,\mathcal{L}_i^0, \mathrm{A}^0, \mathcal{P}_{i}^0)$\\
	\textbf{While} not converged \textbf{do}	
	\begin{algorithmic}[1]
		\STATE  Update $\mathcal{G}$-Block by
		\[\mathcal{G}^{k+1} = \textbf{LargNet}( \mathcal{E}_1, \mathcal{E}_3)\]
		\STATE  Update $\mathcal{S}$-Block by 
		\[\mathcal{S}^{k+1} = \textbf{ShrinkNet}(\mathcal{Y}^{k}-\mathcal{G}^{k+1}\times_3 \mathrm{A}^{k}, \mu)\]
		\STATE  Update  $\mathcal{B}_i$-Block by
		\[\mathcal{B}_i^{k+1}=\textbf{ShrinkNet}(\mathcal{F}_i,  \lambda\gamma_1/l)\]
		\STATE  Update  $\mathcal{L}_i$-Block by
		\[\mathcal{L}_i^{k+1}=\textbf{SvtNet}(\mathcal{B}_{i}^{k+1} - \mathcal{P}_{i}^k/\beta,  \lambda \gamma_2/\beta)\]
		\STATE  Update  $\mathrm{A}$-Block by
		\[\mathrm{A}^{k+1} = \textbf{LargNet}(\mathrm{U}, \mathrm{V}^\top)\]
		\STATE  Update  $\mathcal{P}_{i}$-Block by
		\[\mathcal{P}_{i}^{k+1} = \textbf{Linear}(\Theta_i)\]
	\end{algorithmic}
	\textbf{End While}\\
	\textbf{Output:} Denoised RSI data $\mathcal{X} = \mathcal{G}^{k+1} \times_3 \mathrm{A}^{k+1}$
\end{algorithm}

\subsubsection{$\mathcal{G}$-Block} Fix other variables, $\mathcal{G}^{k+1}$ can be obtained by solving the following optimization problem
\begin{equation}\label{GG2}
	\begin{aligned}
		\mathop{\min}\limits_{\mathcal{G}}~& \frac{1}{2}\|\mathcal{Y}-\mathcal{G}\times_3 \mathrm{A}^{k} - \mathcal{S}^{k}\|_\textrm{F}^2+\lambda \sum_i \frac{1}{2}\|\mathcal{R}_i\mathcal{G}-\mathcal{B}_i^{k}\times_1 \mathrm{D}_1 \times_2 \mathrm{D}_2 \times_3 \mathrm{D}_3\|_\textrm{F}^2.
	\end{aligned}
\end{equation}
It is not hard to obtain
\begin{equation}\label{GGG2}
	\begin{aligned}
		&\mathcal{G}^{k+1} = (\mathcal{I} + \lambda \sum_i \mathcal{R}_i^\top\mathcal{R}_i )^{-1}(\lambda \sum_i \mathcal{R}_i^\top\mathcal{B}_i^{k}\times_1 \mathrm{D}_1 \\
&~~~\times_2 \mathrm{D}_2 \times_3 \mathrm{D}_3 + \mathcal{Y} \times_3\mathrm{A}^{k\top} - \mathcal{S}^{k} \times_3\mathrm{A}^{k\top}).
	\end{aligned}
\end{equation}
Denote $\mathcal{E}_3 = \lambda \sum_i \mathcal{R}_i^\top\mathcal{B}_i^k\times_1 \mathrm{D}_1 \times_2 \mathrm{D}_2 \times_3 \mathrm{D}_3  + \mathcal{Y} \times_3\mathrm{A}^{k\top}- \mathcal{S}^{k} \times_3\mathrm{A}^{k\top}$. Then, $\mathcal{G}^{k+1}$ can be updated by 
\begin{equation}
	\begin{aligned}
		\mathcal{G}^{k+1} = \textbf{LargNet}( \mathcal{E}_1, \mathcal{E}_3)  .
	\end{aligned}
\end{equation}

\subsubsection{$\mathcal{S}$-Block} Fix other variables, $\mathcal{S}^{k+1}$ can be obtained by optimizing the following problem
\begin{equation}\label{SS2}
	\begin{aligned}
		\mathop{\min}\limits_{\mathcal{S}}~&\frac{1}{2}\|\mathcal{Y}-\mathcal{G}^{k+1}\times_3 \mathrm{A}^{k} - \mathcal{S}\|_\textrm{F}^2 + \mu\|\mathcal{S}\|_1.
	\end{aligned}
\end{equation}
Similar to \eqref{BBBB}, it has the solution
\begin{equation} \label{SSS2}
	\begin{aligned}
		\mathcal{S}^{k+1} = \textbf{ShrinkNet}(\mathcal{Y}-\mathcal{G}^{k+1}\times_3 \mathrm{A}^{k}, \mu).
	\end{aligned}
\end{equation}

Besides, the deep unrolling scheme of STAR-Net-S is summarized in Algorithm \ref{stars}.

\section{Experiments and Discussions}\label{sec4}
In this section, STAR-Net and STAR-Net-S are compared with ten state-of-the-art methods, including five model-based methods, i.e., BM4D \cite{6253256}, LLRT \cite{8100108}, LRTDTV \cite{8233403}, NGMeet \cite{8954159}, and NLSSR \cite{10106506}, and five deep learning-based methods, i.e., FastHyMix \cite{9552462}, HSI-SDeCNN \cite{8913713},  SMDS-Net \cite{9855427}, Eigen-CNN \cite{10475370}, and RCILD \cite{10413643}.

\subsection{Experimental Settings}
\subsubsection{Training and Testing}
Similar to \cite{9046853}, we select 100 RSIs from the ICVL dataset to serve as the training set. 
All RSIs are captured using a Specim PS Kappa DX4 hyperspectral camera. 
The images have a resolution of 1392 $\times$ 1300 pixels and encompass 31 spectral bands ranging from 400 to 700 nm. 
Before training, the images are randomly flipped, cropped, and resized to a final size of 56 $\times$ 56 $\times$ 31 before being fed into the network.

The testing set includes synthetic and real-world datasets. 
The ICVL and PaviaU datasets are the synthetic datasets used in this paper.
The PaviaU dataset is collected using the reflective optics system imaging spectrometer (ROSIS) sensor over the city of Pavia in Italy.
The PaviaU dataset has a size of 610 $\times$ 340 $\times$ 103 and is recognized as a large-scale dataset \cite{10677367}.

The real-world datasets consist of the Beijing Capital Airport RSI and the Indian Pines RSI. Here, the Beijing Capital Airport RSI is captured by the Gaofen-5 satellite with 155 bands and 300 $\times$ 300 pixels. As in \cite{8913713}, the Indian Pines dataset is captured by the airborne visible infrared imaging spectrometer (AVIRIS) and consists of 145 $\times$ 145 pixels and 206 spectral bands.
To generate noisy RSI, Gaussian noise is introduced to each band of clean RSI.
Specifically, we consider four different scenarios with standard variance $\sigma$ set at 10, 30, 50, and 70, respectively.
Furthermore, to simulate the effects of non-Gaussian noise, a non-Gaussian noise case is constructed by adding 20\% salt-and-pepper noise to each band and adding dead lines to 20\% of the bands.

\subsubsection{Initialization} 
In general, neural networks are trained using parameters that are initialized randomly.
To accelerate the training process, the network parameters are initialized using the original parameters obtained from the ADMM as described below.

For STAR-Net, first initialize $\mathcal{E}_1, \mathcal{E}_2$ in $\mathcal{G}$-Block through \eqref{Gnet}, mainly involving $\lambda$ in the ADMM.
Then, initialize $\lambda\gamma_1/l$ corresponding to each layer of $\mathcal{B}_i$-Block components according to \eqref{Bnet}, where $\lambda, \gamma_1$ and $l$ are parameters related to the ADMM.
Lastly, initialize $\lambda, \gamma_2, \beta$ involved in $\mathcal{L}_i$-Block and $\mathcal{P}_i$-Block.
For STAR-Net-S, there is an additional parameter $\mu$.
In this paper, all learnable parameters $\gamma_1$, $\gamma_2$, $l$, $\lambda$, $\mu$, $\beta$ are initialized to 0.02. 

\subsubsection{Loss Function} 
For a given training dataset, the loss function is specified as the Euclidean distance between the output of STAR-Net and the ground truth, i.e.,
\begin{equation} \label{loss}
	\begin{aligned}
		Loss = \|\textrm{STAR-Net}(\mathcal{Y}) - \mathcal{X}\|_\textrm{F}^2,
	\end{aligned}
\end{equation}
where $\mathcal{Y}$ denotes the denoised RSI  generated by the network and $\mathcal{X}$ denotes the ground truth, which is the original RSI without noise.
\subsubsection{Implement Details}
The proposed STAR-Net and STAR-Net-S are implemented using the PyTorch framework and trained for 300 epochs on the NVIDIA GeForce RTX 4090 GPU. 
The initial learning rate is set to $\text{5$\times10^{-3}$}$ and is reduced by a factor of 0.35 after every 80 iterations. 
Additionally, the Adam optimizer is employed with a patch size of $\text{56$\times$56}$ and a batch size of 2.
The unrolling iteration \textit{K} is set to 9. 
In addition, the dictionary is initialized using a discrete cosine transform (DCT) basis with dimensions [9, 9, 9], resulting in a dictionary size of $\text{9$\times$9}$ along the three modes. 
Finally, the remaining parameters $\gamma_1$, $\gamma_2$, $l$, $\lambda$, $\mu$, $\beta$ are initialized to 0.02.

\subsubsection{Evaluation Indexes}
To quantitatively assess the denoising performance of all compared methods, we select four commonly used indexes, i.e., peak signal-to-noise ratio (PSNR), structural similarity (SSIM), spectral angle mapper (SAM), and erreur relative globale adimensionnelle de synthese (ERGAS).
In particular, PSNR evaluates the reconstruction accuracy after lossy compression, SSIM measures the perceived alterations in structural information, and SAM describes the spectral difference between the clean and the denoised RSI. 
Considering that the first three metrics have limitations in evaluating structural preservation and spectral fidelity, the ERGAS metric provides a more comprehensive measure of the overall reconstruction error, thereby further validating the effectiveness of the method in maintaining global spectral consistency.
Generally speaking, better denoising performance is indicated by higher PSNR and SSIM values with lower SAM and ERGAS values.

\subsection{Comparison on Synthetic Datasets}

\subsubsection{Experiments on ICVL}

\begin{table*}[t]
	\centering
	\renewcommand\arraystretch{1.2}
	\caption{Comparison of all methods on ICVL. The top two values are marked as \textcolor[rgb]{1.00,0.00,0.00}{red} and \textcolor[rgb]{0.00,0.00,1.00}{blue}.}\label{icvltable}
	\vskip-0.2cm
	\setlength{\tabcolsep}{4.5 pt}
\resizebox{\textwidth}{!}{
	\begin{tabular}{ccccccccccccccc}
		\toprule
		{$\sigma$}   & {Index} & {Noisy}  
		                                                  & \makecell[c]{BM4D\\}   & \makecell[c]{LLRT\\}                       & \makecell[c]{LRTDTV\\} & \makecell[c]{NGMeet\\}& \makecell[c]{NLSSR\\}&\makecell[c]{FastHy\\Mix} & \makecell[c]{HSI-SDe\\CNN}  & \makecell[c]{SMDS-\\Net}    &\makecell[c]{Eigen-\\CNN}   & \makecell[c]{RCILD\\}         & \makecell[c]{STAR-\\Net}        & \makecell[c]{STAR-\\Net-S}    \\ \hline 
		\multirow{4}{*}{10}      & PSNR $\uparrow$    & 29.018   & 42.987 & 39.810 & 43.882  & 42.383 & 45.928 & 43.628         & 41.519     & 46.371      & \textcolor[rgb]{0.00,0.00,1.00}{47.321}         & 42.458      & 47.286         & \textcolor[rgb]{1.00,0.00,0.00}{47.345} \\
                    & SSIM $\uparrow$    & 0.521    & 0.973  & 0.962  & 0.979   & 0.968  & 0.984 & \textcolor[rgb]{0.00,0.00,1.00}{0.988}            & 0.969      & 0.985       & \textcolor[rgb]{1.00,0.00,0.00}{0.989} & 0.987       & \textcolor[rgb]{0.00,0.00,1.00}{ 0.988}    & \textcolor[rgb]{1.00,0.00,0.00}{0.989}  \\
                    & SAM $\downarrow$   & 0.229    & 0.080  & 0.045  & 0.077   & 0.074 & 0.066 & 0.035           & 0.075      & \textcolor[rgb]{0.00,0.00,1.00}{ 0.028} & 0.032          & 0.044       & \textcolor[rgb]{1.00,0.00,0.00}{0.025} & \textcolor[rgb]{1.00,0.00,0.00}{0.025}  \\
                    & ERGAS $\downarrow$ & 243.021  & 36.420 & 59.279       & 44.893  & 34.764 & 28.026 & 24.893          & 61.289     & 20.056      & 25.355         & 25.800      & \textcolor[rgb]{0.00,0.00,1.00}{ 18.124}   & \textcolor[rgb]{1.00,0.00,0.00}{17.951}   \\ \hline
		\multirow{4}{*}{30}      & PSNR $\uparrow$    & 21.591   & 37.630 & 34.250 & 38.245  & 36.791 & 41.629 & 38.286          & 36.840     & 42.337      & 41.491         & 38.514      & \textcolor[rgb]{0.00,0.00,1.00}{42.435}         & \textcolor[rgb]{1.00,0.00,0.00}{42.500} \\
                    & SSIM $\uparrow$    & 0.146    & 0.930  & 0.921  & 0.877   & 0.915  & 0.968    & 0.966   & 0.926           & \textcolor[rgb]{1.00,0.00,0.00}{0.972 }      & 0.963          & \textcolor[rgb]{0.00,0.00,1.00}{ 0.971} & \textcolor[rgb]{1.00,0.00,0.00}{0.972} & \textcolor[rgb]{1.00,0.00,0.00}{0.972}  \\
                    & SAM $\downarrow$   & 0.535    & 0.142  & 0.084  & 0.149   & 0.139  & 0.084 & 0.068             & 0.124      & 0.040       & 0.060          & 0.067       & \textcolor[rgb]{0.00,0.00,1.00}{ 0.039}    & \textcolor[rgb]{1.00,0.00,0.00}{0.038}  \\
                    & ERGAS $\downarrow$ & 729.026  & 62.662 & 40.725       & 69.464  & 60.101  & 41.388  & 48.675         & 103.084    & 32.393      & 49.458         & 49.362      & \textcolor[rgb]{0.00,0.00,1.00}{ 32.056}   & \textcolor[rgb]{1.00,0.00,0.00}{31.862} \\ \hline
		\multirow{4}{*}{50}      & PSNR $\uparrow$    & 18.402   & 35.242 & 32.067 & 33.659  & 34.399 & 39.713 & 35.397           & 34.342     & 37.481      & 36.579         & 35.838      & \textcolor[rgb]{0.00,0.00,1.00}{ 39.853}   & \textcolor[rgb]{1.00,0.00,0.00}{39.963} \\
                    & SSIM $\uparrow$    & 0.042    & 0.888  & 0.899  & 0.862   & 0.887  & \textcolor[rgb]{0.00,0.00,1.00}{ 0.955} & 0.941      & 0.893      & 0.907       & 0.879          & 0.951       & \textcolor[rgb]{1.00,0.00,0.00}{0.956} & \textcolor[rgb]{1.00,0.00,0.00}{0.956}  \\
                    & SAM $\downarrow$   & 0.779    & 0.190  & 0.107  & 0.195   & 0.177 & 0.109  & 0.096            & 0.134      & 0.066       & 0.106          & 0.092       & \textcolor[rgb]{0.00,0.00,1.00}{ 0.050}    & \textcolor[rgb]{1.00,0.00,0.00}{0.047}  \\
                    & ERGAS $\downarrow$ & 1215.105 & 81.133 & 54.869       & 110.351 & 80.585 & 53.811 & 68.681           & 136.304    & 55.786      & 85.426         & 68.945      & \textcolor[rgb]{0.00,0.00,1.00}{ 43.362}   & \textcolor[rgb]{1.00,0.00,0.00}{42.923}  \\ \hline
		\multirow{4}{*}{70}      & PSNR $\uparrow$    & 18.126   & 33.586 & 30.746 & 30.565  & 32.389 & \textcolor[rgb]{0.00,0.00,1.00}{ 37.450} & 33.377     & 32.794     & 37.197      & 32.194         & 33.980      & 37.342         & \textcolor[rgb]{1.00,0.00,0.00}{38.237} \\
                    & SSIM $\uparrow$    & 0.038    & 0.844  & 0.852  & 0.762   & 0.858  & \textcolor[rgb]{0.00,0.00,1.00}{ 0.934} & 0.915       & 0.855      & 0.923       & 0.752          & 0.930       & \textcolor[rgb]{1.00,0.00,0.00}{0.943} & \textcolor[rgb]{1.00,0.00,0.00}{0.943}  \\
                    & SAM $\downarrow$   & 0.897    & 0.231  & 0.214  & 0.304   & 0.217 & 0.128  & 0.120            & 0.186      & 0.066       & 0.154          & 0.098       & \textcolor[rgb]{0.00,0.00,1.00}{ 0.058}    & \textcolor[rgb]{1.00,0.00,0.00}{0.055}  \\
                    & ERGAS $\downarrow$ & 1701.060 & 97.267 & 66.690       & 163.788 & 97.309  & 65.893  & 88.790         & 173.430    & 58.223      & 126.017        & 89.340      & \textcolor[rgb]{0.00,0.00,1.00}{ 57.341}   & \textcolor[rgb]{1.00,0.00,0.00}{52.261} \\ \hline
		\multirow{4}{*}{Average} & PSNR $\uparrow$    & 21.784   & 37.361  & 34.218  & 36.588  & 36.490  & 41.180        & 37.672          & 36.374     & 40.846      & 39.396      & 37.697       & \textcolor[rgb]{0.00,0.00,1.00}{ 41.729}   & \textcolor[rgb]{1.00,0.00,0.00}{42.011}  \\
                         & SSIM $\uparrow$    & 0.187    & 0.909   & 0.909   & 0.870   & 0.907   & \textcolor[rgb]{0.00,0.00,1.00}{ 0.960}   & 0.953           & 0.911      & 0.947       & 0.896       & \textcolor[rgb]{0.00,0.00,1.00}{ 0.960}  & \textcolor[rgb]{1.00,0.00,0.00}{0.965} & \textcolor[rgb]{1.00,0.00,0.00}{0.965}   \\
                         & SAM $\downarrow$   & 0.610    & 0.161   & 0.113   & 0.181   & 0.152   & 0.097         & 0.080           & 0.130      & 0.050       & 0.088       & 0.075        & \textcolor[rgb]{0.00,0.00,1.00}{ 0.043}    & \textcolor[rgb]{1.00,0.00,0.00}{0.041}   \\
                         & ERGAS $\downarrow$ & 972.053  & 69.370  & 55.391  & 97.124  & 68.190  & 47.280        & 57.760          & 118.527    & 41.614      & 71.564      & 58.362       & \textcolor[rgb]{0.00,0.00,1.00}{ 37.721}   & \textcolor[rgb]{1.00,0.00,0.00}{36.249} \\ \bottomrule
	\end{tabular}}
\end{table*}

Table \ref{icvltable} lists the comparison results of PSNR, SSIM, and SAM across various noise levels on the 30 testing RSIs of the ICVL dataset.
In this paper, the best results are highlighted in red, while the second-best results are indicated in blue.
The performance of the model-based method is slightly inferior to that of the deep learning-based method, and there is also a significant residual noise.
Among all methods, STAR-Net and STAR-Net-S achieve the first and second best performances under noise variances of 10, 30, 50, and 70.
In addition, we summarize the average performance of all methods  in Table \ref{icvltable}, where STAR-Net-S shows excellent overall results in four indexes.
This further demonstrates the robust denoising capability of STAR-Net and STAR-Net-S across various noise scenarios.

\begin{figure*}[t]
	\centering
	\subfigure[PSNR(dB)]{
		\includegraphics[width=0.619 in,height=0.619 in]{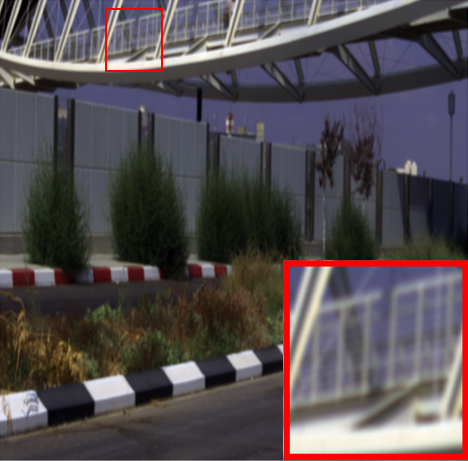}\label{ROC-b21}}
	\subfigure[18.402]{
		\includegraphics[width=0.619 in,height=0.619 in]{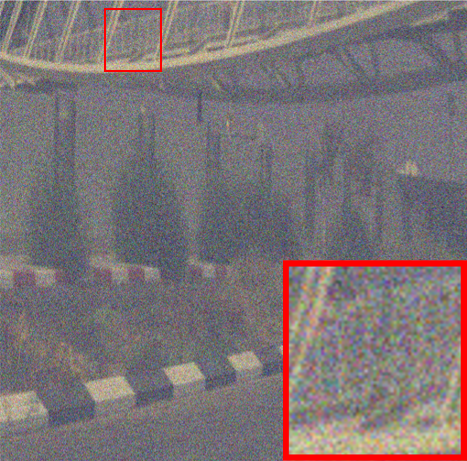}\label{ROC-b22}}
	\subfigure[35.242]{
		\includegraphics[width=0.619 in,height=0.619 in]{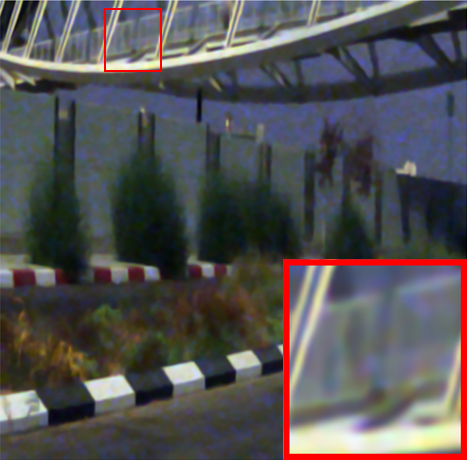}\label{ROC-b22}}
	\subfigure[32.067]{
		\includegraphics[width=0.619 in,height=0.619 in]{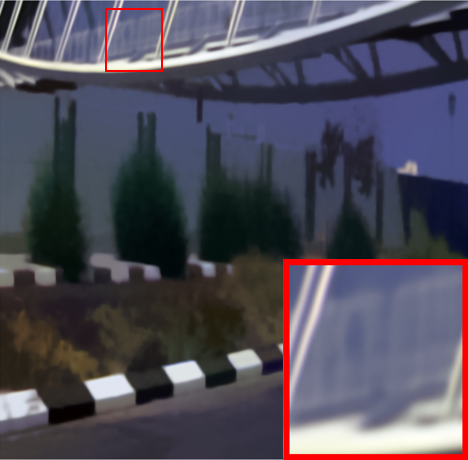}\label{ROC-b22}}
	\subfigure[33.659]{
		\includegraphics[width=0.619 in,height=0.619 in]{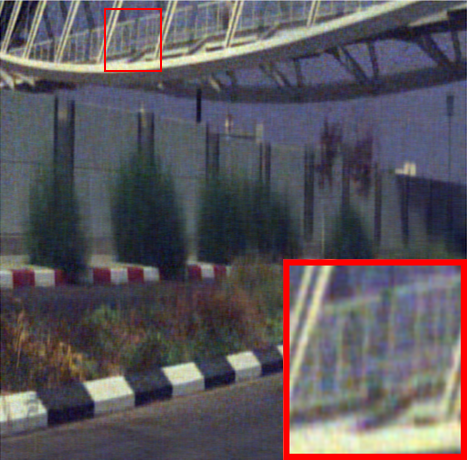}\label{ROC-b22}}
	\subfigure[34.399]{
		\includegraphics[width=0.619 in,height=0.619 in]{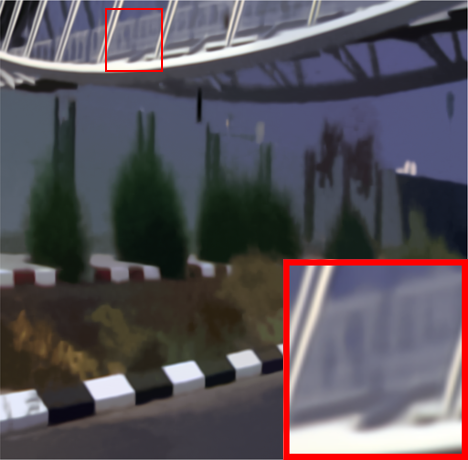}\label{ROC-b22}}
	\subfigure[39.713]{
		\includegraphics[width=0.619 in,height=0.619 in]{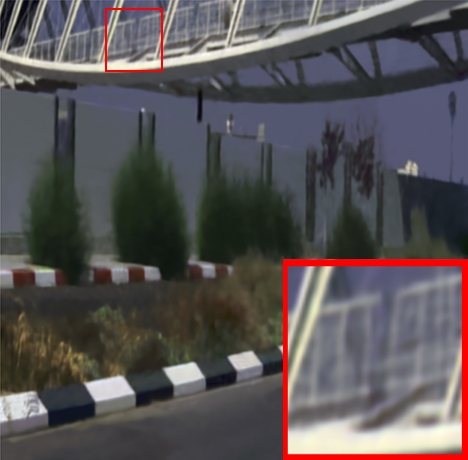}\label{ROC-b22}}
	\subfigure[35.397]{
		\includegraphics[width=0.619 in,height=0.619 in]{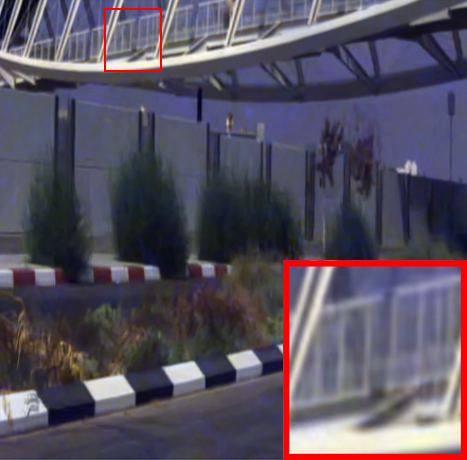}\label{ROC-b22}}
	\subfigure[34.342]{
		\includegraphics[width=0.619 in,height=0.619 in]{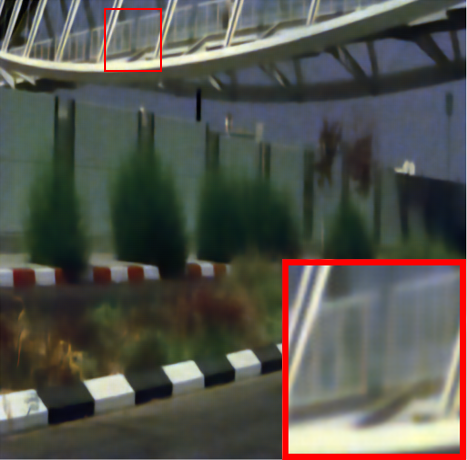}\label{ROC-b22}}
	\subfigure[37.481]{
		\includegraphics[width=0.619 in,height=0.619 in]{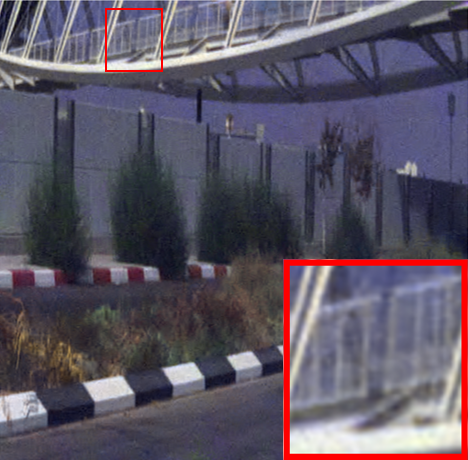}\label{ROC-b23}}
	\subfigure[36.579]{
		\includegraphics[width=0.619 in,height=0.619 in]{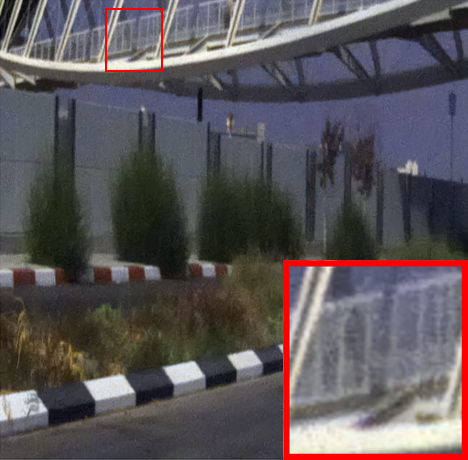}\label{ROC-b22}}
	\subfigure[35.838]{
		\includegraphics[width=0.619 in,height=0.619 in]{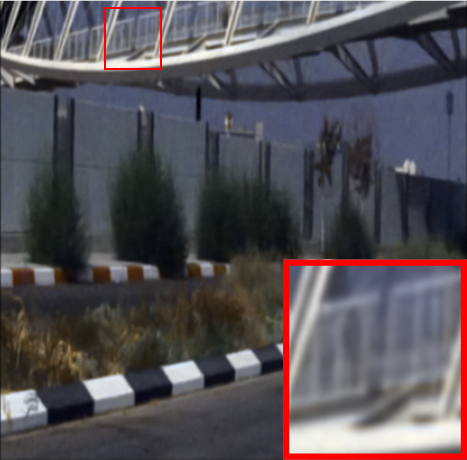}\label{ROC-b22}}
	\subfigure[39.853]{
		\includegraphics[width=0.619 in,height=0.619 in]{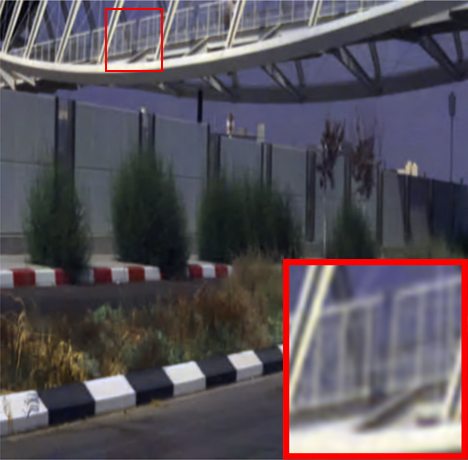}\label{ROC-b24}}
	\subfigure[39.963]{
		\includegraphics[width=0.619 in,height=0.619 in]{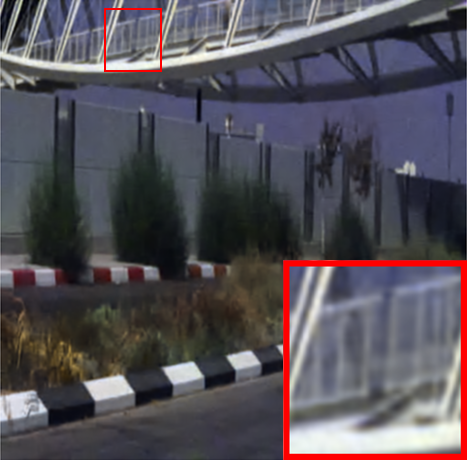}\label{ROC-b24}}
	\vskip-0.2cm
	\caption{Denoising results on gavyam$\text{\_0823-0933}$  with the noise variance of 50. The false-color images are generated by combining bands 5, 18, and 25. (a) Clean, (b) Noisy, (c) BM4D, (d) LLRT, (e) LRTDTV, (f) NGMeet, (g) NLSSR, (h) FastHyMix, (i) HSI-SDeCNN, (j) SMDS-Net, (k) Eigen-CNN, (l) RCILD, (m) STAR-Net, (n) STAR-Net-S. }\label{icvl}
\end{figure*}

\begin{figure*}[t]
  \centering
    \subfigure[]{
    \includegraphics[width=0.619 in,height=0.464 in]{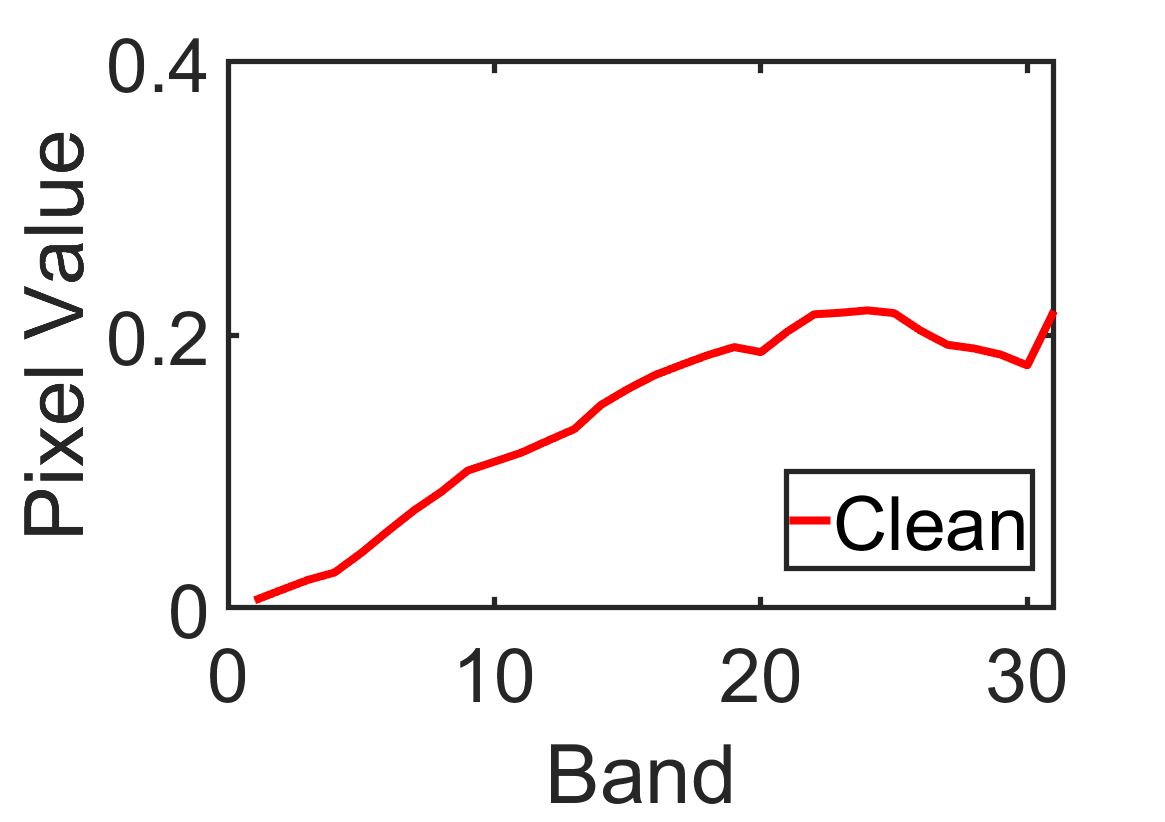}\label{ROC-b21}}
    \subfigure[]{
    \includegraphics[width=0.619 in,height=0.464 in]{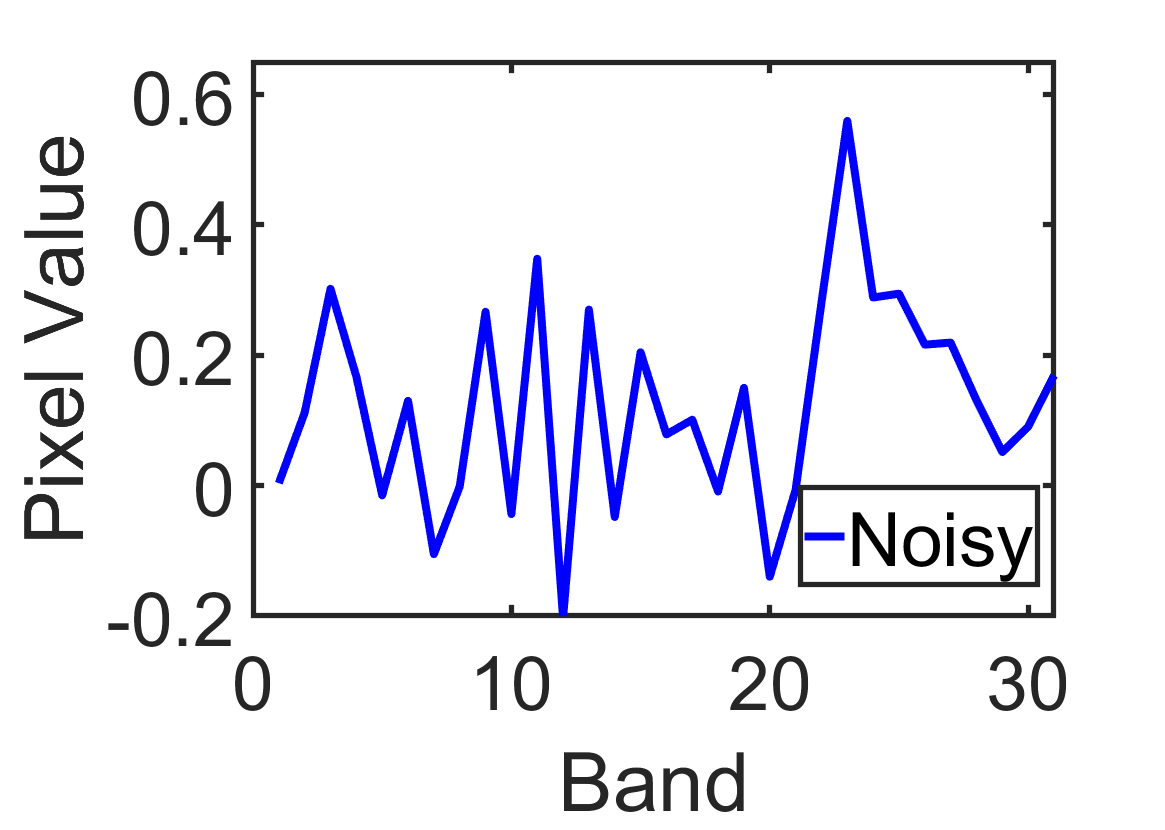}\label{ROC-b22}}
    \subfigure[]{
    \includegraphics[width=0.619 in,height=0.464 in]{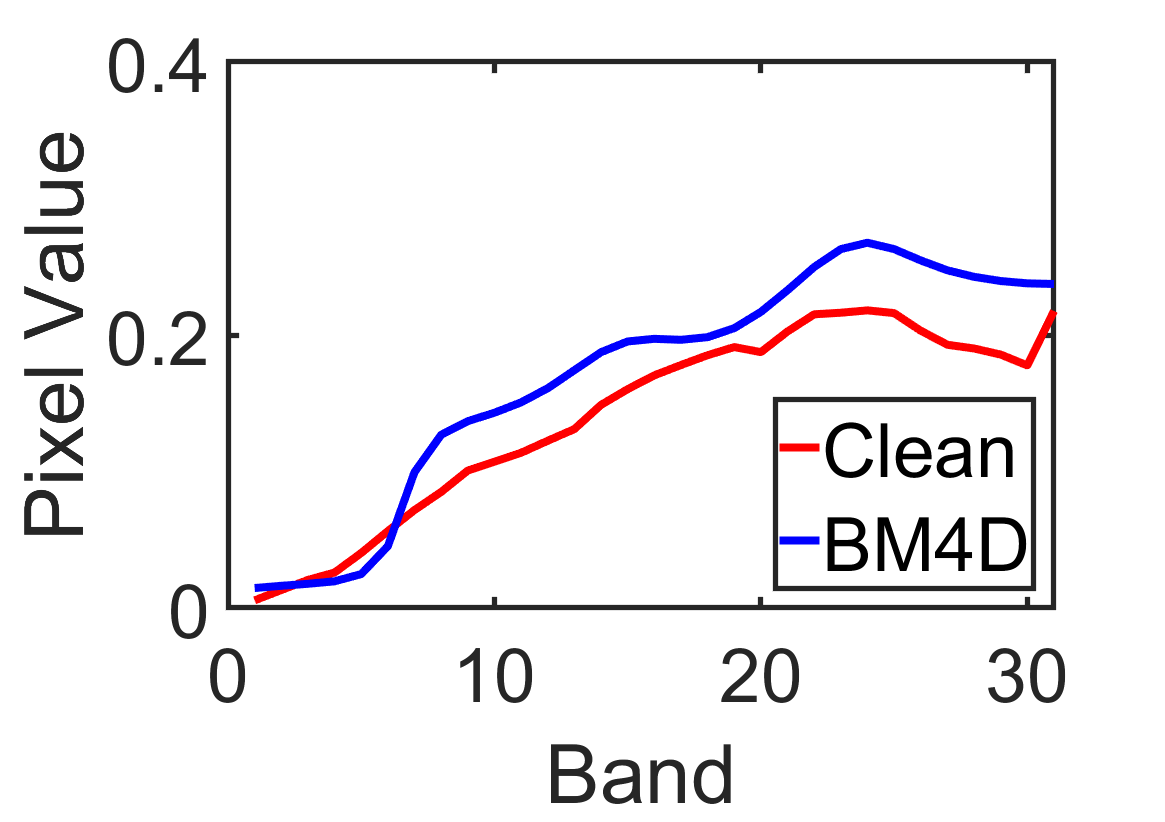}\label{ROC-b22}}
    \subfigure[]{
    \includegraphics[width=0.619 in,height=0.464 in]{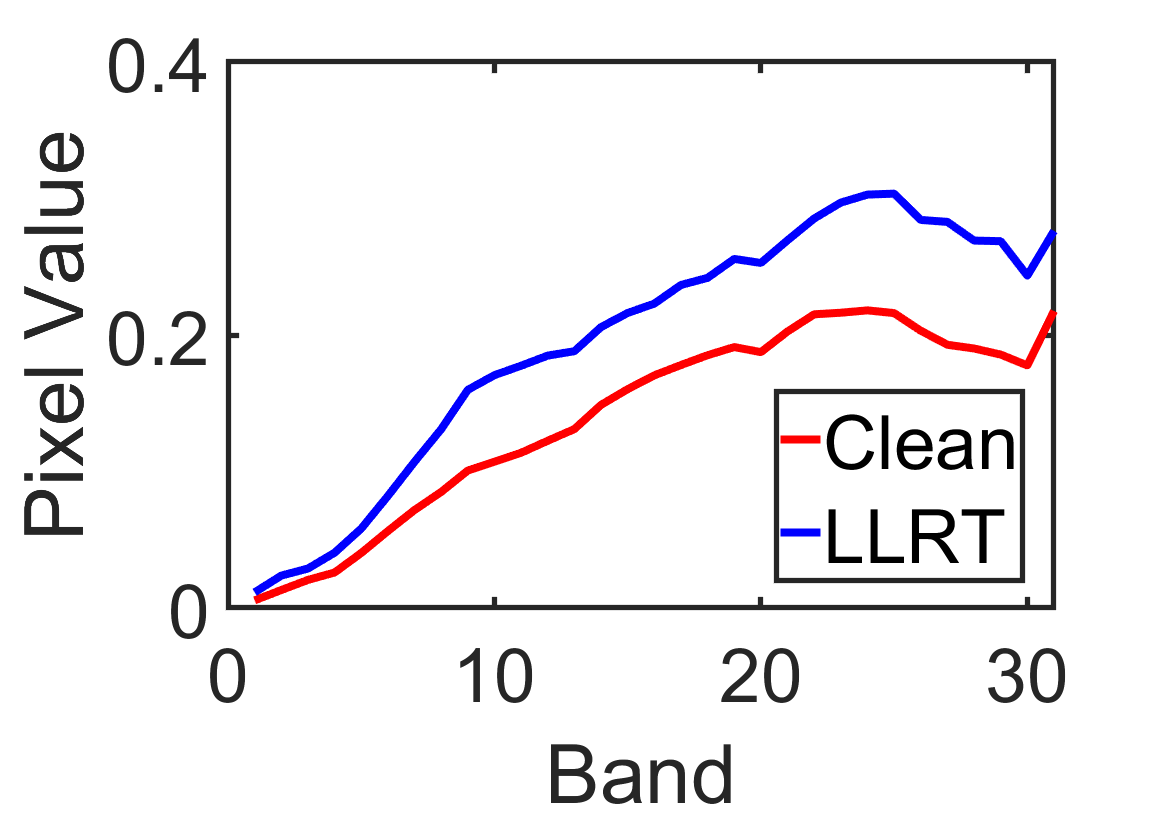}\label{ROC-b23}}
    \subfigure[]{
    \includegraphics[width=0.619 in,height=0.464 in]{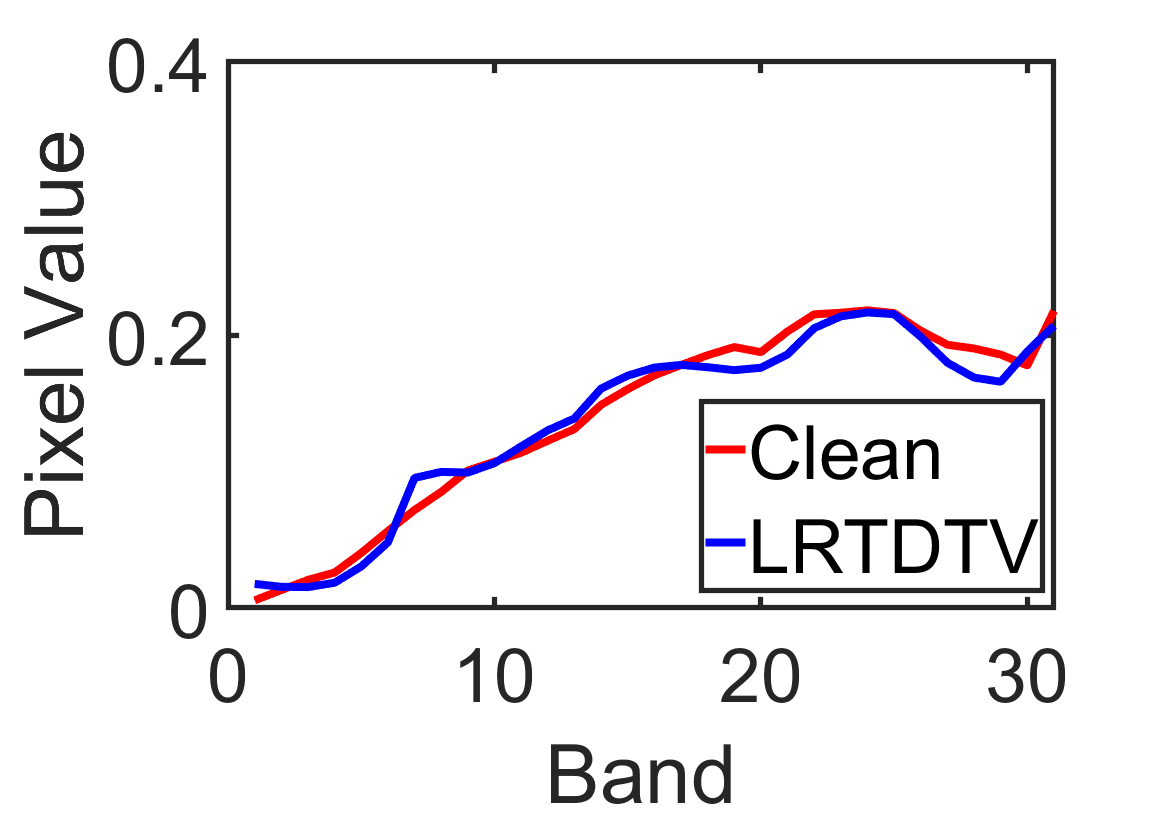}\label{ROC-b22}}
    \subfigure[]{
    \includegraphics[width=0.619 in,height=0.464 in]{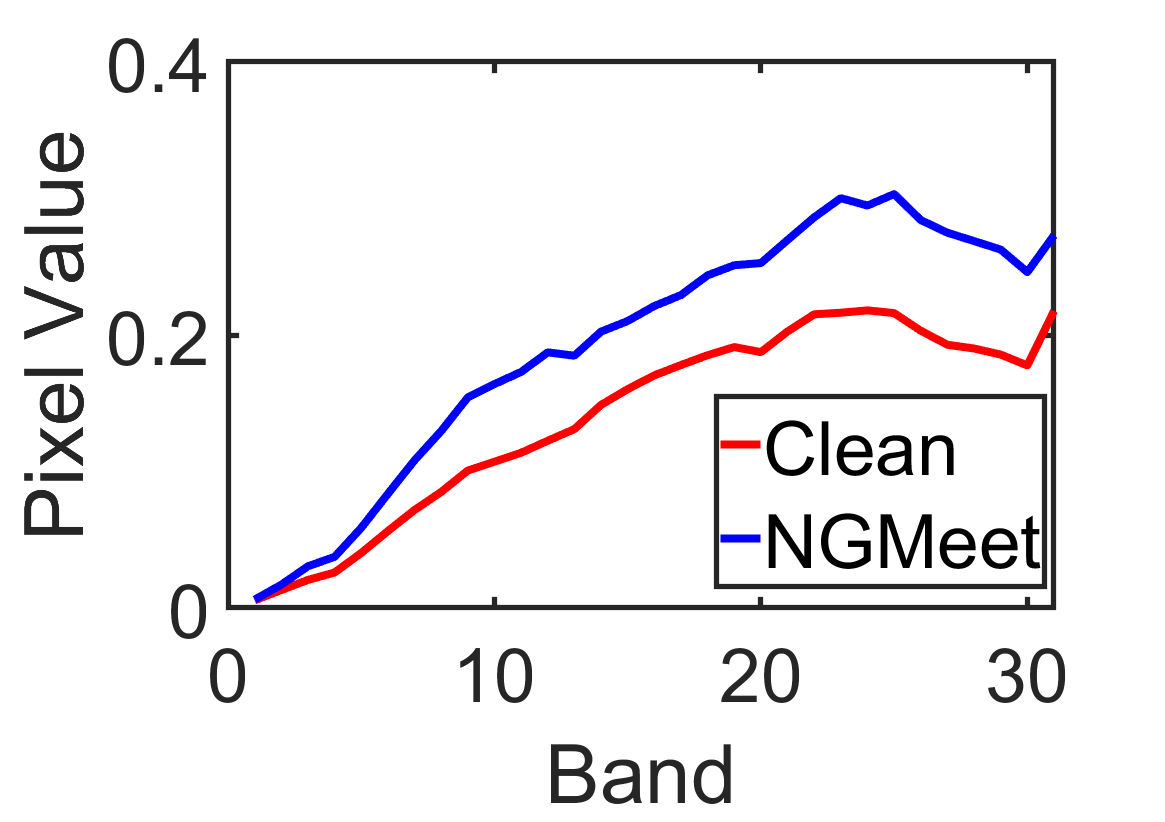}\label{ROC-b22}}
    \subfigure[]{
    \includegraphics[width=0.619 in,height=0.464 in]{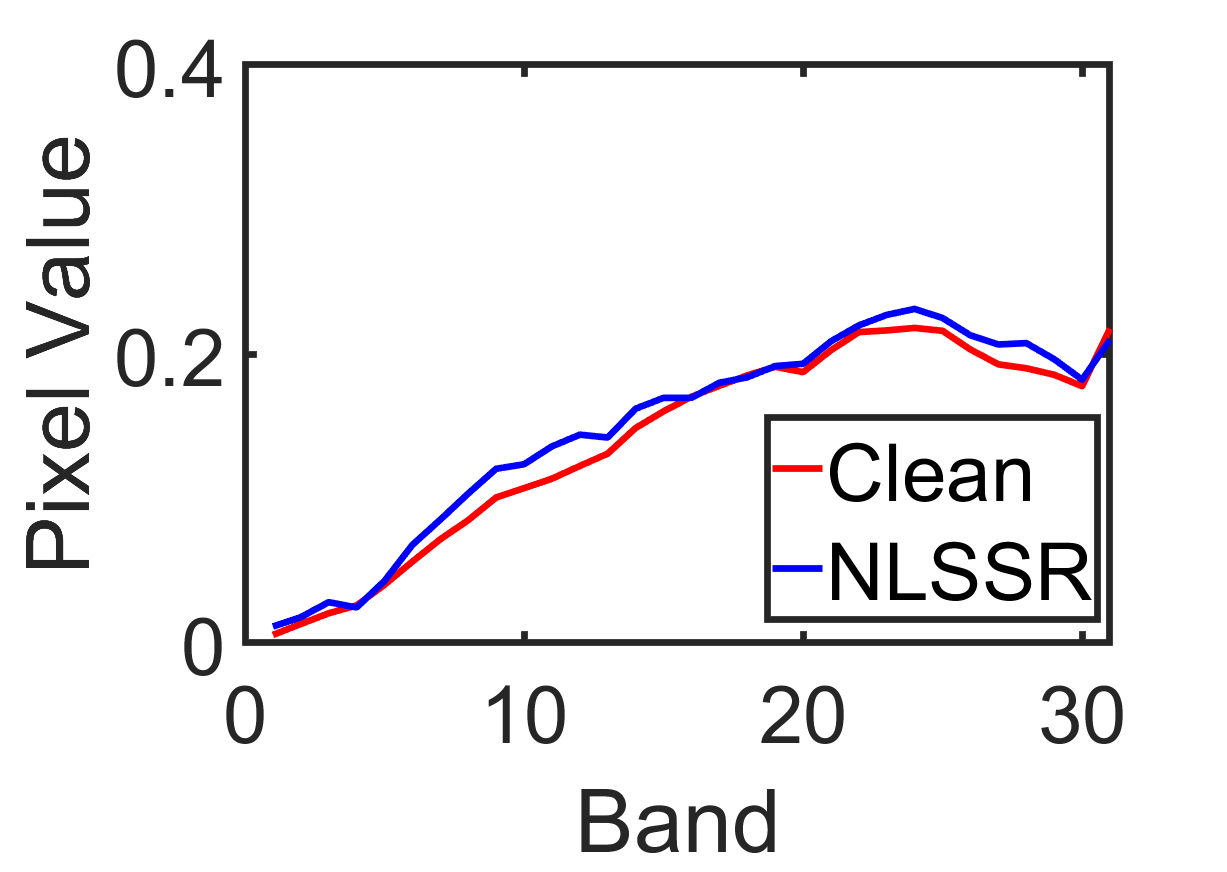}\label{ROC-b22}}
    \subfigure[]{
    \includegraphics[width=0.619 in,height=0.464 in]{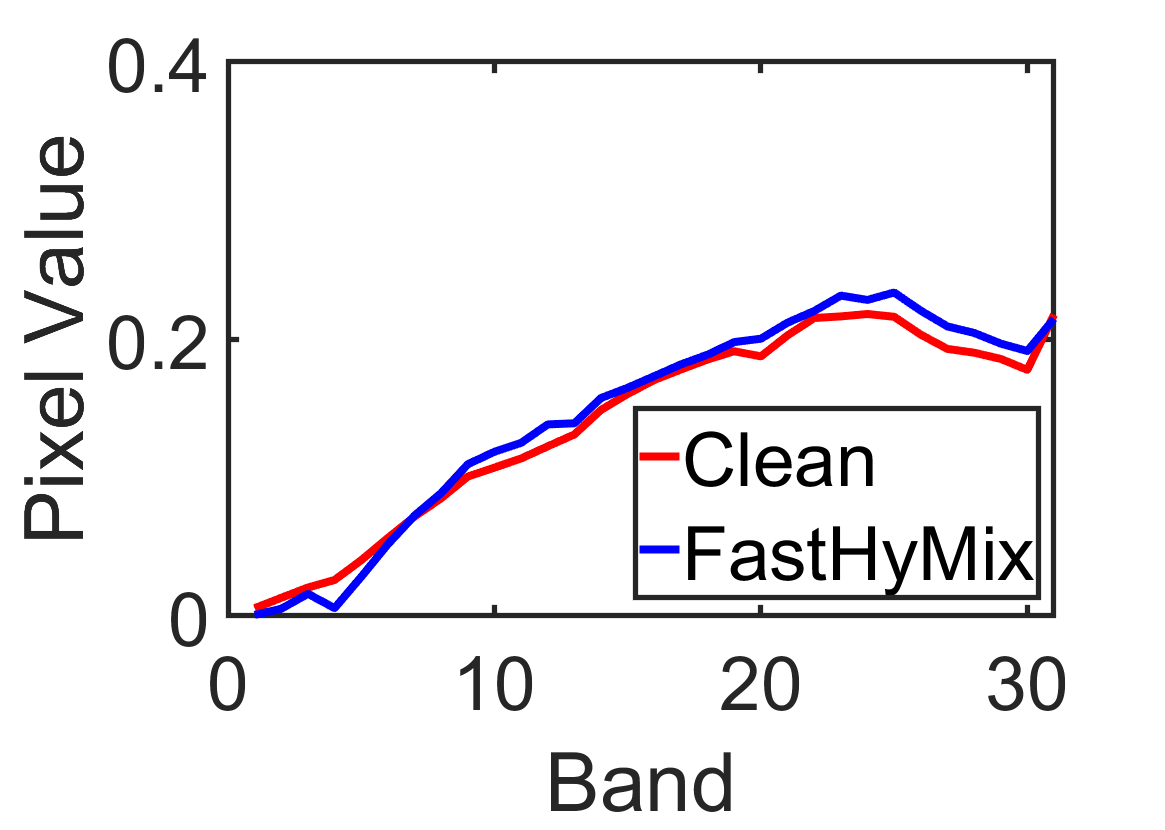}\label{ROC-b22}}
    \subfigure[]{
    \includegraphics[width=0.619 in,height=0.464 in]{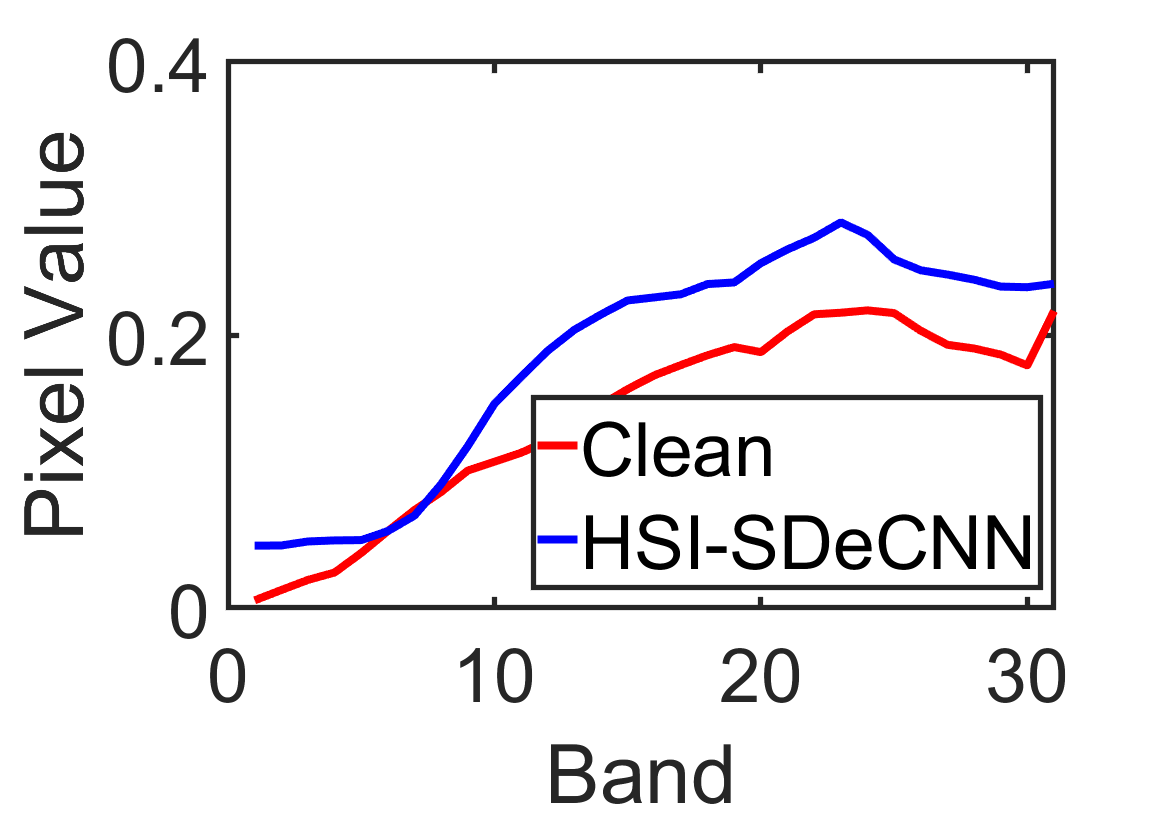}\label{ROC-b23}}
    \subfigure[]{
    \includegraphics[width=0.619 in,height=0.464 in]{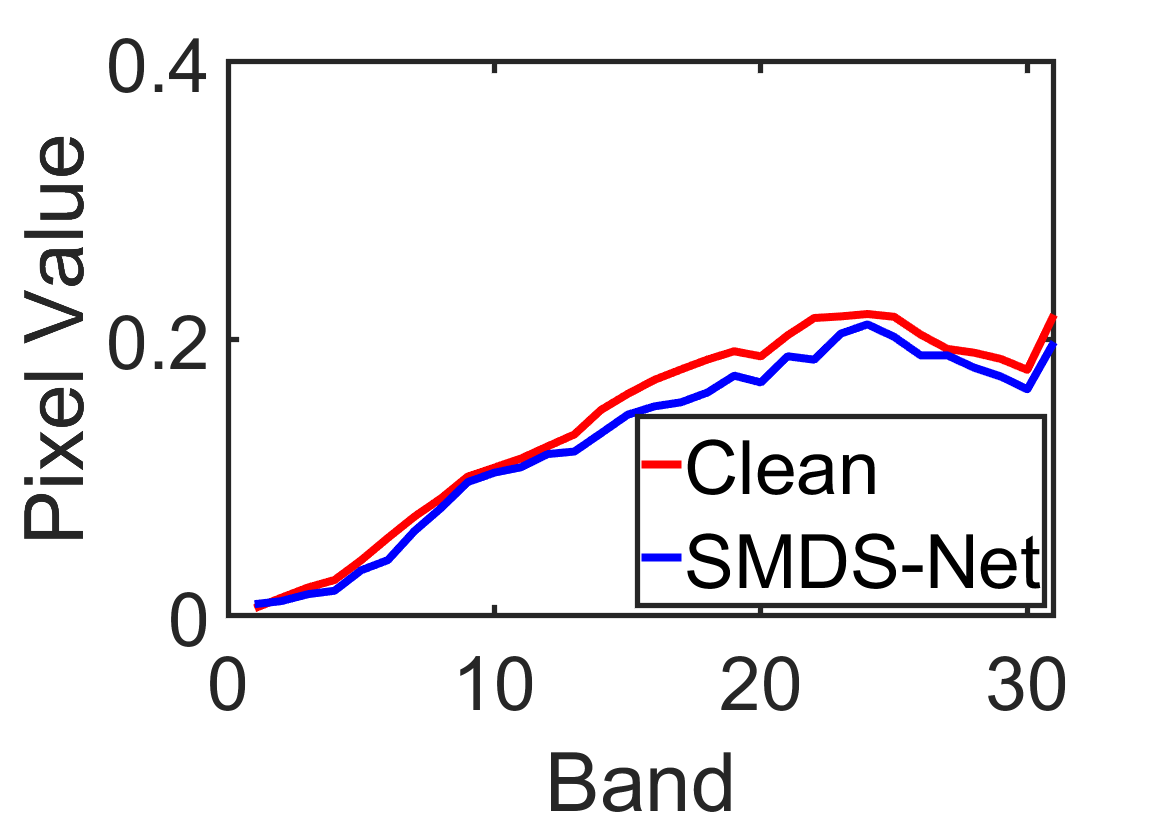}\label{ROC-b23}}
    \subfigure[]{
    \includegraphics[width=0.619 in,height=0.464 in]{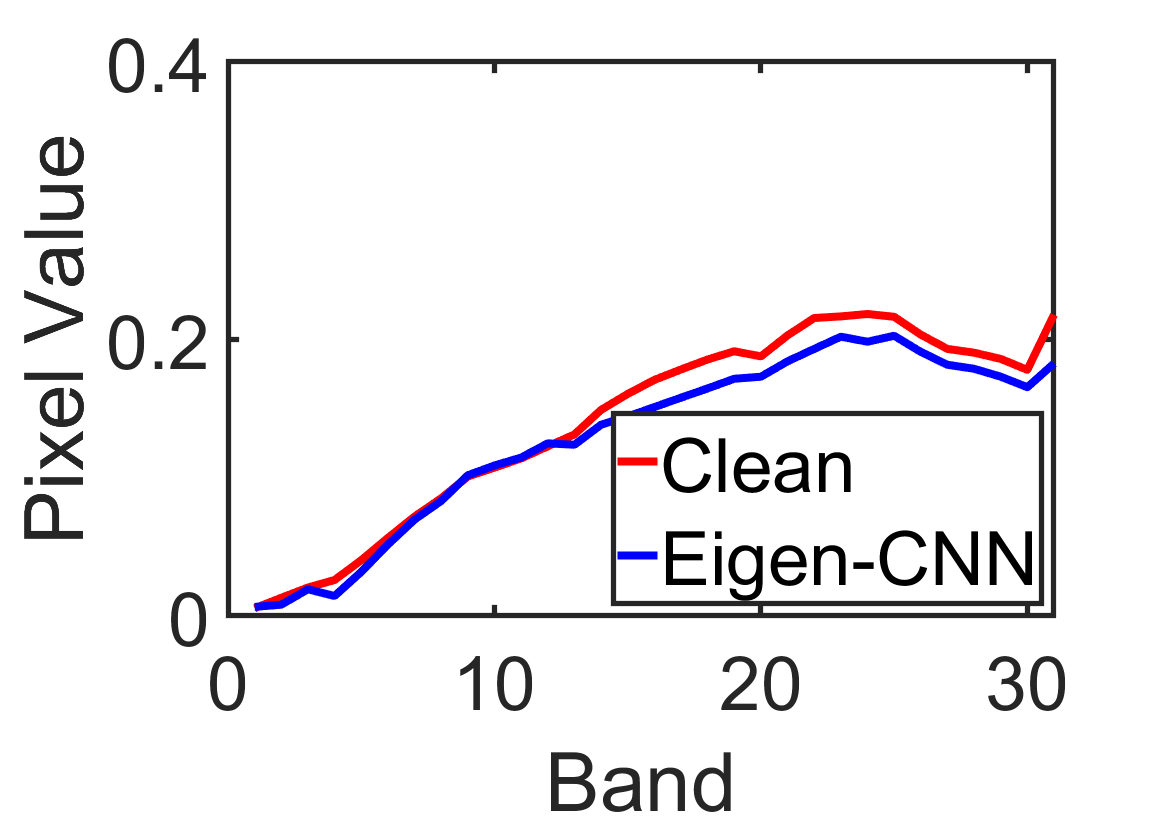}\label{ROC-b22}}
    \subfigure[]{
    \includegraphics[width=0.619 in,height=0.464 in]{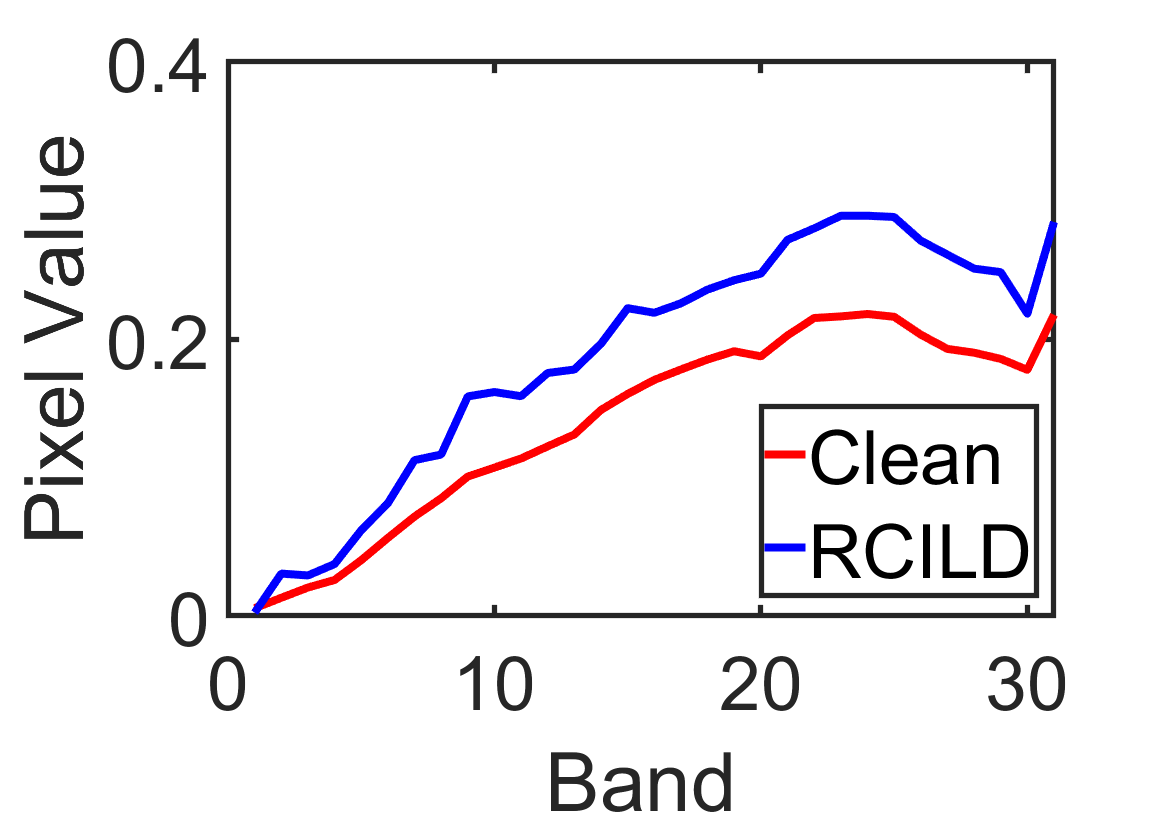}\label{ROC-b23}}
    \subfigure[]{
    \includegraphics[width=0.619 in,height=0.464 in]{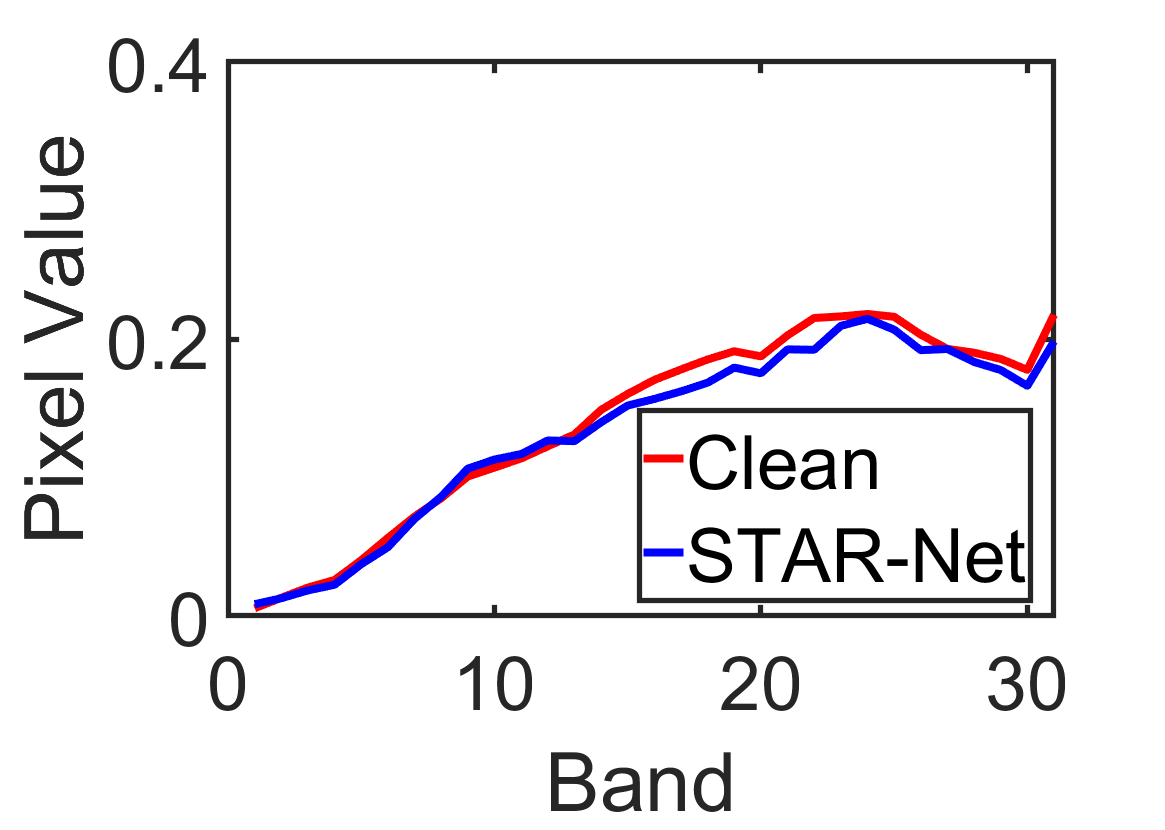}\label{ROC-b24}}
    \subfigure[]{
    \includegraphics[width=0.619 in,height=0.464 in]{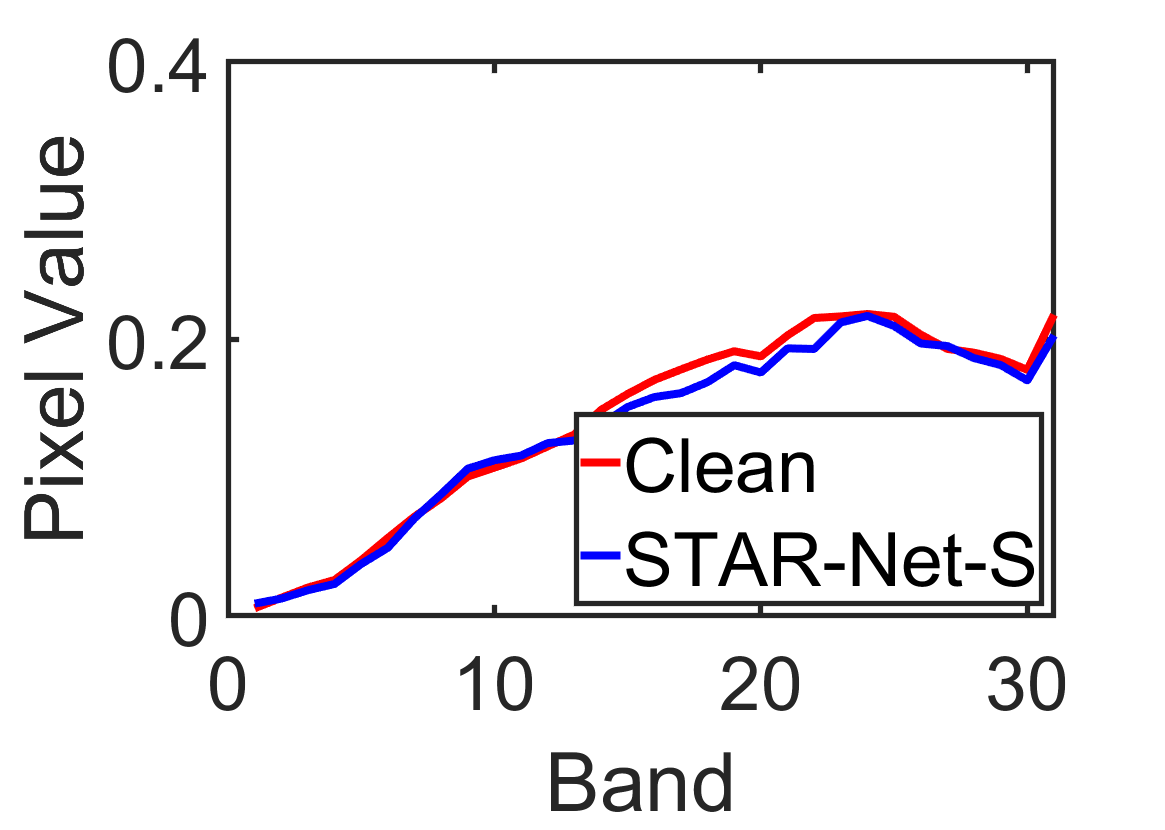}\label{ROC-b24}}
         \vskip-0.2cm
  \caption{Denoising results of pixel (400, 100) on gavyam$\text{\_0823-0933}$ with the noise variance of 50. (a) Clean, (b) Noisy, (c) BM4D, (d) LLRT, (e) LRTDTV, (f) NGMeet, (g) NLSSR, (h) FastHyMix, (i) HSI-SDeCNN, (j) SMDS-Net, (k) Eigen-CNN, (l) RCILD, (m) STAR-Net, (n) STAR-Net-S.}\label{icvlpixel}
\end{figure*}

To better illustrate the denoising effect of each method, the gavyam$\text{\_0823-0933}$ RSI from the ICVL dataset is selected for visualization.
Figure \ref{icvl} shows the clean RSI, the RSI with added noise, and the RSI denoised by all methods with the variance  of 50.
It can be found that LRTDTV and NLSSR perform relatively well, but both exhibit residual noise and do not achieve optimal visual results.
Among deep learning-based methods, except HSI-SDeCNN, all methods are able to filter out noise and approximate the clean RSI.
To further assess the overall denoising performance of all methods, spectral curves are introduced to evaluate spectral recovery. The closer the post-denoising spectral trend is to that of the clean RSI, the better the spectral recovery capability and the stronger the preservation of spectral–spatial structural correlations.
Figure \ref{icvlpixel} shows the spectral reflectance of pixel (400,100) in the gavyam$\text{\_0823-0933}$ RSI before and after denoising.
It can be seen that STAR-Net and STAR-Net-S have stronger spectral recovery capabilities and are close to the spectra before denoising. This further verifies the denoising and spectral recovery capabilities of our proposed methods.

\begin{table*}[t]
\centering
\renewcommand\arraystretch{1.2}
\caption{Comparison of all methods on PaviaU. The top two values are marked as \textcolor[rgb]{1.00,0.00,0.00}{red} and \textcolor[rgb]{0.00,0.00,1.00}{blue}.}\label{PUtable}
\vskip-0.2cm
\setlength{\tabcolsep}{4.5 pt}
\resizebox{\textwidth}{!}{
	\begin{tabular}{ccccccccccccccc}
		\toprule
		{$\sigma$}   & {Index} & {Noisy}  
		                                                  & \makecell[c]{BM4D\\}   & \makecell[c]{LLRT\\}                       & \makecell[c]{LRTDTV\\} & \makecell[c]{NGMeet\\}& \makecell[c]{NLSSR\\}&\makecell[c]{FastHy\\Mix} & \makecell[c]{HSI-SDe\\CNN}  & \makecell[c]{SMDS-\\Net}    &\makecell[c]{Eigen-\\CNN}   & \makecell[c]{RCILD\\}         & \makecell[c]{STAR-\\Net}        & \makecell[c]{STAR-\\Net-S}   \\ \hline 
		\multirow{4}{*}{10}      & PSNR $\uparrow$    & 29.181   & 34.044  & 30.671  & 31.796  & 36.493  & 35.419        & 39.646          & 38.509     & 35.882      & 39.660      & 40.509       & \textcolor[rgb]{0.00,0.00,1.00}{40.881}   & \textcolor[rgb]{1.00,0.00,0.00}{41.172}  \\
                         & SSIM $\uparrow$    & 0.654    & 0.902   & 0.821   & 0.856   & 0.933   & 0.925         & \textcolor[rgb]{0.00,0.00,1.00}{0.968}     & 0.953      & 0.938       & 0.969       & 0.962        & \textcolor[rgb]{1.00,0.00,0.00}{0.971} & \textcolor[rgb]{1.00,0.00,0.00}{0.971}   \\
                         & SAM $\downarrow$   & 0.221    & 0.108   & 0.159   & 0.141   & 0.084   & 0.092         & 0.060           & 0.067      & 0.068       & 0.060       & \textcolor[rgb]{0.00,0.00,1.00}{ 0.056}  & \textcolor[rgb]{1.00,0.00,0.00}{0.049} & \textcolor[rgb]{1.00,0.00,0.00}{0.049}   \\
                         & ERGAS $\downarrow$ & 157.975  & 77.328  & 116.590 & 102.736 & 60.251  & 67.076        & 44.100          & 48.351     & 67.165      & 44.016      & 44.690       & \textcolor[rgb]{0.00,0.00,1.00}{ 40.944}   & \textcolor[rgb]{1.00,0.00,0.00}{39.927}  \\ \hline
\multirow{4}{*}{30}      & PSNR $\uparrow$    & 21.409   & 28.398  & 30.394  & 31.756  & 30.448  & 33.835        & 32.960          & 32.236     & 30.010      & 33.028      & \textcolor[rgb]{0.00,0.00,1.00}{ 35.947} & 35.637         & \textcolor[rgb]{1.00,0.00,0.00}{35.976}  \\
                         & SSIM $\uparrow$    & 0.237    & 0.746   & 0.809   & 0.855   & 0.818   & 0.901         & 0.919           & 0.849      & 0.929       & 0.922       & 0.899        & \textcolor[rgb]{0.00,0.00,1.00}{ 0.935}    & \textcolor[rgb]{1.00,0.00,0.00}{0.936}   \\
                         & SAM $\downarrow$   & 0.580    & 0.202   & 0.165   & 0.141   & 0.610   & 0.111         & 0.125           & 0.134      & 0.077       & 0.124       & 0.090        & \textcolor[rgb]{0.00,0.00,1.00}{ 0.074}    & \textcolor[rgb]{1.00,0.00,0.00}{0.073}   \\
                         & ERGAS $\downarrow$ & 473.723  & 147.139 & 120.494 & 103.199 & 119.247 & 80.904        & 96.429          & 98.539     & 73.304      & 95.860      & 69.178       & \textcolor[rgb]{0.00,0.00,1.00}{
                          68.764}   & \textcolor[rgb]{1.00,0.00,0.00}{66.131}  \\ \hline
\multirow{4}{*}{50}      & PSNR $\uparrow$    & 18.760   & 26.159  & 27.022  & 31.648  & 27.756  & 31.750        & 31.909          & 29.198     & 31.748      & 32.031      & 31.416       & \textcolor[rgb]{0.00,0.00,1.00}{ 33.227}   & \textcolor[rgb]{1.00,0.00,0.00}{33.243}  \\
                         & SSIM $\uparrow$    & 0.114   & 0.650   & 0.666   & 0.850   & 0.722   & 0.856         & 0.897           & 0.757      & 0.878       & \textcolor[rgb]{0.00,0.00,1.00}{ 0.899} & 0.831        & 0.898          & \textcolor[rgb]{1.00,0.00,0.00}{0.902}   \\
                         & SAM $\downarrow$   & 0.816    & 0.259   & 0.242   & 0.196   & 0.220   & 0.140         & 0.138           & 0.182      & 0.098       & 0.136       & 0.150        & \textcolor[rgb]{0.00,0.00,1.00}{ 0.093}    & \textcolor[rgb]{1.00,0.00,0.00}{0.092}   \\
                         & ERGAS $\downarrow$ & 789.831  & 188.949 & 177.238 & 104.514 & 162.735 & 103.138       & 105.438         & 138.135    & 104.315     & 104.304     & 115.949      & \textcolor[rgb]{0.00,0.00,1.00}{ 89.223}   & \textcolor[rgb]{1.00,0.00,0.00}{88.743}  \\ \hline
\multirow{4}{*}{70}      & PSNR $\uparrow$    & 16.705   & 24.940  & 26.626  & 30.644  & 26.281  & 30.180        & 31.087          & 27.435     & 30.989      & 31.097      & 30.560       & \textcolor[rgb]{0.00,0.00,1.00}{ 31.658}   & \textcolor[rgb]{1.00,0.00,0.00}{31.694}  \\
                         & SSIM $\uparrow$    & 0.064    & 0.595   & 0.643   & 0.820   & 0.659   & 0.814         & \textcolor[rgb]{1.00,0.00,0.00}{0.876}  & 0.694      & 0.859       & \textcolor[rgb]{0.00,0.00,1.00}{ 0.872} & 0.779        & 0.868          & 0.868            \\
                         & SAM $\downarrow$   & 0.971    & 0.298   & 0.254   & 0.160   & 0.265   & 0.166         & 0.150           & 0.215      & \textcolor[rgb]{0.00,0.00,1.00}{ 0.113} & 0.150       & 0.157        & \textcolor[rgb]{1.00,0.00,0.00}{0.105} & \textcolor[rgb]{1.00,0.00,0.00}{0.105}   \\
                         & ERGAS $\downarrow$ & 1105.483 & 216.652 & 185.492 & 117.070 & 193.479 & 122.756       & 114.277         & 169.228    & 113.503     & 114.060     & 126.980      & \textcolor[rgb]{0.00,0.00,1.00}{ 106.249}  & \textcolor[rgb]{1.00,0.00,0.00}{105.060} \\ \hline
\multirow{4}{*}{Average} & PSNR $\uparrow$    & 21.514   & 28.385  & 28.678  & 31.461  & 30.245  & 32.796        & 33.901          & 31.844     & 32.157      & 33.954      & 34.608       & \textcolor[rgb]{0.00,0.00,1.00}{  35.351}   & \textcolor[rgb]{1.00,0.00,0.00}{35.521}  \\
                         & SSIM $\uparrow$    & 0.267    & 0.723   & 0.735   & 0.845   & 0.783   & 0.874         & 0.915           & 0.813      & 0.901       & 0.915       & 0.868        & \textcolor[rgb]{0.00,0.00,1.00}{  0.918}    & \textcolor[rgb]{1.00,0.00,0.00}{0.919}   \\
                         & SAM $\downarrow$   & 0.647    & 0.217   & 0.205   & 0.159   & 0.295   & 0.127         & 0.118           & 0.149      & \textcolor[rgb]{0.00,0.00,1.00}{  0.089} & 0.118       & 0.113        & \textcolor[rgb]{1.00,0.00,0.00}{0.080} & \textcolor[rgb]{1.00,0.00,0.00}{0.080}   \\
                         & ERGAS $\downarrow$ & 631.753  & 157.517 & 149.953 & 106.880 & 133.928 & 93.468        & 90.061          & 113.563    & 89.572      & 89.560      & 89.199       & \textcolor[rgb]{0.00,0.00,1.00}{  76.295}   & \textcolor[rgb]{1.00,0.00,0.00}{74.965}  \\ \bottomrule
	\end{tabular}}
\end{table*}

\begin{figure*}[h!]
  \centering
    \subfigure[PSNR(dB)]{
    \includegraphics[width=0.619 in,height=1.111 in]{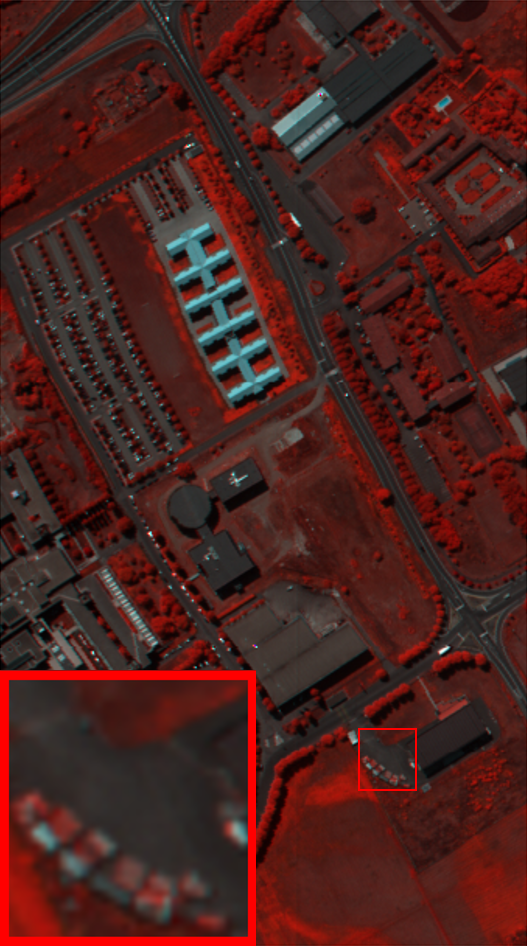}\label{ROC-b21}}
    \subfigure[18.760]{
    \includegraphics[width=0.619 in,height=1.111 in]{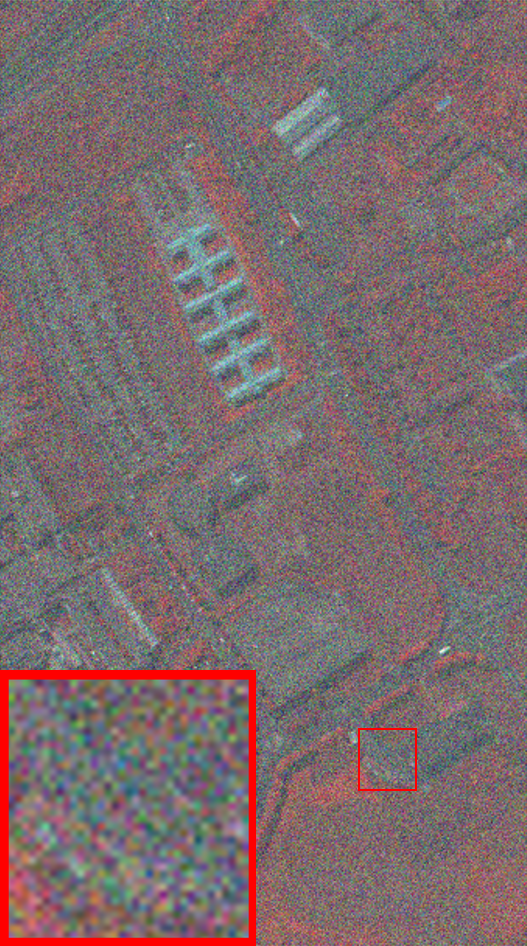}\label{ROC-b22}}
    \subfigure[26.159]{
    \includegraphics[width=0.619 in,height=1.111 in]{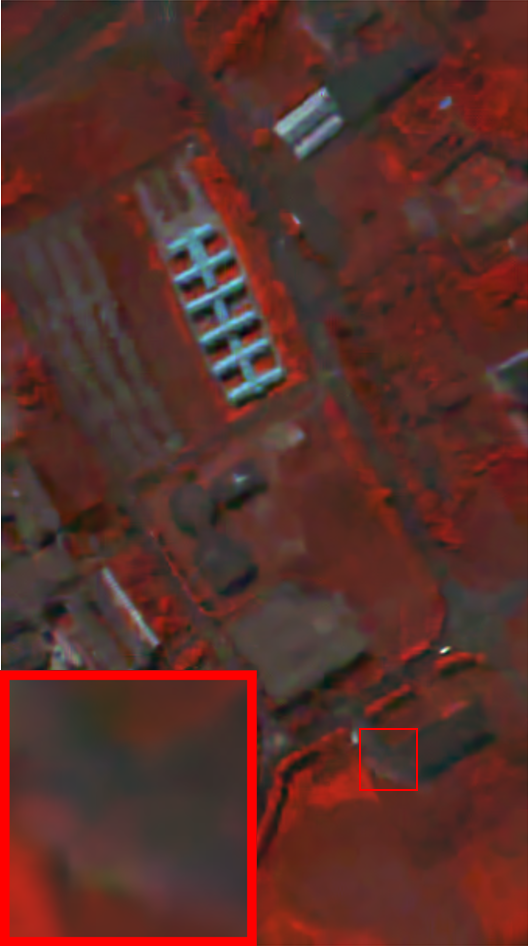}\label{ROC-b22}}
    \subfigure[27.022]{
    \includegraphics[width=0.619 in,height=1.111 in]{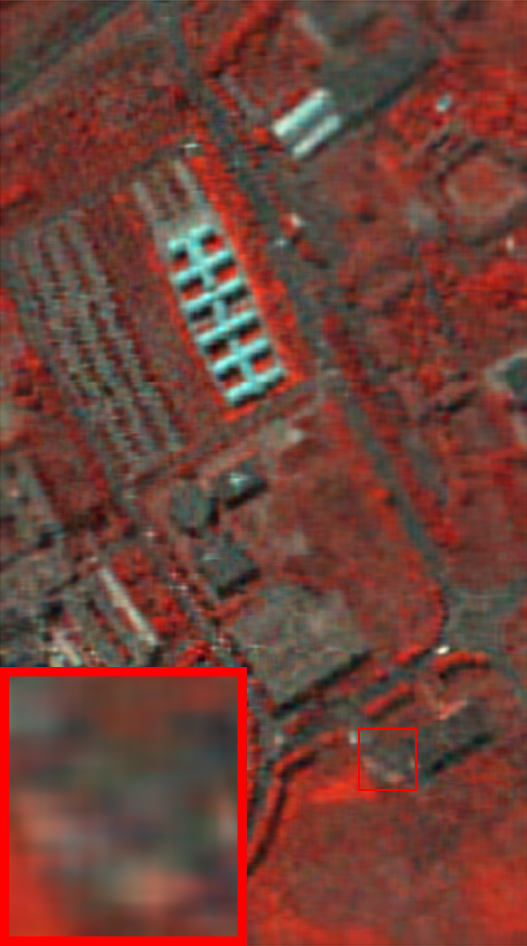}\label{ROC-b22}}
    \subfigure[31.648]{
    \includegraphics[width=0.619 in,height=1.111 in]{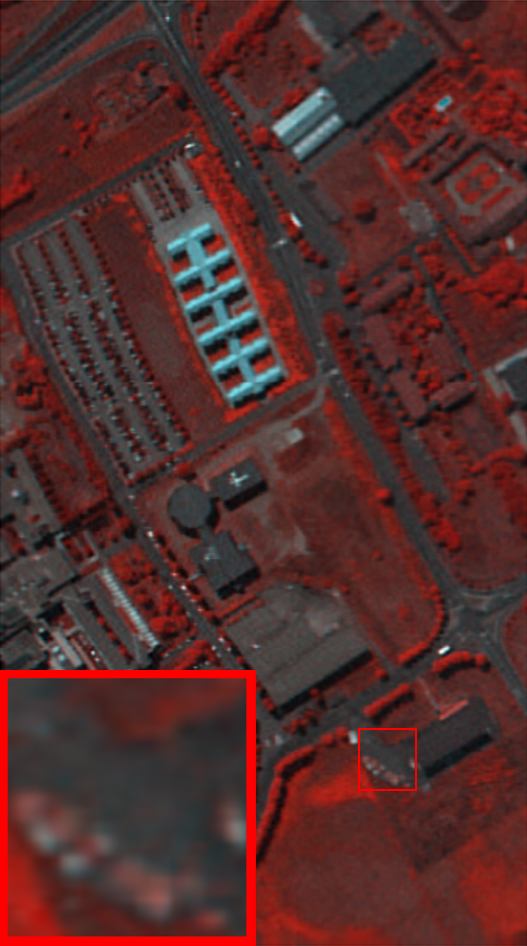}\label{ROC-b22}}
    \subfigure[27.756]{
    \includegraphics[width=0.619 in,height=1.111 in]{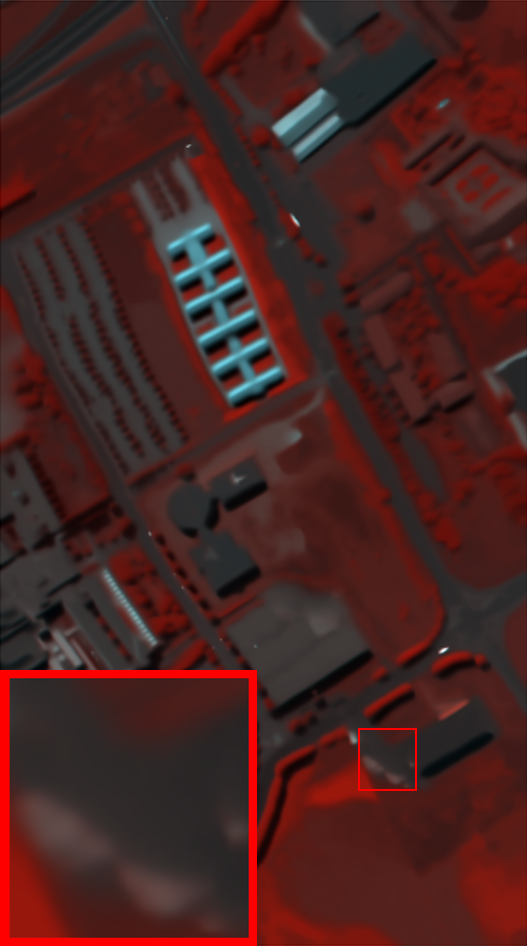}\label{ROC-b22}}
    \subfigure[31.750]{
    \includegraphics[width=0.619 in,height=1.111 in]{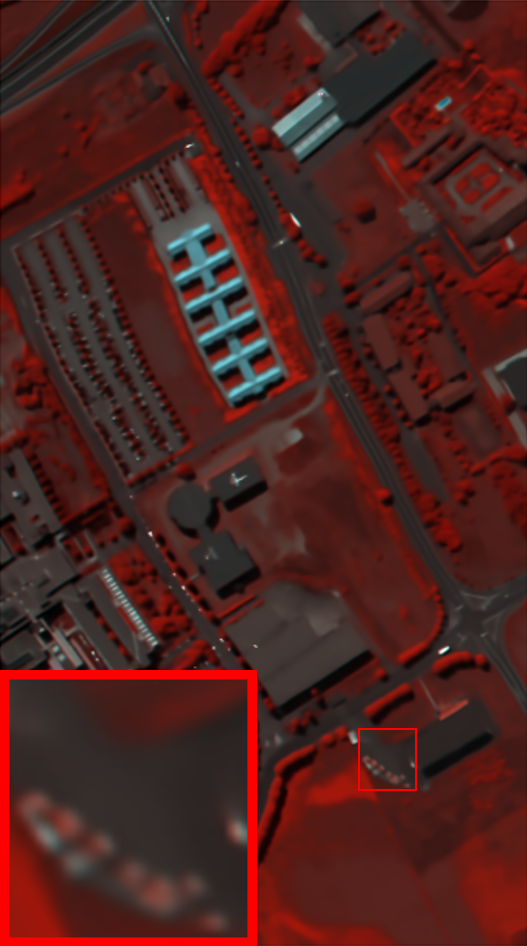}\label{ROC-b22}}
    \subfigure[31.909]{
    \includegraphics[width=0.619 in,height=1.111 in]{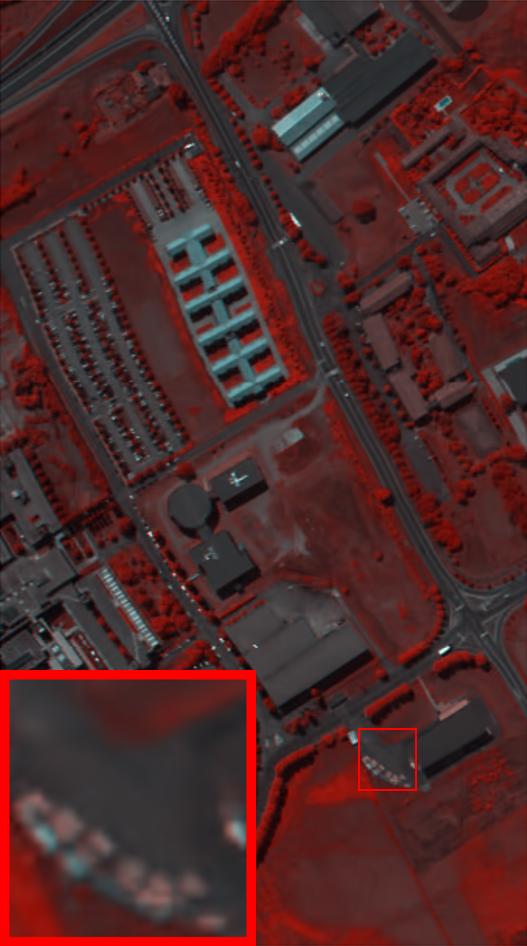}\label{ROC-b22}}
    \subfigure[29.198]{
    \includegraphics[width=0.619 in,height=1.111 in]{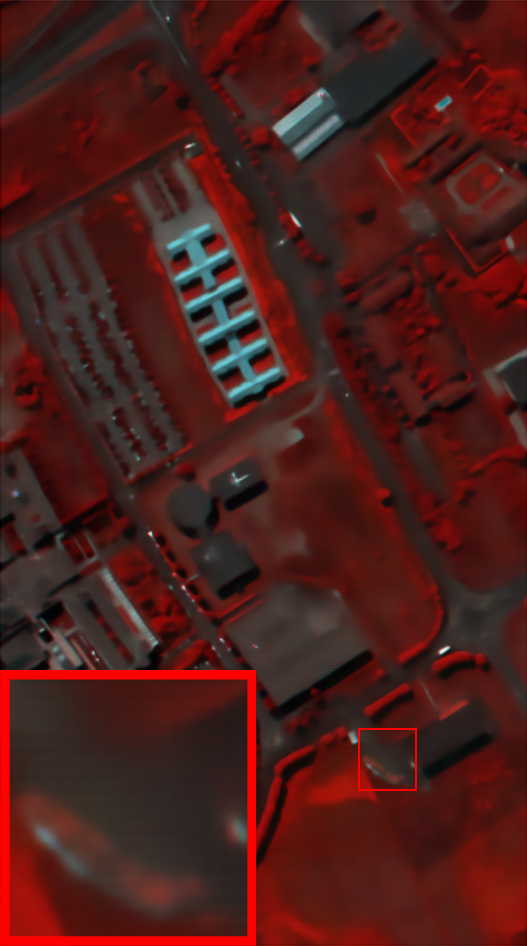}\label{ROC-b22}}
    \subfigure[31.748]{
    \includegraphics[width=0.619 in,height=1.111 in]{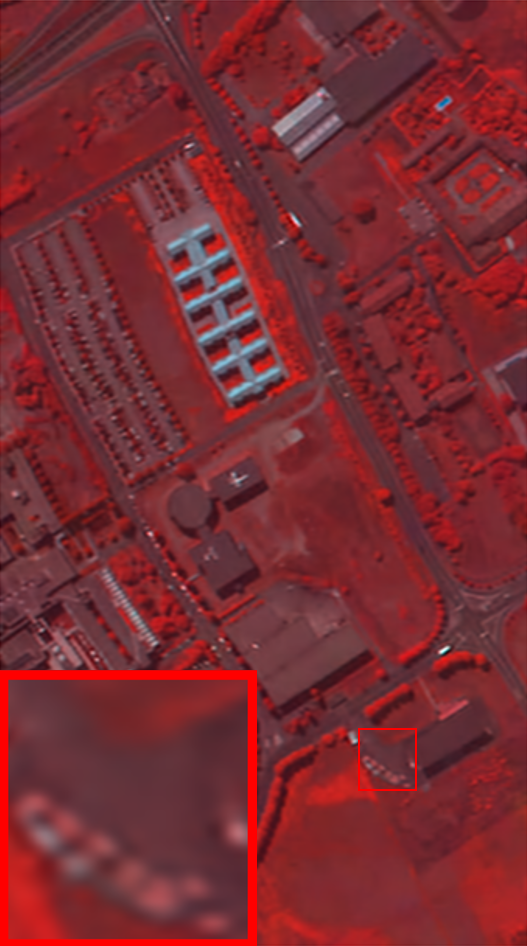}\label{ROC-b23}}
    \subfigure[32.031]{
    \includegraphics[width=0.619 in,height=1.111 in]{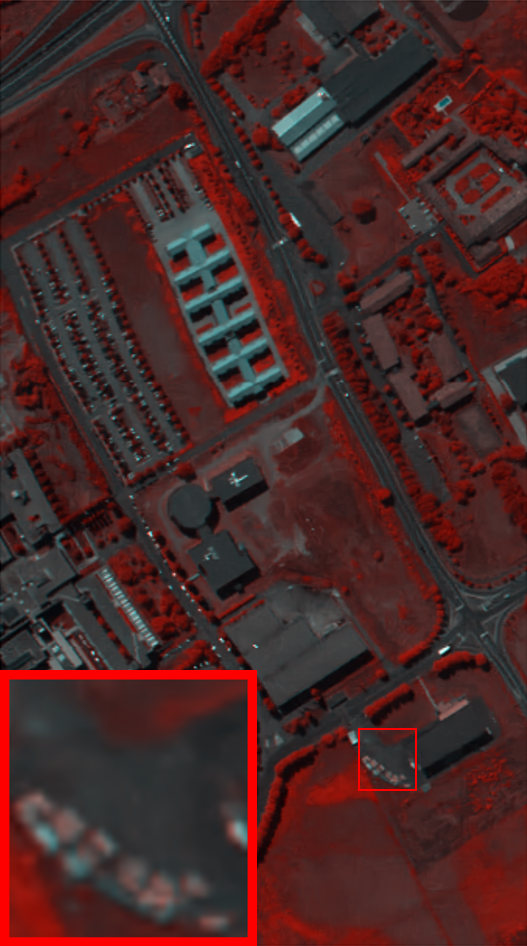}\label{ROC-b23}}
    \subfigure[31.416]{
    \includegraphics[width=0.619 in,height=1.111 in]{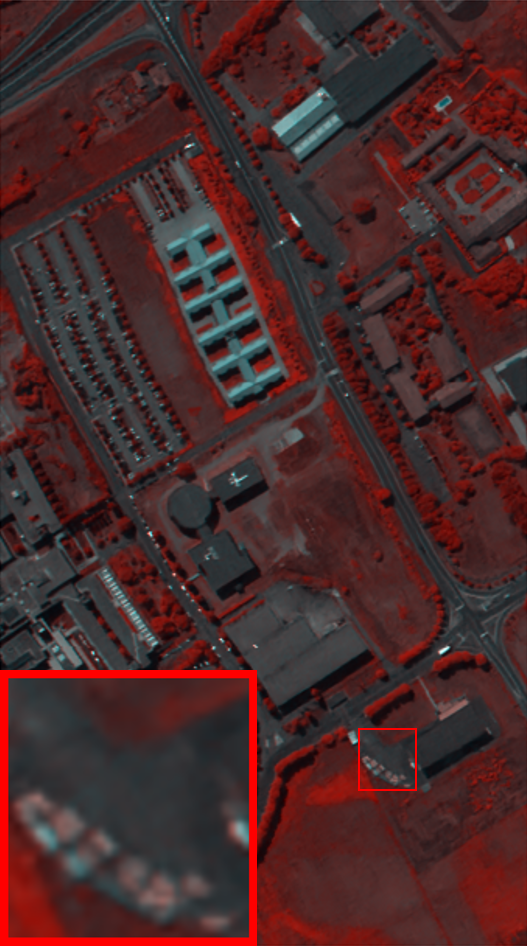}\label{ROC-b22}}
    \subfigure[33.227]{
    \includegraphics[width=0.619 in,height=1.111 in]{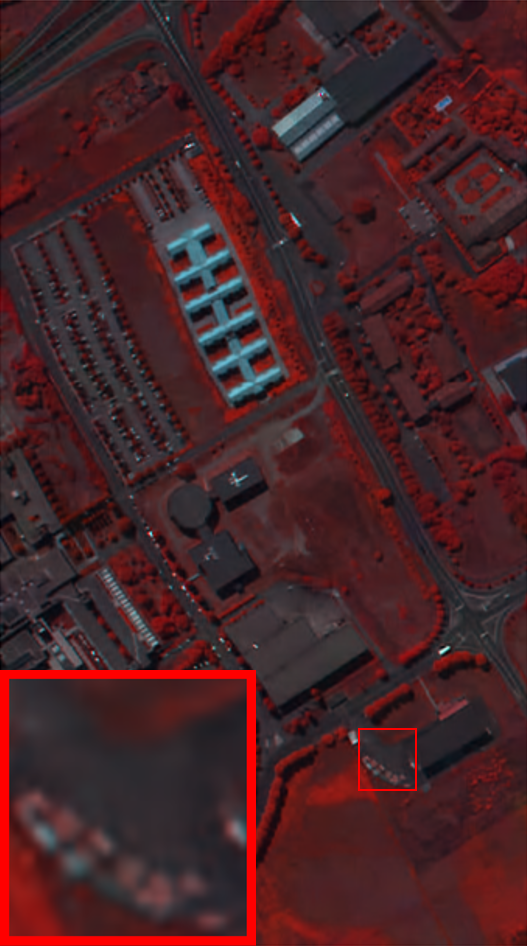}\label{ROC-b24}}
    \subfigure[33.243]{
    \includegraphics[width=0.619 in,height=1.111 in]{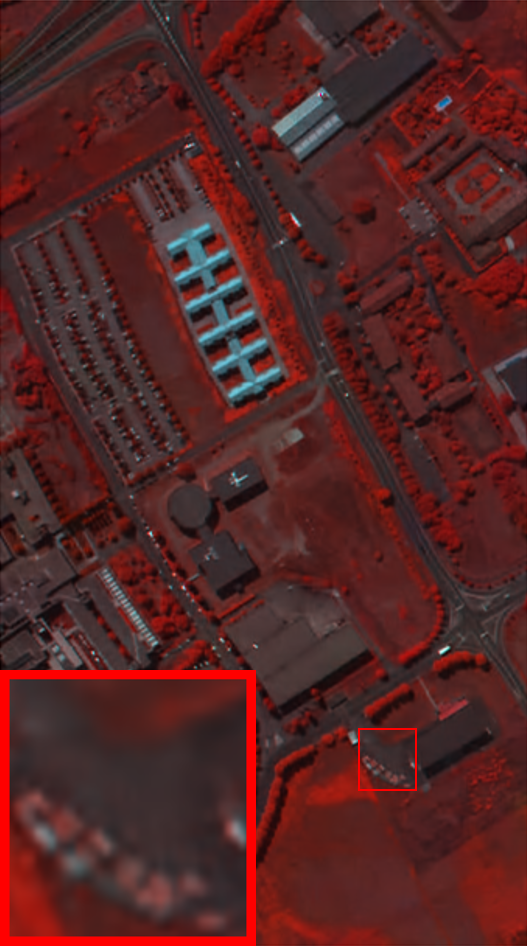}\label{ROC-b24}}
         \vskip-0.2cm
  \caption{Denoising results on PaviaU with  the noise variance of 50. The false-color images are generated by combining bands 22, 28, and 88. (a) Clean, (b) Noisy, (c) BM4D, (d) LLRT, (e) LRTDTV, (f) NGMeet, (g) NLSSR, (h) FastHyMix, (i) HSI-SDeCNN, (j) SMDS-Net, (k) Eigen-CNN, (l) RCILD, (m) STAR-Net, (n) STAR-Net-S.}\label{PU}
\end{figure*}

\begin{figure*}[t]
  \centering
    \subfigure[]{
    \includegraphics[width=0.619 in,height=0.464 in]{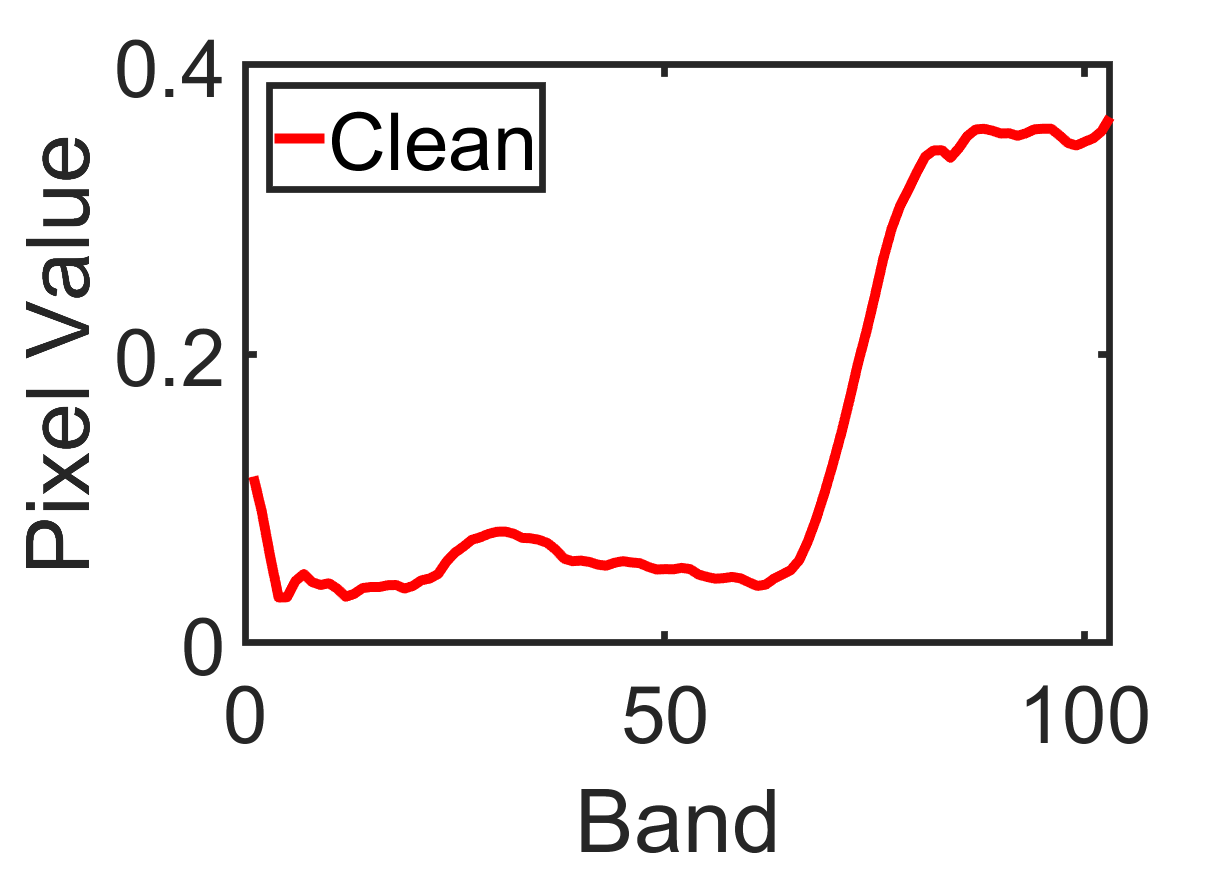}\label{PUb21}}
    \subfigure[]{
    \includegraphics[width=0.619 in,height=0.464 in]{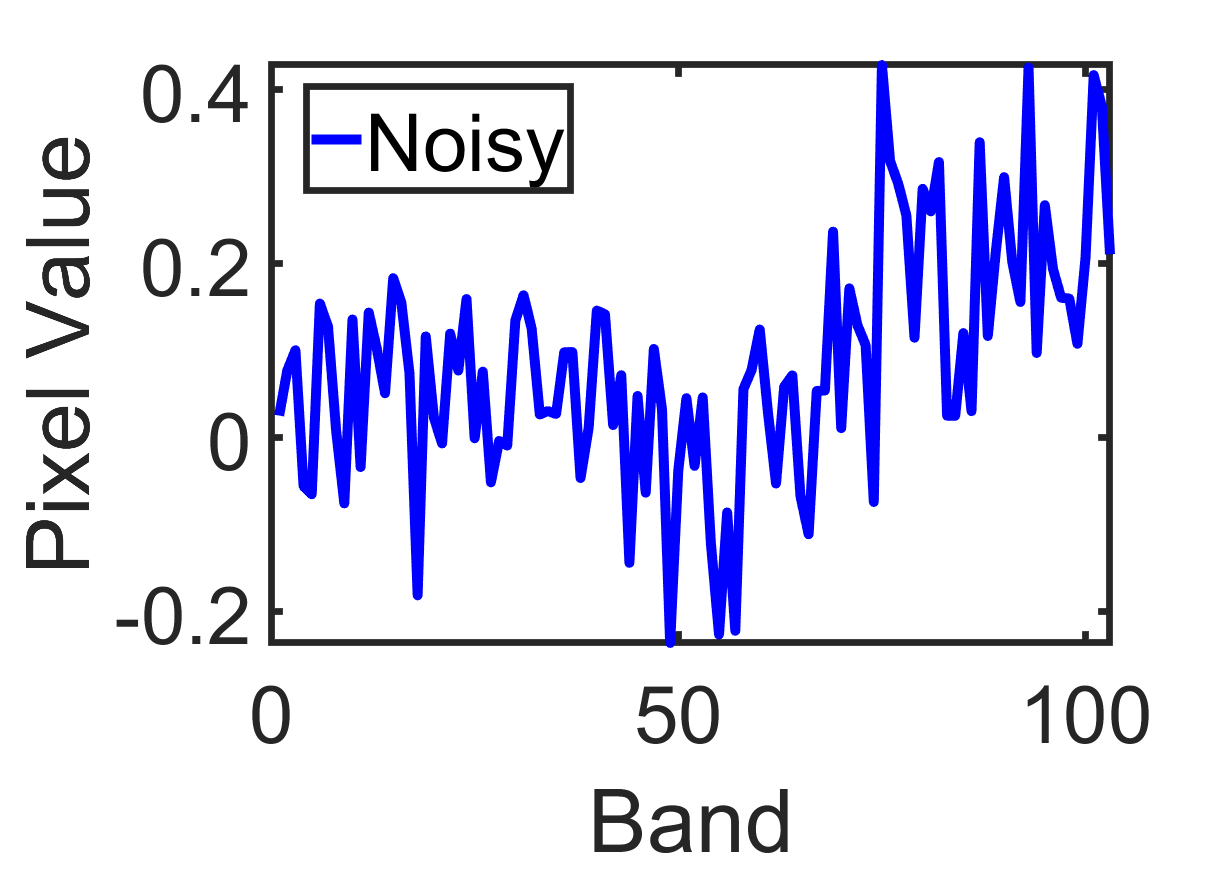}\label{PUb22}}
    \subfigure[]{
    \includegraphics[width=0.619 in,height=0.464 in]{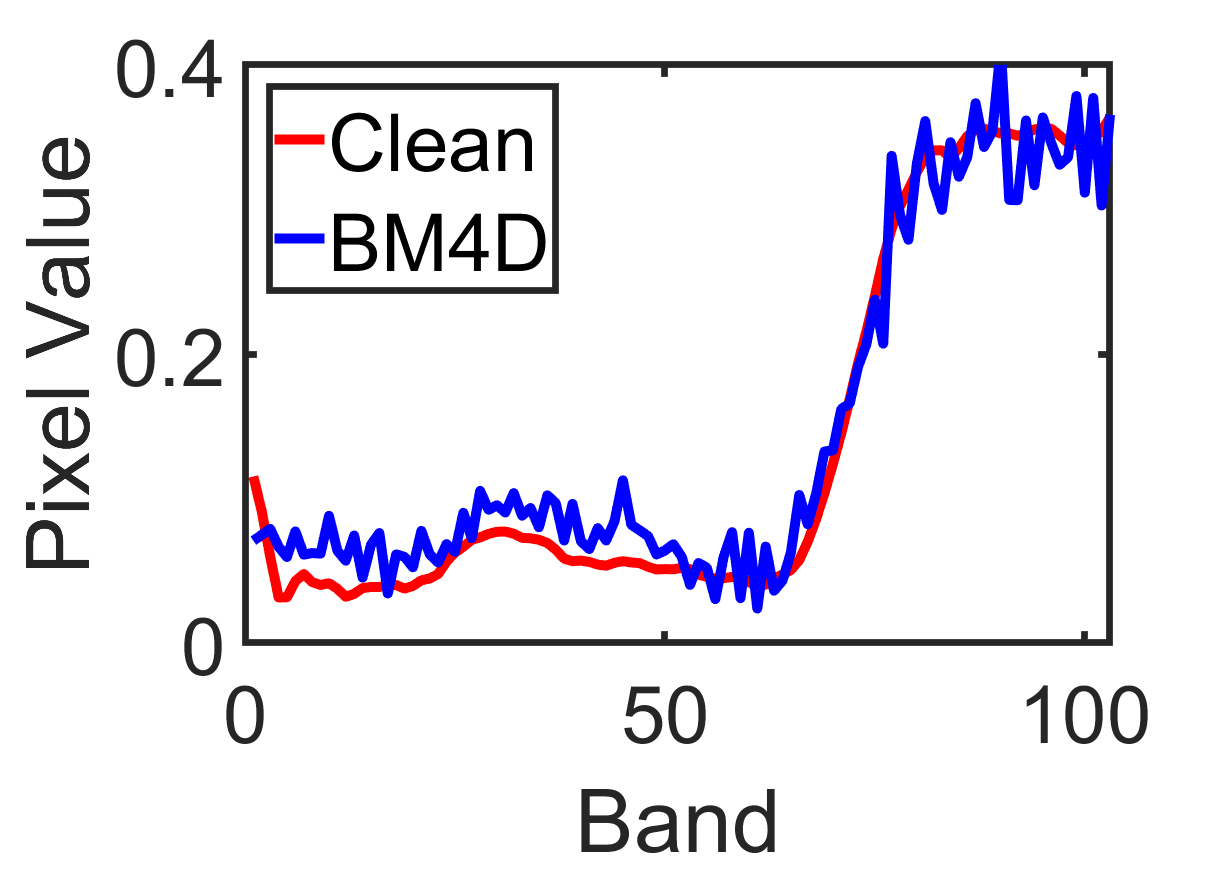}\label{ROC-b22}}
    \subfigure[]{
    \includegraphics[width=0.619 in,height=0.464 in]{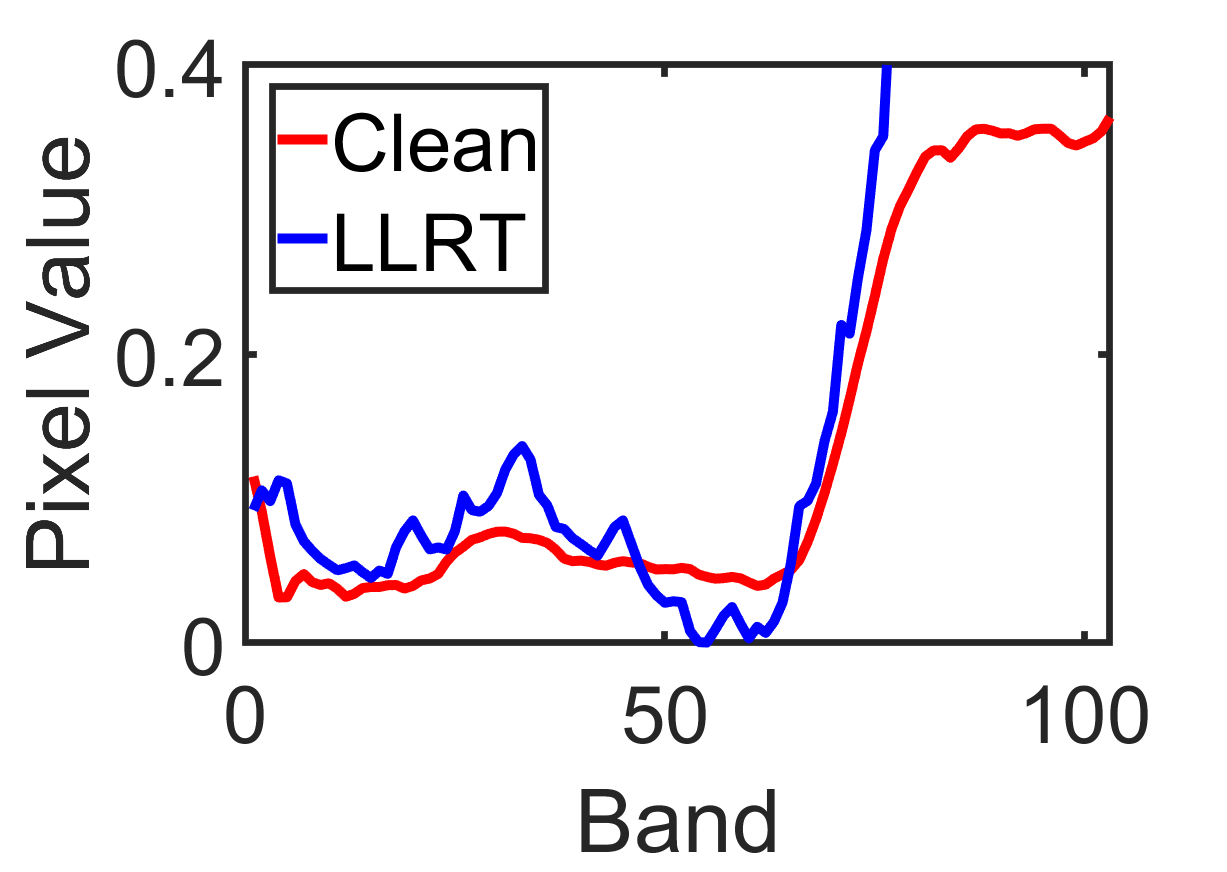}\label{ROC-b22}}
    \subfigure[]{
    \includegraphics[width=0.619 in,height=0.464 in]{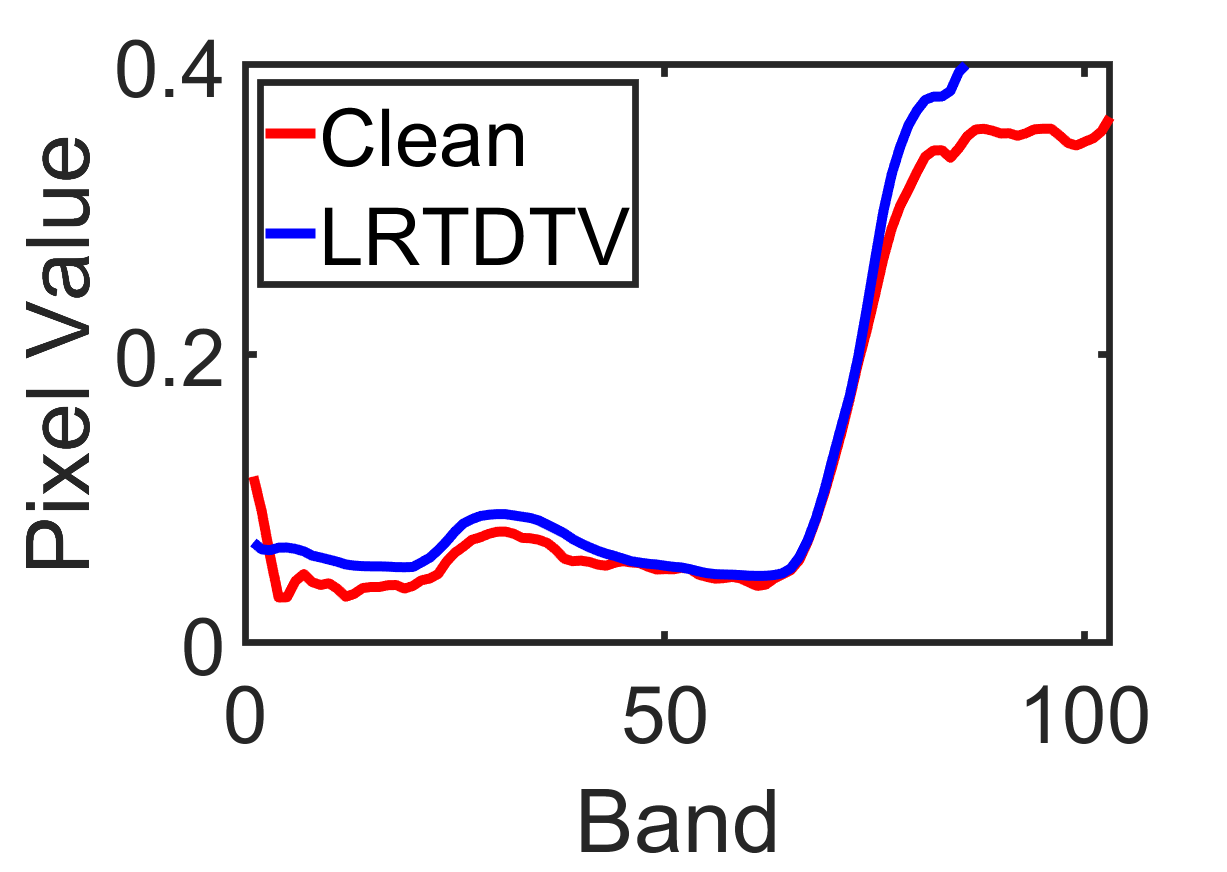}\label{ROC-b23}}
    \subfigure[]{
    \includegraphics[width=0.619 in,height=0.464 in]{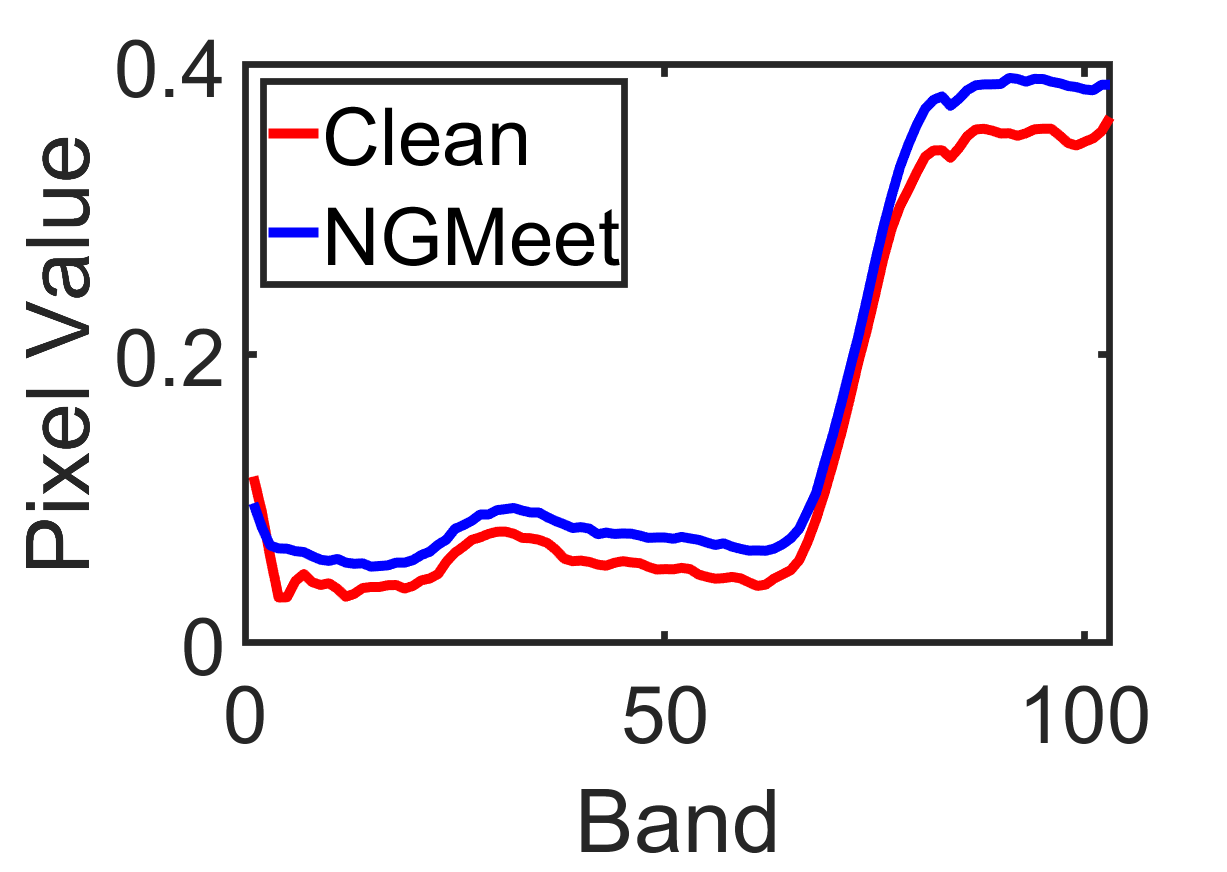}\label{ROC-b22}}
    \subfigure[]{
    \includegraphics[width=0.619 in,height=0.464 in]{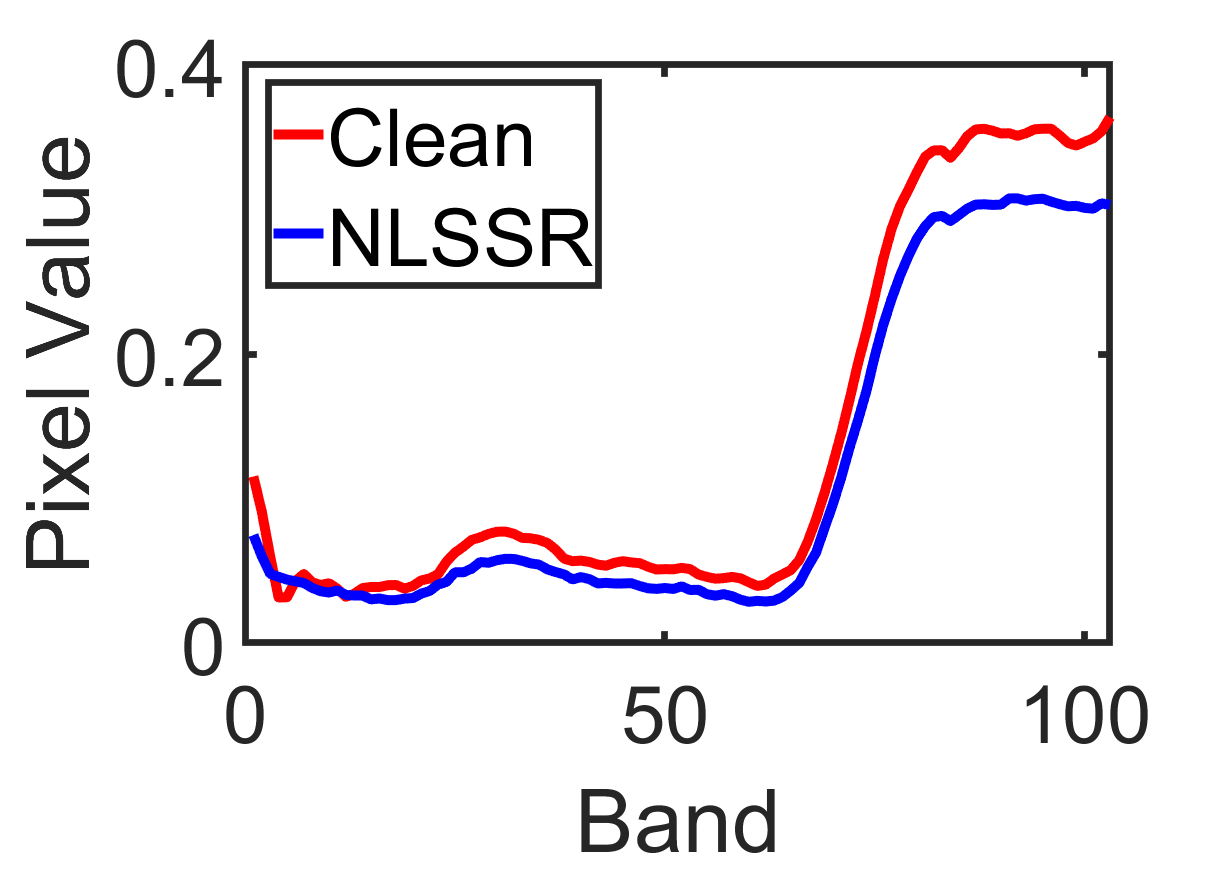}\label{ROC-b22}}
    \subfigure[]{
    \includegraphics[width=0.619 in,height=0.464 in]{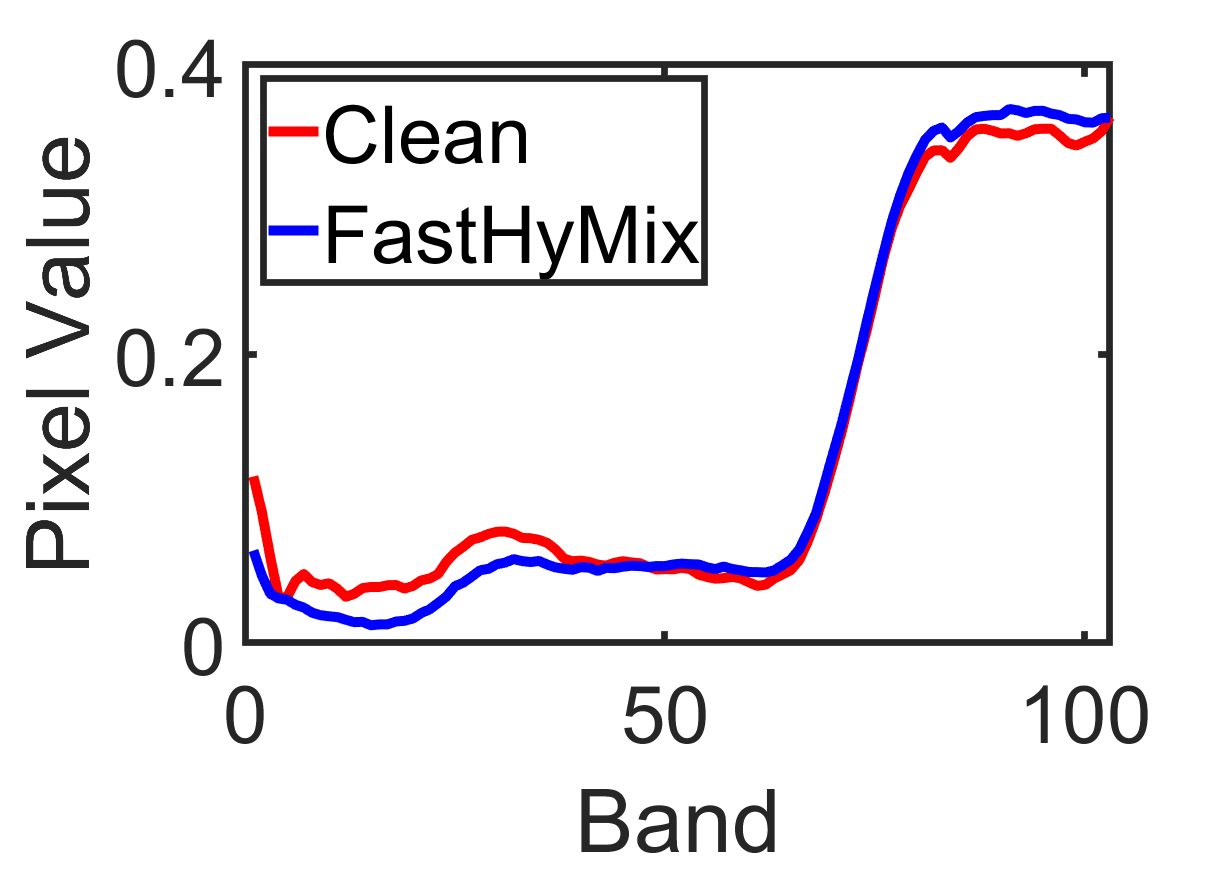}\label{ROC-b22}}
    \subfigure[]{
    \includegraphics[width=0.619 in,height=0.464 in]{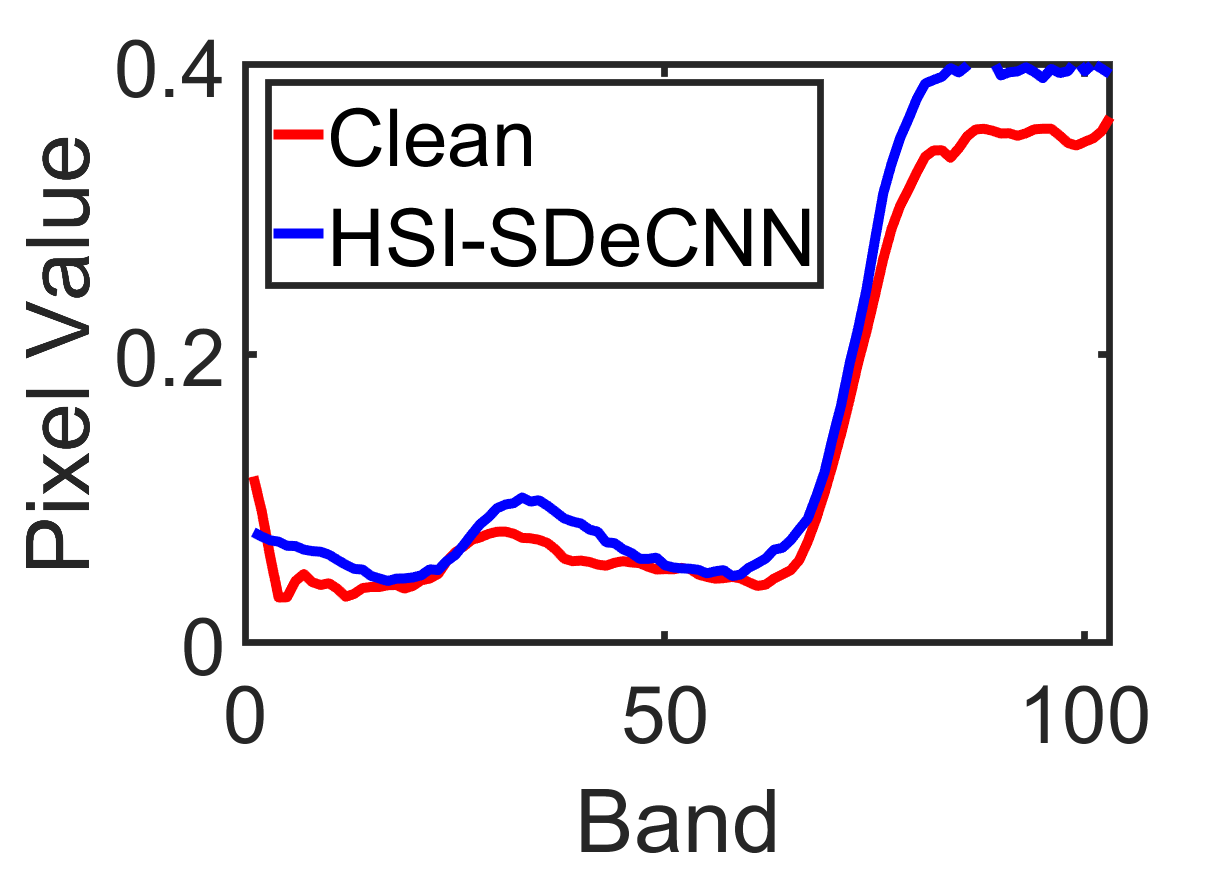}\label{ROC-b23}}
    \subfigure[]{
    \includegraphics[width=0.619 in,height=0.464 in]{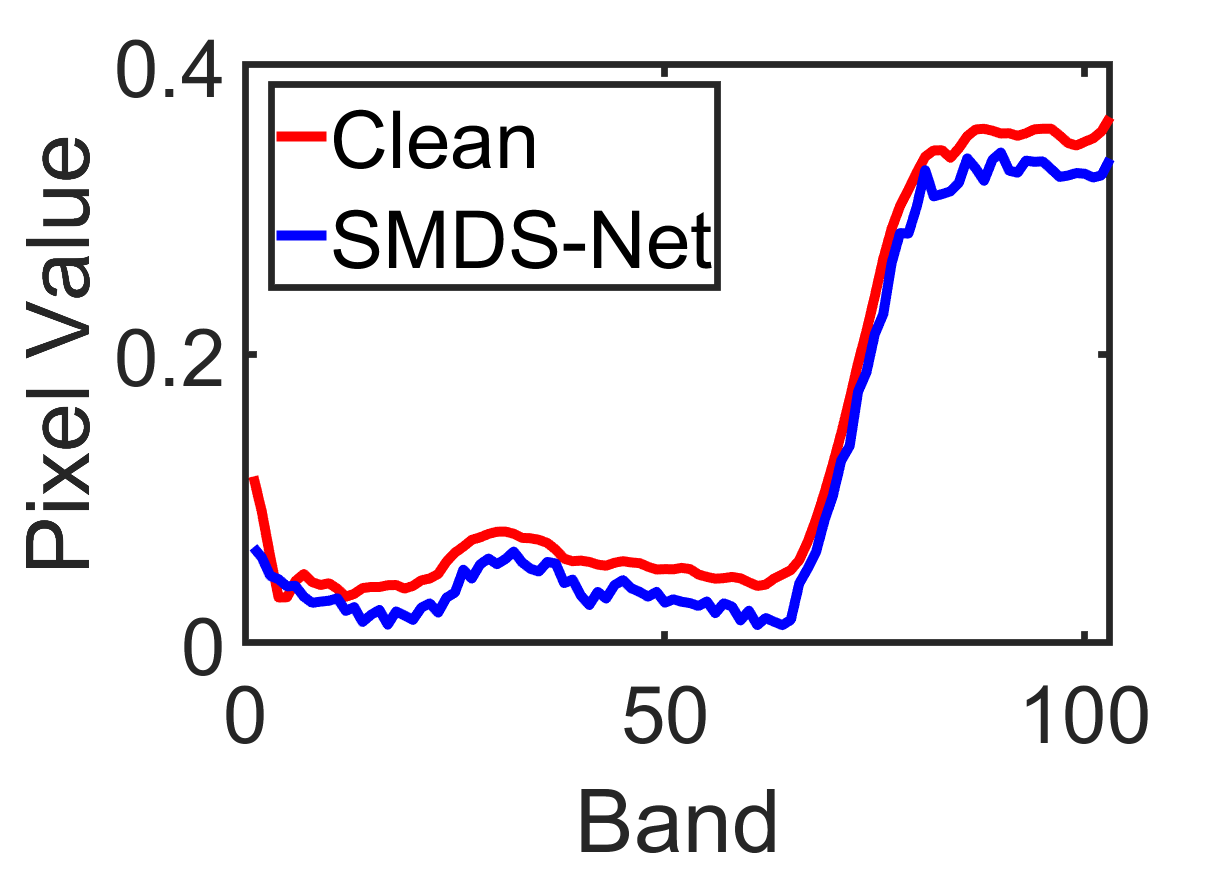}\label{ROC-b23}}
    \subfigure[]{
    \includegraphics[width=0.619 in,height=0.464 in]{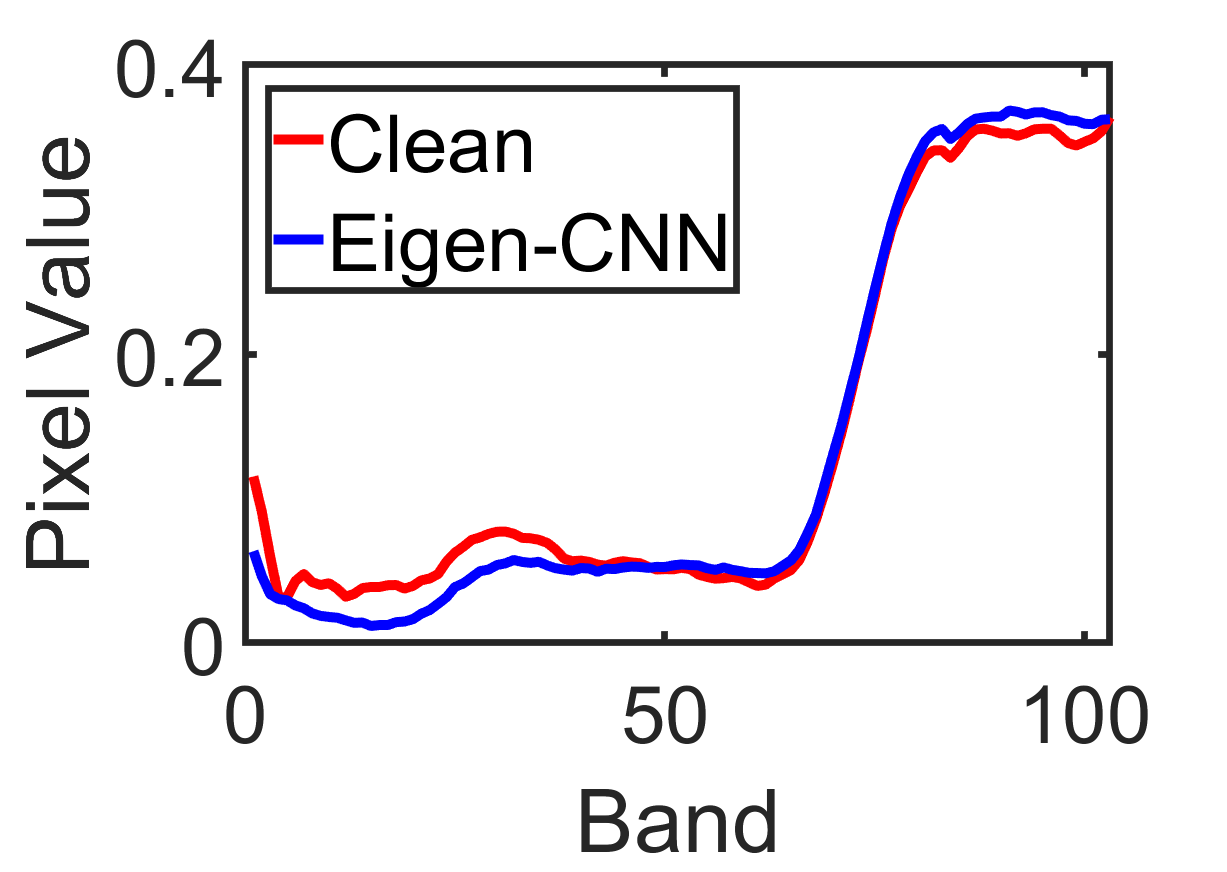}\label{ROC-b22}}
    \subfigure[]{
    \includegraphics[width=0.619 in,height=0.464 in]{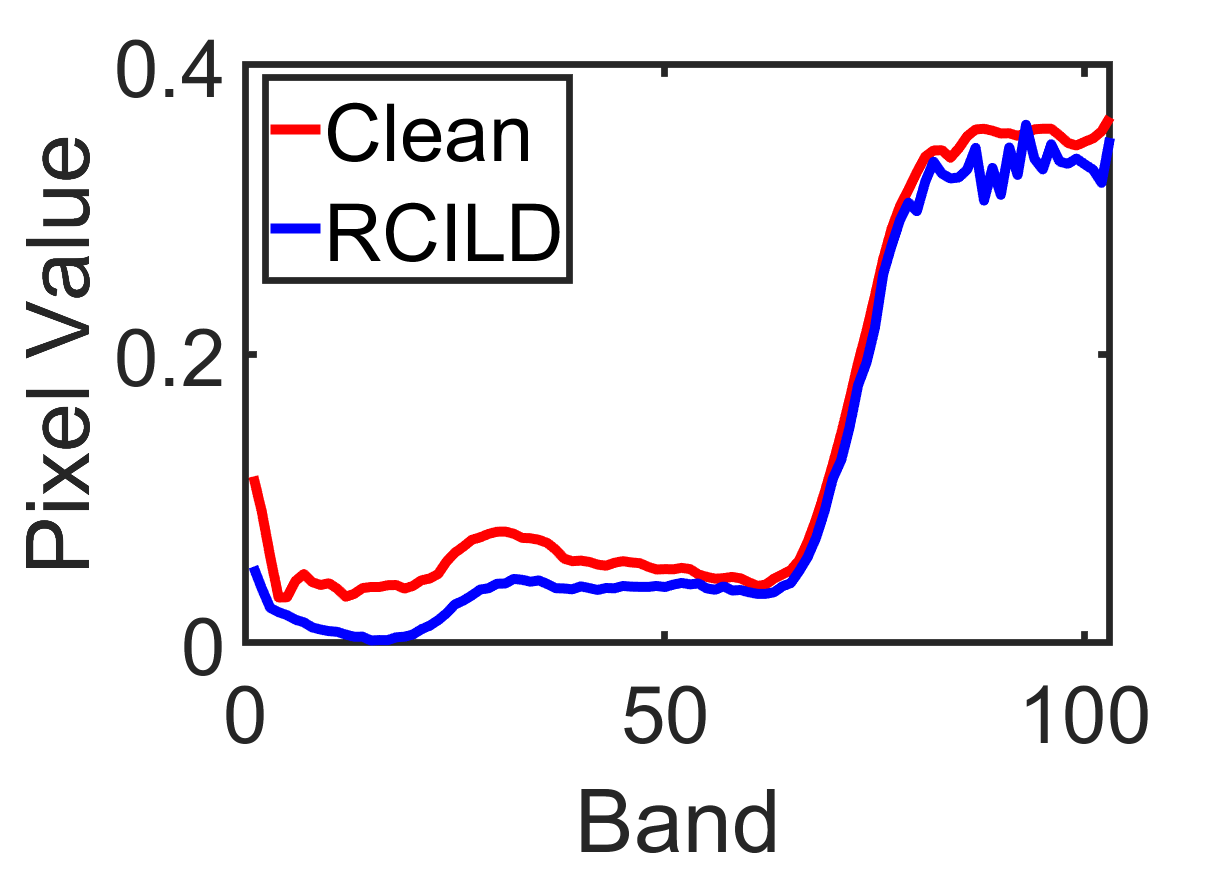}\label{ROC-b23}}
    \subfigure[]{
    \includegraphics[width=0.619 in,height=0.464 in]{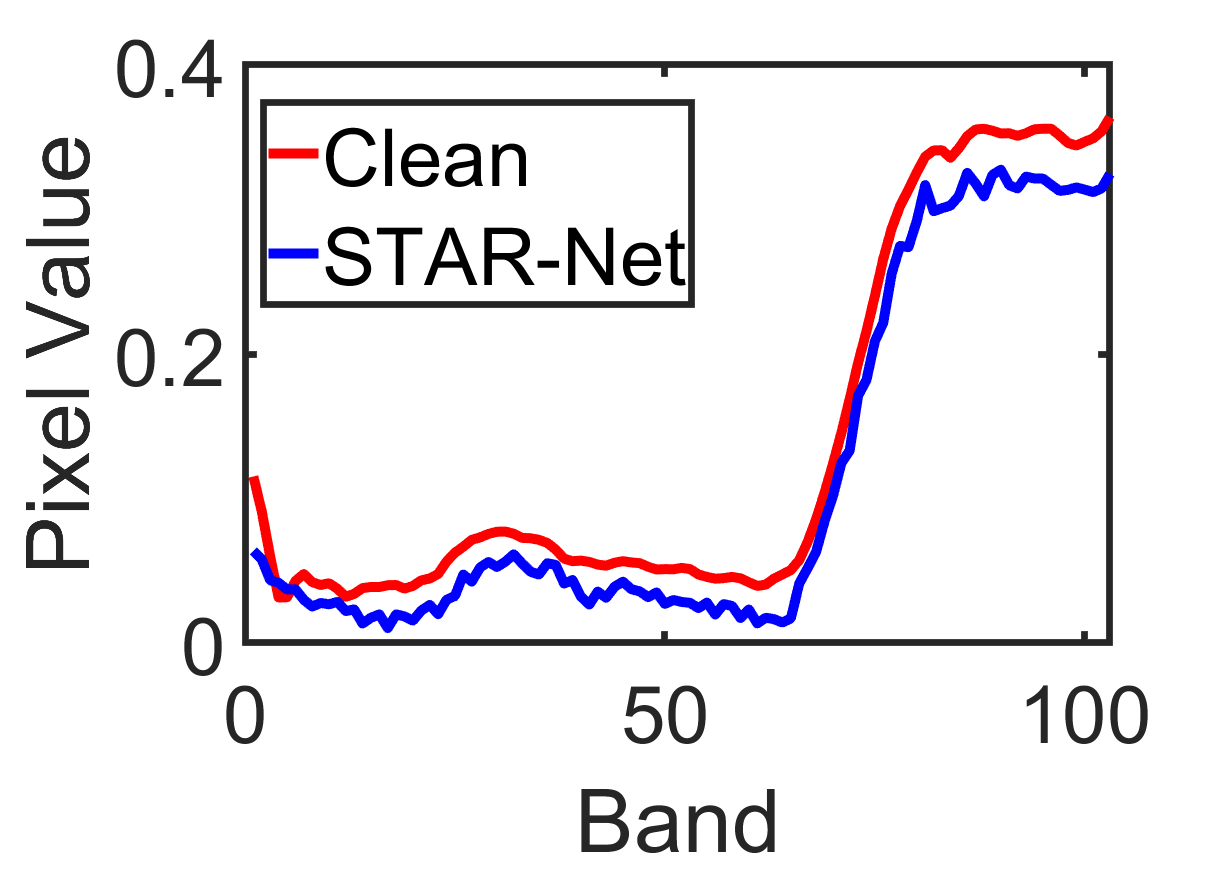}\label{ROC-b24}}
    \subfigure[]{
    \includegraphics[width=0.619 in,height=0.464 in]{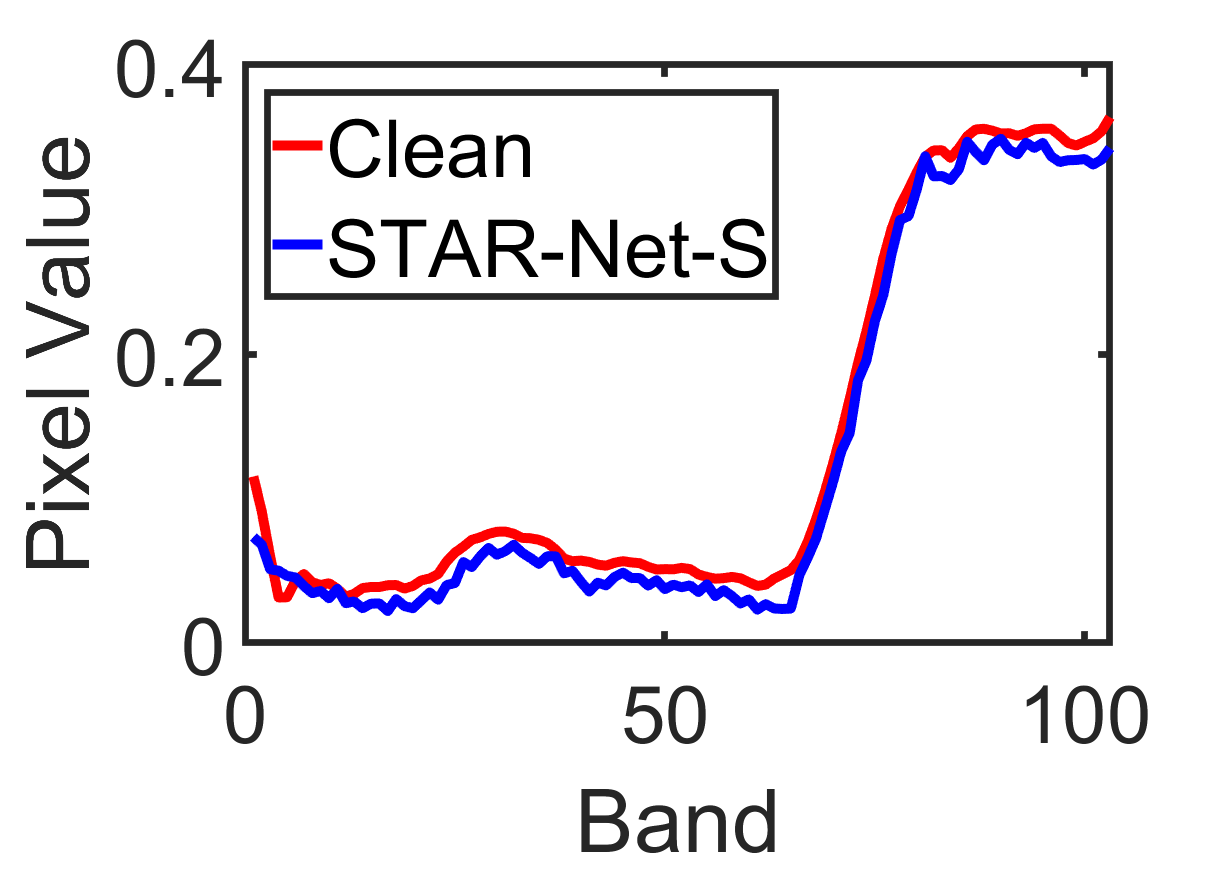}\label{ROC-b24}}
         \vskip-0.2cm
  \caption{The denoising results of pixel (90, 130) on PaviaU with the noise variance of 50.  (a) Clean, (b) Noisy, (c) BM4D, (d) LLRT, (e) LRTDTV, (f) NGMeet, (g) NLSSR, (h) FastHyMix, (i) HSI-SDeCNN, (j) SMDS-Net, (k) Eigen-CNN, (l) RCILD, (m) STAR-Net, (n) STAR-Net-S}.\label{PaviaUpixel}
\end{figure*}

\begin{table*}[h!]
	\centering
	\renewcommand\arraystretch{1.2}
	\caption{Comparison of all methods on the non-Gaussian noise case. The top two values are marked as \textcolor[rgb]{1.00,0.00,0.00}{red} and \textcolor[rgb]{0.00,0.00,1.00}{blue}.}\label{nogaussable}
	\vskip-0.2cm
	\setlength{\tabcolsep}{4.5 pt}
\resizebox{\textwidth}{!}{
	\begin{tabular}{ccccccccccccccc}
		\toprule
		{Dataset}   & {Index} & {Noisy}  & \makecell[c]{BM4D\\}   & \makecell[c]{LLRT\\}                       & \makecell[c]{LRTDTV\\} & \makecell[c]{NGMeet\\}& \makecell[c]{NLSSR\\}&\makecell[c]{FastHy\\Mix} & \makecell[c]{HSI-SDe\\CNN}  & \makecell[c]{SMDS-\\Net}    &\makecell[c]{Eigen-\\CNN}   & \makecell[c]{RCILD\\}         & \makecell[c]{STAR-\\Net}        & \makecell[c]{STAR-\\Net-S}  \\ \hline 
		\multirow{4}{*}{ICVL}     & PSNR $\uparrow$    & 11.695   & 21.274  & 19.308  & 22.587  & 21.916  & 20.305   & 23.319       & 22.033     & 23.573      & 23.436         & 22.832      & \textcolor[rgb]{0.00,0.00,1.00}{ 23.767}         & \textcolor[rgb]{1.00,0.00,0.00}{23.847} \\
                    & SSIM $\uparrow$    & 0.070    & 0.549   & 0.331   & 0.743   & 0.516   & 0.417     & 0.779        & 0.593      & 0.754       & 0.729          & 0.812       & \textcolor[rgb]{0.00,0.00,1.00}{ 0.840} & \textcolor[rgb]{1.00,0.00,0.00}{0.843}   \\
                    & SAM $\downarrow$   & 0.623    & 0.178   & 0.262   & 0.169   & 0.205   & 0.233     & 0.154        & 0.181      &\textcolor[rgb]{0.00,0.00,1.00}{ 0.097}       & 0.163          & 0.158       & \textcolor[rgb]{0.00,0.00,1.00}{ 0.097}          & \textcolor[rgb]{1.00,0.00,0.00}{0.096}\\
                     & ERGAS $\downarrow$ & 715.955  & 247.539 & 319.053 & 233.654 & 247.198 & 256.319   & 205.134      & 237.444    & 184.877     & 201.632        & 200.985     & \textcolor[rgb]{0.00,0.00,1.00}{ 180.920}        & \textcolor[rgb]{1.00,0.00,0.00}{179.610} \\ \hline
		\multirow{4}{*}{PaviaU}     & PSNR $\uparrow$    & 11.200   & 19.846  & 20.828  & 21.977  & 21.262  & 22.119        & \textcolor[rgb]{1.00,0.00,0.00}{24.363} & 21.142     & 22.360      & 22.434      & 21.594       & 22.372         & \textcolor[rgb]{0.00,0.00,1.00}{ 22.558}     \\
                         & SSIM $\uparrow$    & 0.079    & 0.482   & 0.540   & 0.655   & 0.576   & 0.714         & \textcolor[rgb]{1.00,0.00,0.00}{0.761}  & 0.595      & 0.722       & 0.736       & 0.738        & 0.728          & \textcolor[rgb]{0.00,0.00,1.00}{ 0.746}      \\
                         & SAM $\downarrow$   & 0.839    & 0.364   & 0.316   & 0.265   & 0.322   & 0.260         & 0.282           & 0.270      & \textcolor[rgb]{0.00,0.00,1.00}{ 0.205} & 0.251       & 0.239        & \textcolor[rgb]{0.00,0.00,1.00}{ 0.205}    & \textcolor[rgb]{1.00,0.00,0.00}{0.201}   \\
                         & ERGAS $\downarrow$ & 912.728  & 405.510 & 394.990 & 354.904 & 352.518 & \textcolor[rgb]{0.00,0.00,1.00}{ 259.708} & 272.636         & 367.037    & 264.036     & 315.859     & 260.234      & 263.723        & \textcolor[rgb]{1.00,0.00,0.00}{259.505} \\ \hline
		\multirow{4}{*}{Average} & PSNR $\uparrow$    & 11.448  & 20.560                                                  & 20.068                                                  & 22.282                                                    & 21.589                                                    & 21.212                                                           & \textcolor[rgb]{1.00,0.00,0.00}{23.841}                                                              & 21.587                                                                        & 22.966                                                                      & 22.935                                                                       & 22.213                                                   & 23.069                                                                      & \textcolor[rgb]{0.00,0.00,1.00}{ 23.202}                                                                  \\
                         & SSIM $\uparrow$    & 0.075   & 0.516                                                   & 0.436                                                   & 0.699                                                     & 0.546                                                     & 0.565                                                            & 0.770                                                                        & 0.594                                                                         & 0.738                                                                       & 0.732                                                                        & 0.775                                                    & \textcolor[rgb]{0.00,0.00,1.00}{ 0.784}                                                                 & \textcolor[rgb]{1.00,0.00,0.00}{0.794}                                                                \\
                         & SAM $\downarrow$   & 0.731   & 0.271                                                   & 0.289                                                   & 0.217                                                     & 0.264                                                     & 0.246                                                            & 0.218                                                                        & 0.226                                                                         & \textcolor[rgb]{0.00,0.00,1.00}{ 0.151}                                                                 & 0.207                                                                        & 0.199                                                    & \textcolor[rgb]{0.00,0.00,1.00}{ 0.151}                                                                 & \textcolor[rgb]{1.00,0.00,0.00}{0.148}                                                                \\
                         & ERGAS $\downarrow$ & 814.342 & 326.524                                                 & 357.021                                                 & 294.279                                                   & 299.858                                                   & 258.014                                                          & 238.885                                                                      & 302.241                                                                       & 224.456                                                                     & 258.745                                                                      & 230.609                                                  & \textcolor[rgb]{0.00,0.00,1.00}{222.321}                                                               & \textcolor[rgb]{1.00,0.00,0.00}{219.557}   \\ \bottomrule
	\end{tabular}}
\end{table*}

\subsubsection{Experiments on PaviaU}

In addition to the small-scale dataset, we also perform denoising experiments on the large-scale PaviaU dataset.
Table \ref{PUtable} summarizes the denoising performance of all methods across five different noise scenarios on PaviaU.
It can be observed from Table \ref{PUtable} that both STAR-Net and STAR-Net-S achieve superior performance when only Gaussian noise is present.
In the case of sparse noise, although FastHyMix achieves higher PSNR and SSIM values than STAR-Net-S, its performance on the SAM and ERGAS metrics is less competitive.
Therefore, considering all four metrics, STAR-Net-S demonstrates more stable and balanced denoising performance.
Meanwhile, Table \ref{PUtable} also summarizes the average denoising performance, showing that both STAR-Net and STAR-Net-S maintain strong denoising capability even on the large-scale dataset.

The visual denoising results of all methods on the PaviaU dataset are presented in Figure \ref{PU}.
As shown in Figure \ref{PU}, the denoising results of FastHyMix, Eigen-CNN, and RCILD appear satisfactory. 
However, the detailed region in the bottom-left corner exhibits noticeable color distortion compared to the clean RSI.
Meanwhile, STAR-Net and STAR-Net-S exhibit denoising performance and color fidelity that are closest to the clean RSI.
Additionally, Figure \ref{PaviaUpixel} presents the spectral reflectance curves at pixel (90, 130) for all methods on the PaviaU dataset. As shown in Figure \ref{PUb21} and Figure \ref{PUb22}, the added noise leads to significant spectral distortion. Figure \ref{PaviaUpixel} shows that the spectral changes of FastHyMix, Eigen-CNN, STAR-Net and Star-Net-S conform to the change trend of clean RSI. However, the spectral curve of STAR-Net-S nearly overlaps with that of the clean RSI, indicating its effectiveness in denoising each band and its strong spectral recovery capability.

In addition, to evaluate the scalability of the methods for non-Gaussian noise denoising, Table \ref{nogaussable} presents the performance of all methods on the two datasets under the non-Gaussian noise scenario.
As shown in Table \ref{nogaussable}, the PSNR of STAR-Net-S is slightly lower than that of FastHyMix on the PaviaU dataset, ranking second among all methods.
Considering all four metrics, STAR-Net-S demonstrates the most balanced performance and exhibits the most stable denoising capability against non-Gaussian noise.

\begin{figure*}[t]
\makeatletter
\renewcommand{\@thesubfigure}{\hskip\subfiglabelskip}
\makeatother
\centering
\raisebox{0.5\height}{
\subfigure[(a)]{
\includegraphics[width=0.13\linewidth]{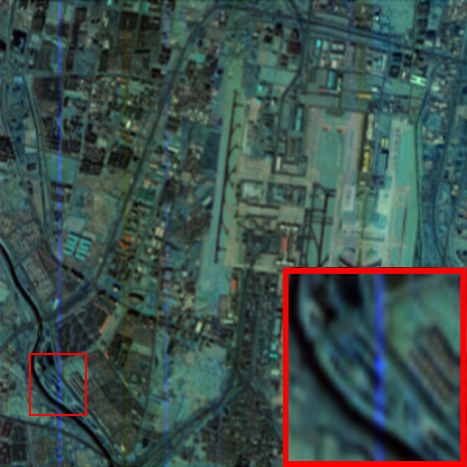}}}
\hspace{0.5mm}
\begin{minipage}[b]{0.13\textwidth}
    \centering
    \subfigure[(b)]{
    \includegraphics[width=\textwidth]{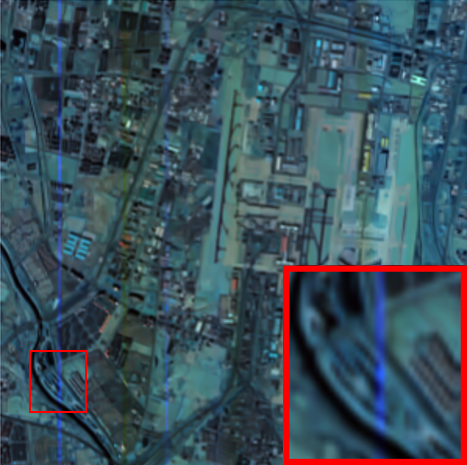}}\\
    \subfigure[(h)]{
    \includegraphics[width=\textwidth]{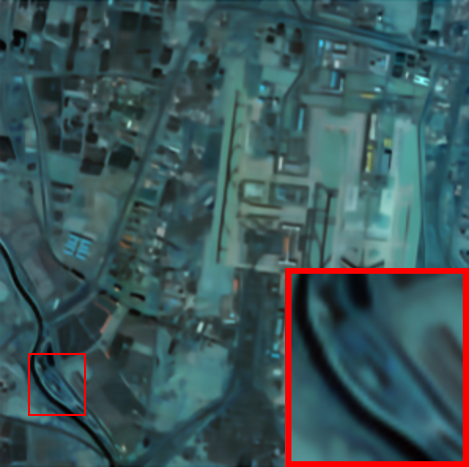}}
\end{minipage}%
\hspace{0.5mm}
\begin{minipage}[b]{0.13\textwidth}
    \centering
    \subfigure[(c)]{
    \includegraphics[width=\textwidth]{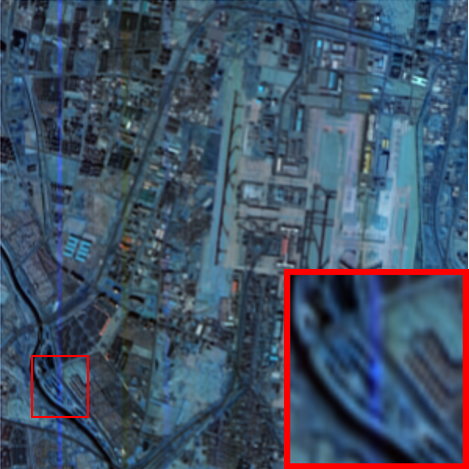}}\\
    \subfigure[(i)]{
    \includegraphics[width=\textwidth]{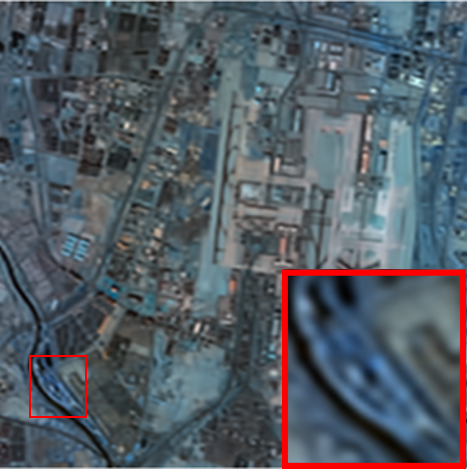}}
\end{minipage}%
\hspace{0.5mm}
\begin{minipage}[b]{0.13\textwidth}
    \centering
    \subfigure[(d)]{
    \includegraphics[width=\textwidth]{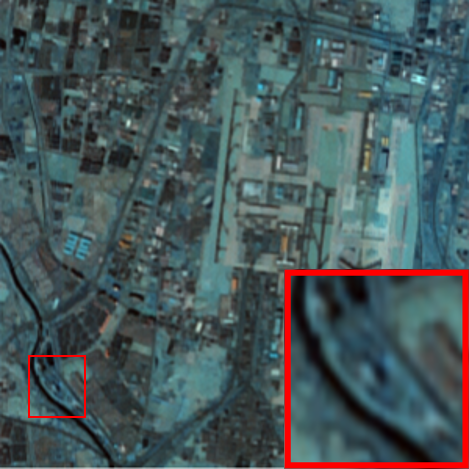}}\\
    \subfigure[(j)]{
    \includegraphics[width=\textwidth]{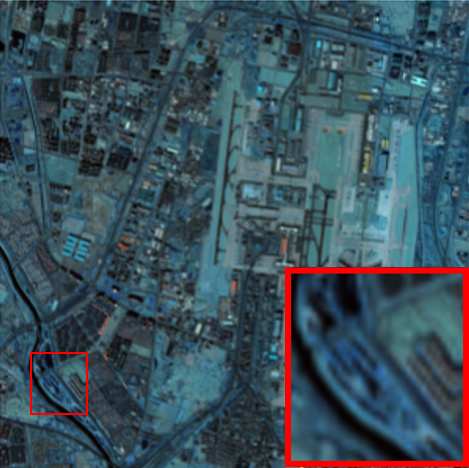}}
\end{minipage}%
\hspace{0.5mm}
\begin{minipage}[b]{0.13\textwidth}
    \centering
    \subfigure[(e)]{
    \includegraphics[width=\textwidth]{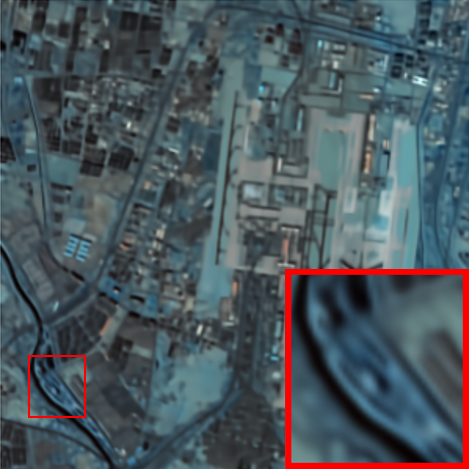}}\\
    \subfigure[(k)]{
    \includegraphics[width=\textwidth]{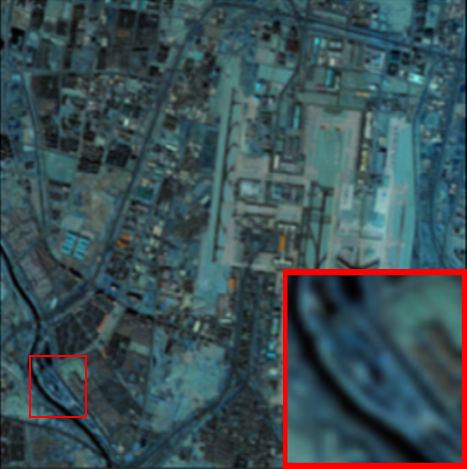}}
\end{minipage}%
\hspace{0.5mm}
\begin{minipage}[b]{0.13\textwidth}
    \centering
    \subfigure[(f)]{
    \includegraphics[width=\textwidth]{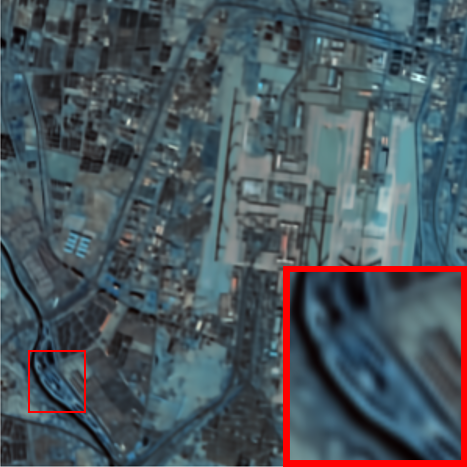}}\\
    \subfigure[(l)]{
    \includegraphics[width=\textwidth]{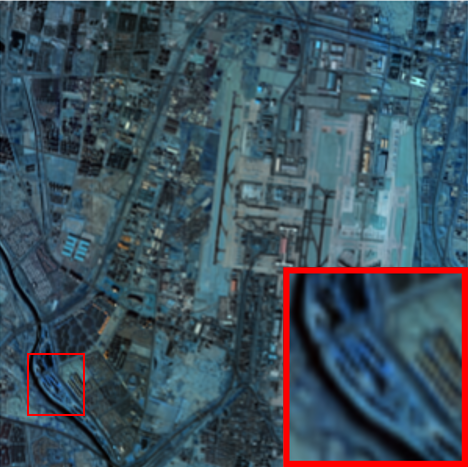}}
\end{minipage}%
\hspace{0.5mm}
\begin{minipage}[b]{0.13\textwidth}
    \centering
    \subfigure[(g)]{
    \includegraphics[width=\textwidth]{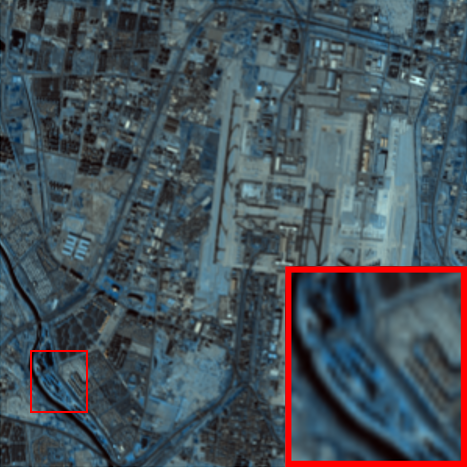}}\\
    \subfigure[(m)]{
    \includegraphics[width=\textwidth]{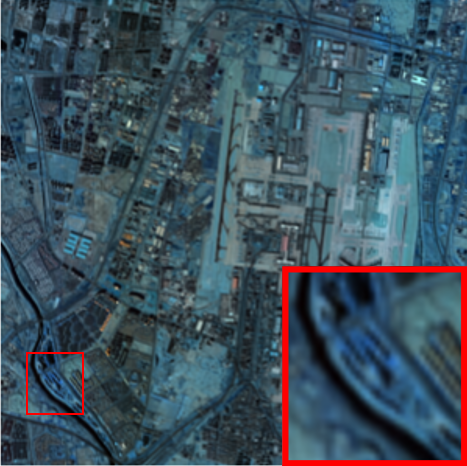}}
\end{minipage}%
\vskip-0.2cm
\caption{Denoising results on Beijing Capital Airport. The false-color images are generated by combining bands 42, 67, and 132. (a) Noisy, (b) BM4D, (c) LLRT, (d) LRTDTV, (e) NGMeet, (f) NLSSR, (g) FastHyMix, (h) HSI-SDeCNN, (i) SMDS-Net, (j) Eigen-CNN, (k) RCILD, (l) STAR-Net, (m) STAR-Net-S.}
\label{captical}
\end{figure*}

\begin{figure*}[t]
\makeatletter
\renewcommand{\@thesubfigure}{\hskip\subfiglabelskip}
\makeatother
\centering
\raisebox{0.5\height}{
\subfigure[(a)]{
\includegraphics[width=0.132\linewidth]{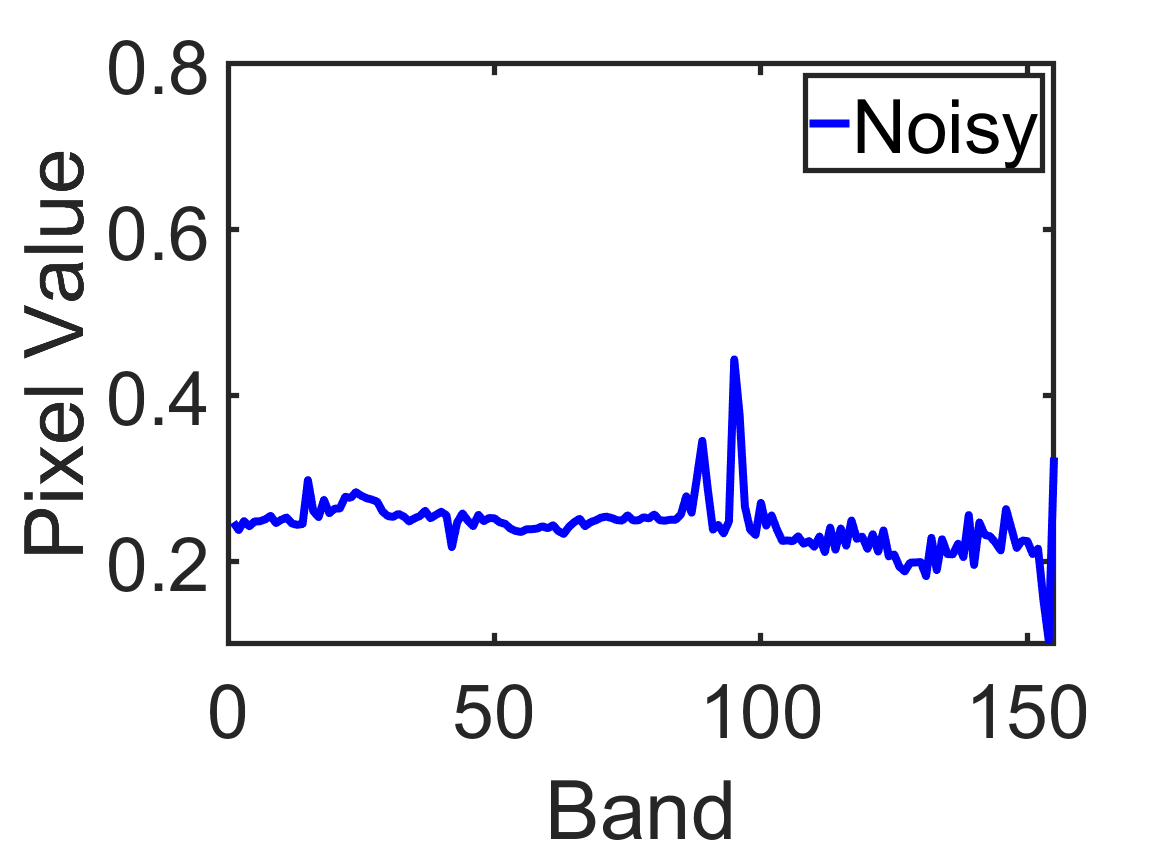}}}
\hspace{0.1mm}
\begin{minipage}[b]{0.132\textwidth}
    \centering
    \subfigure[(b)]{
    \includegraphics[width=\textwidth,height=0.75\textwidth]{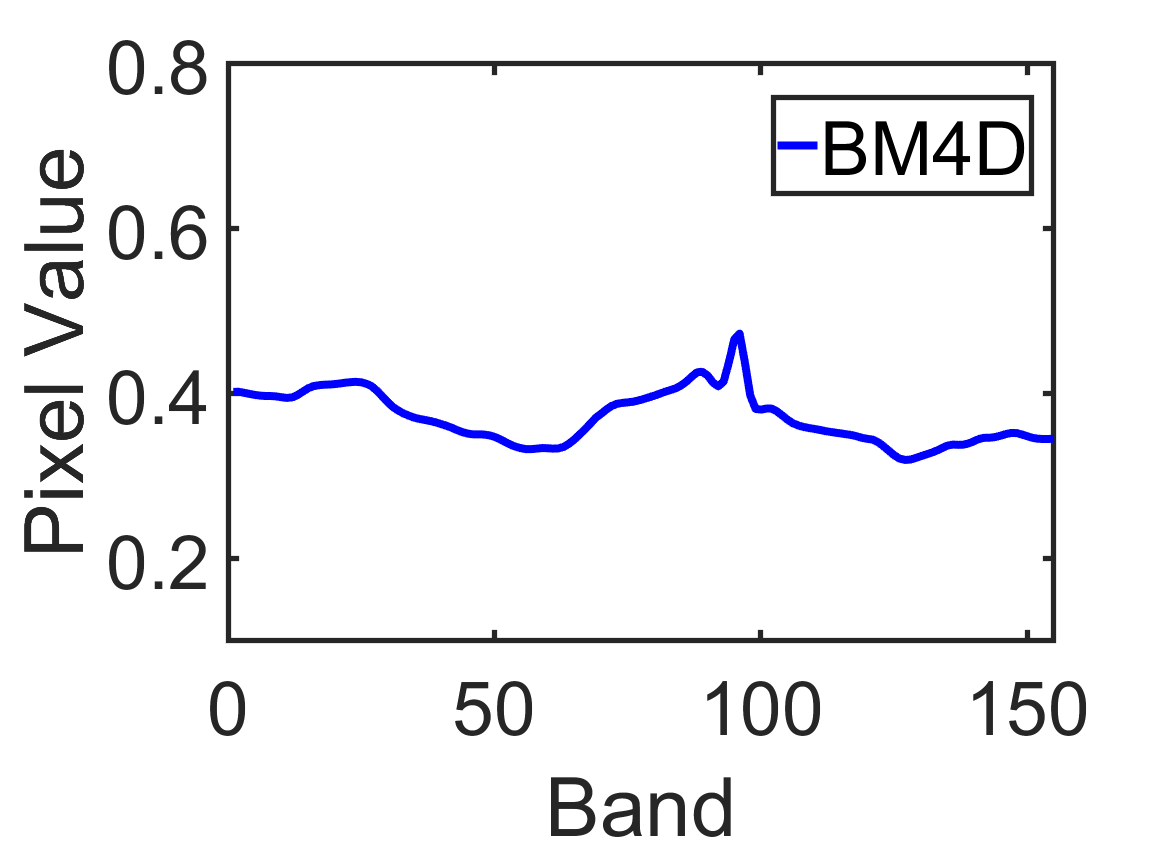}}\\
    \subfigure[(h)]{
    \includegraphics[width=\textwidth,height=0.75\textwidth]{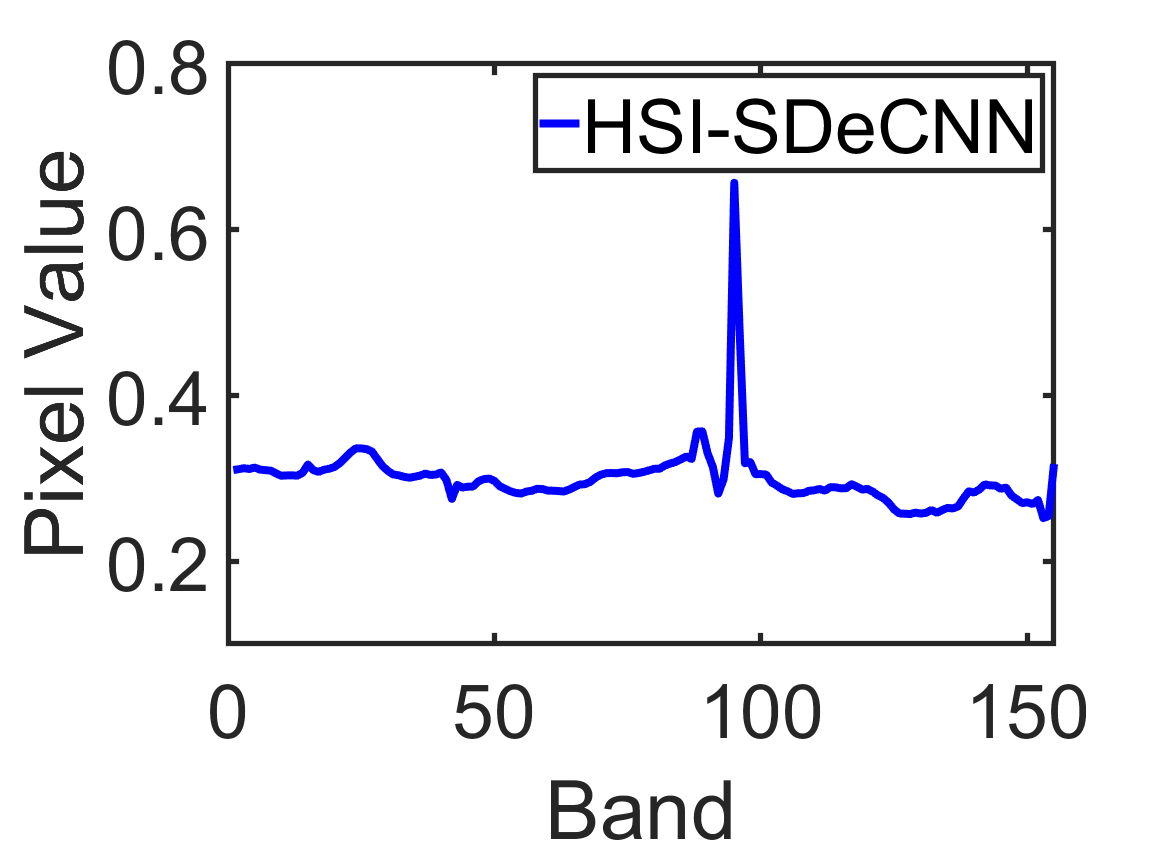}}
\end{minipage}%
\hspace{0.1mm}
\begin{minipage}[b]{0.132\textwidth}
    \centering
    \subfigure[(c)]{
    \includegraphics[width=\textwidth,height=0.75\textwidth]{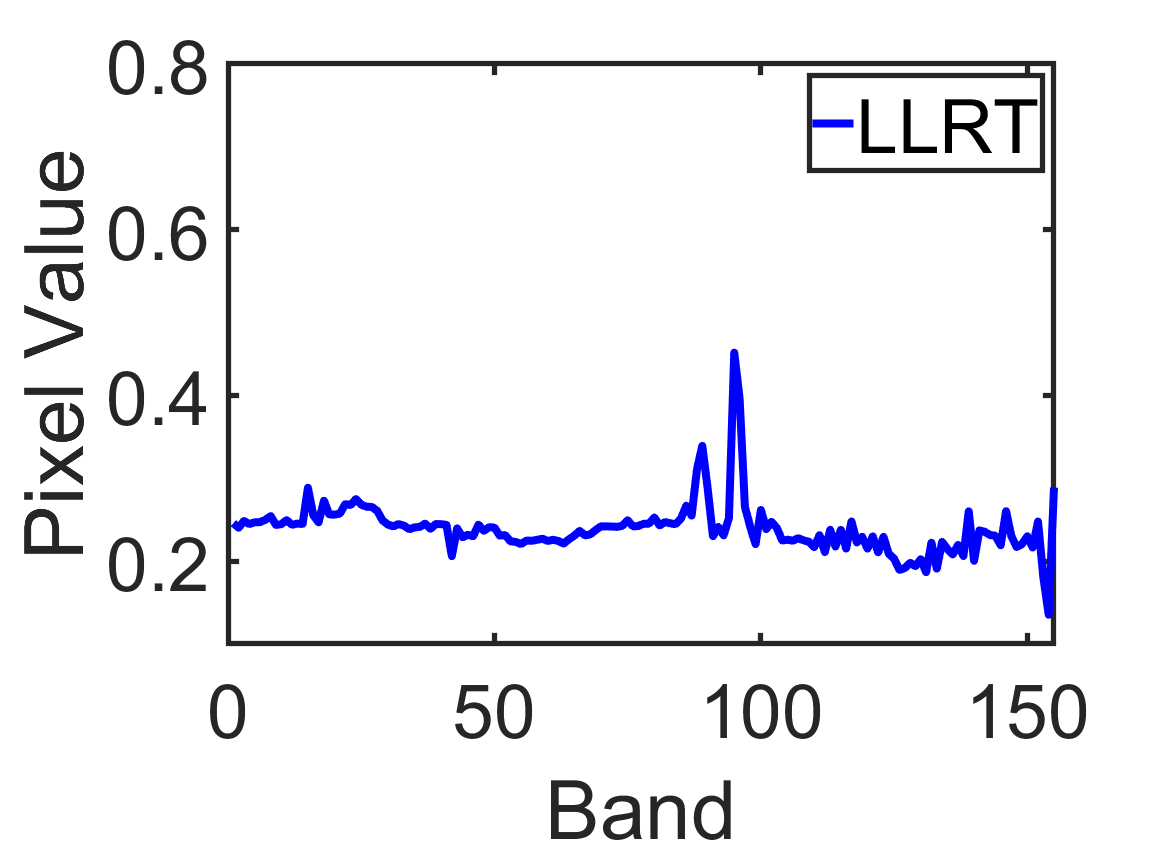}}\\
    \subfigure[(i)]{
    \includegraphics[width=\textwidth,height=0.75\textwidth]{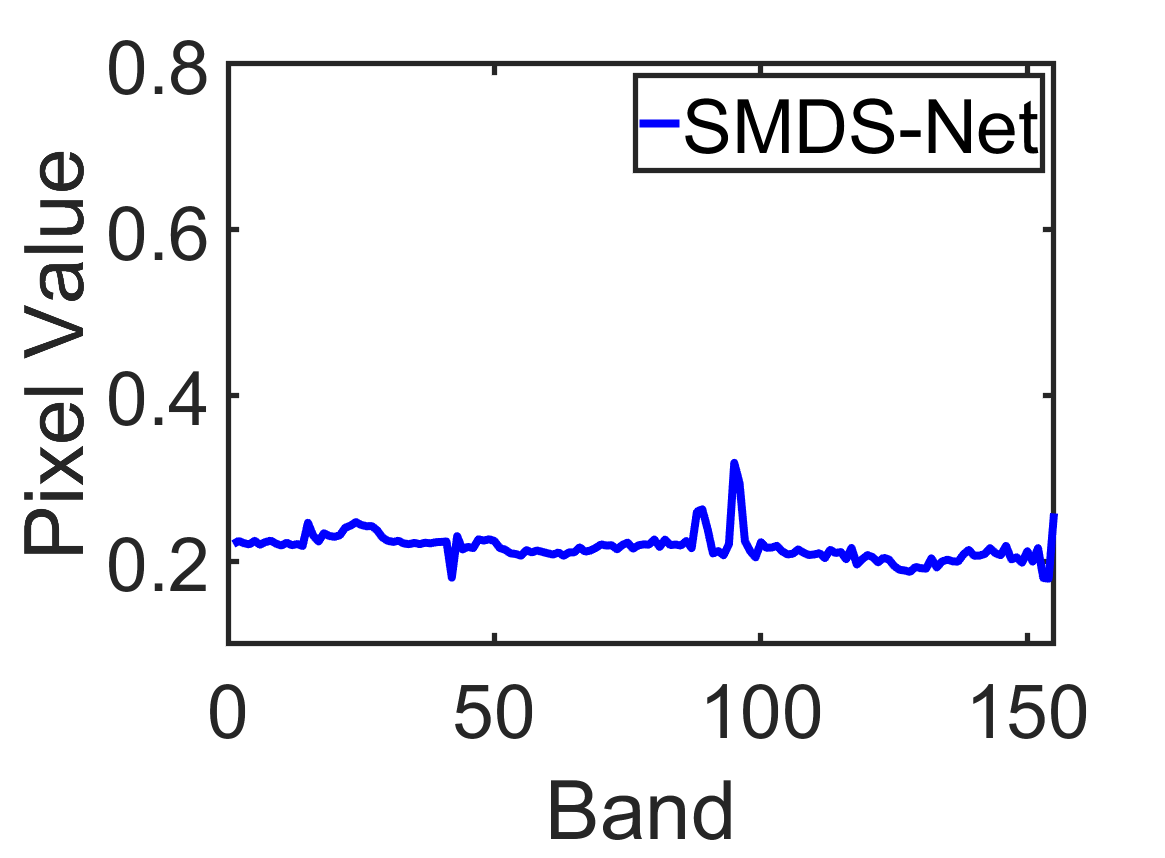}}
\end{minipage}%
\hspace{0.1mm}
\begin{minipage}[b]{0.132\textwidth}
    \centering
    \subfigure[(d)]{
    \includegraphics[width=\textwidth,height=0.75\textwidth]{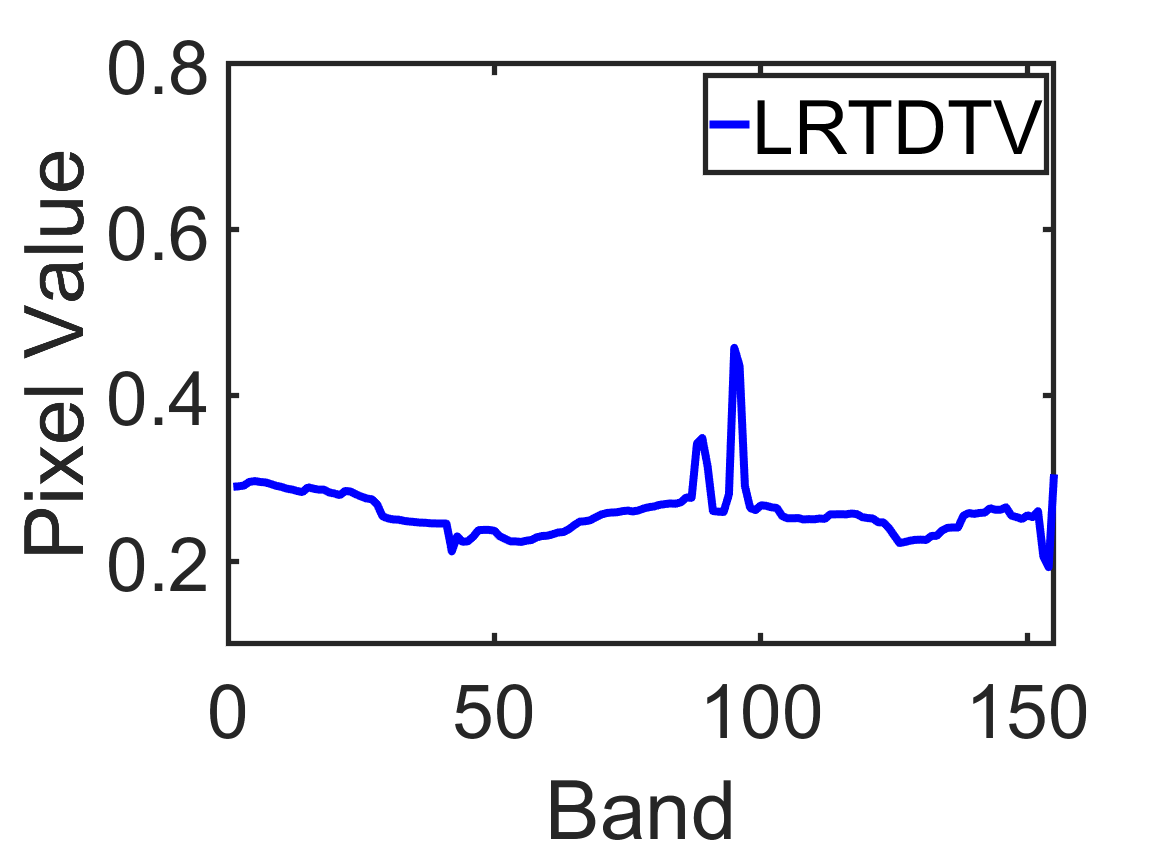}}\\
    \subfigure[(j)]{
    \includegraphics[width=\textwidth,height=0.75\textwidth]{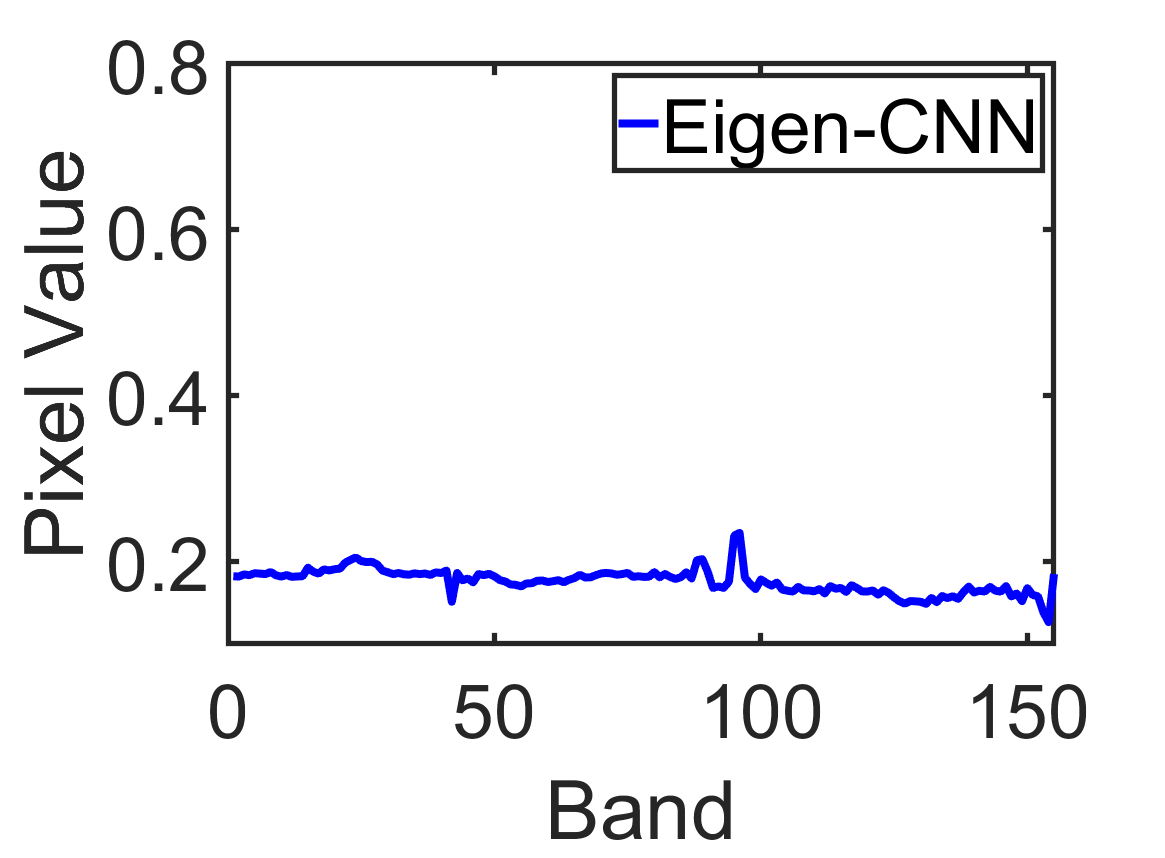}}
\end{minipage}%
\hspace{0.1mm}
\begin{minipage}[b]{0.132\textwidth}
    \centering
    \subfigure[(e)]{
    \includegraphics[width=\textwidth,height=0.75\textwidth]{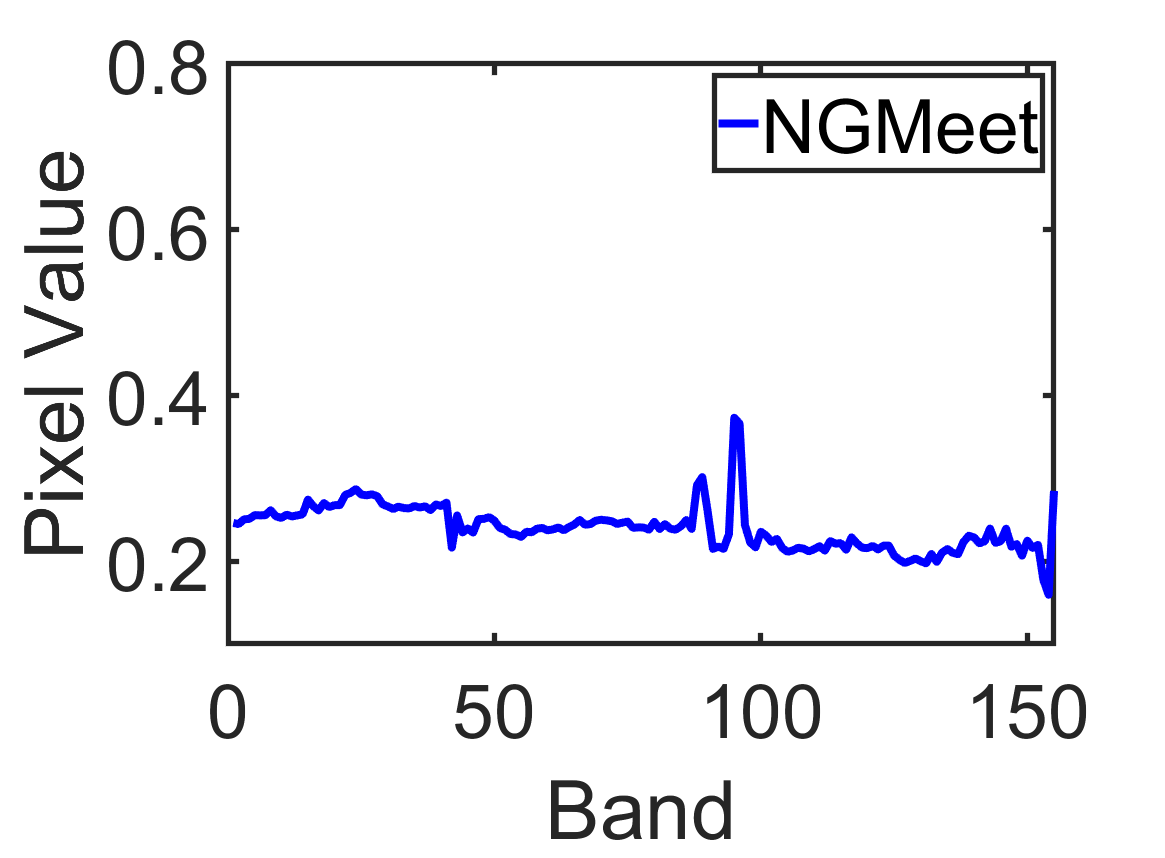}}\\
    \subfigure[(k)]{
    \includegraphics[width=\textwidth,height=0.75\textwidth]{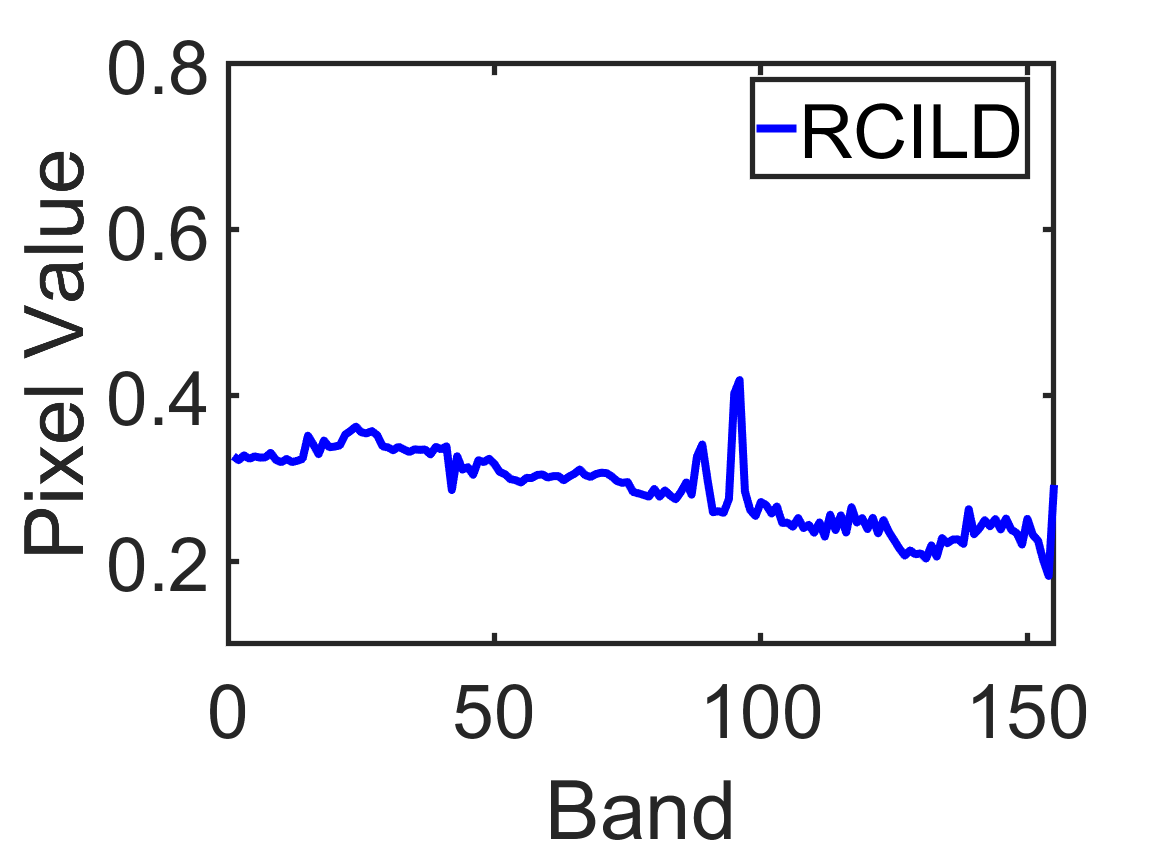}}
\end{minipage}%
\hspace{0.1mm}
\begin{minipage}[b]{0.132\textwidth}
    \centering
    \subfigure[(f)]{
    \includegraphics[width=\textwidth,height=0.75\textwidth]{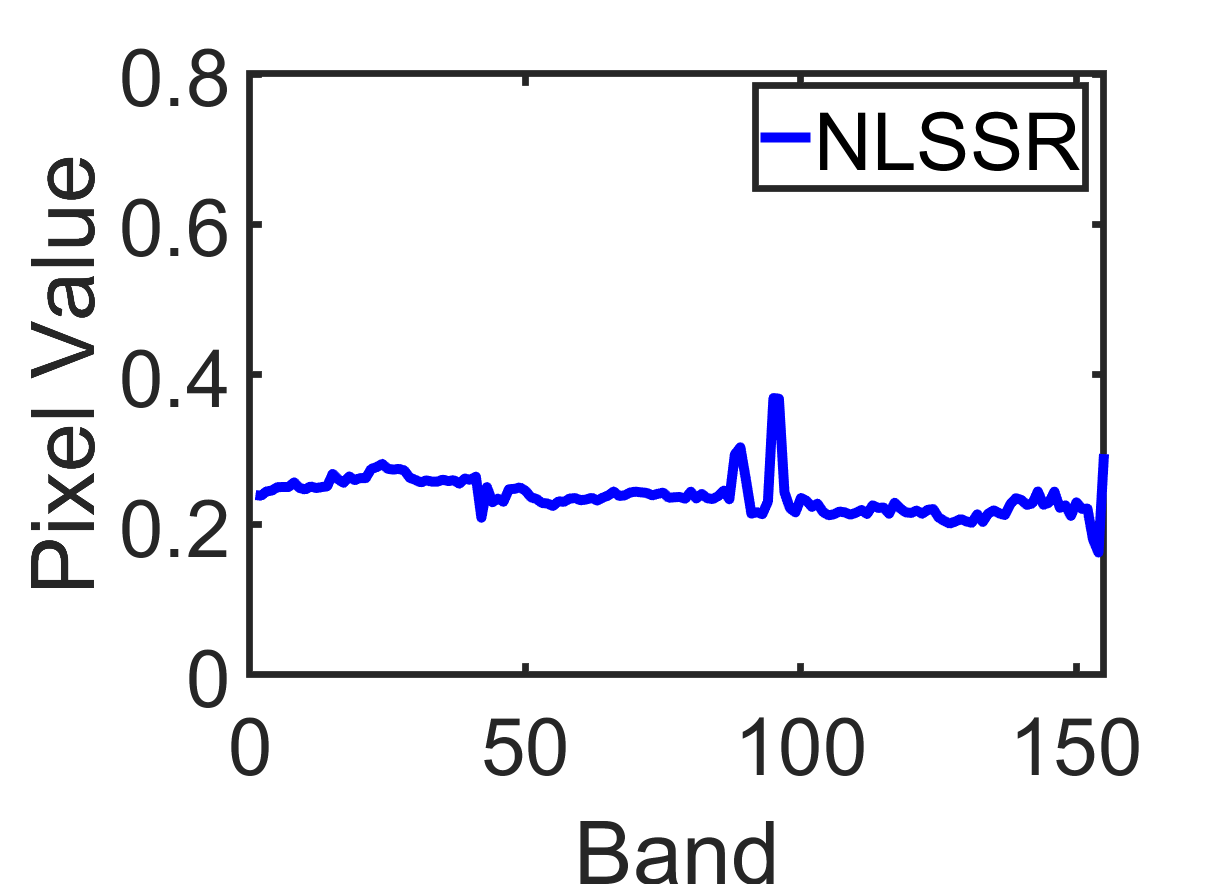}}\\
    \subfigure[(l)]{
    \includegraphics[width=\textwidth,height=0.75\textwidth]{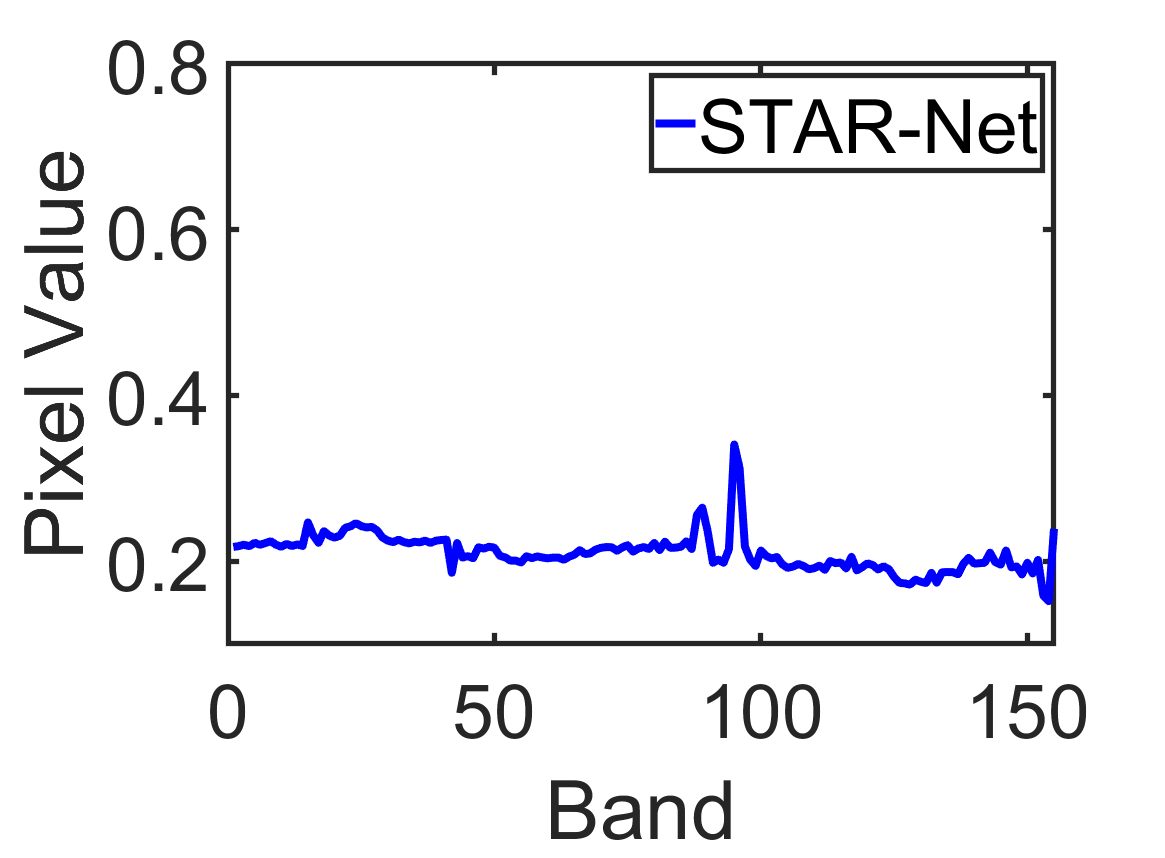}}
\end{minipage}%
\hspace{0.1mm}
\begin{minipage}[b]{0.132\textwidth}
    \centering
    \subfigure[(g)]{
    \includegraphics[width=\textwidth,height=0.75\textwidth]{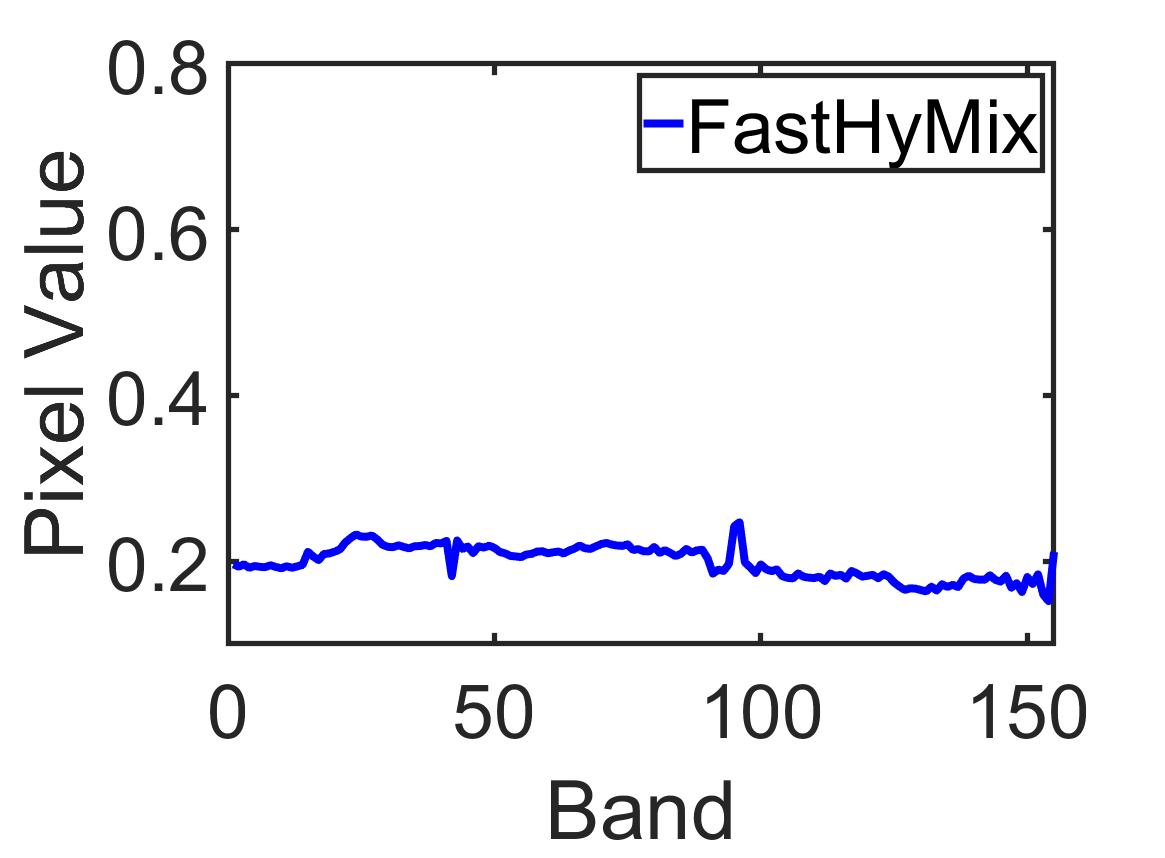}}\\
    \subfigure[(m)]{
    \includegraphics[width=\textwidth,height=0.75\textwidth]{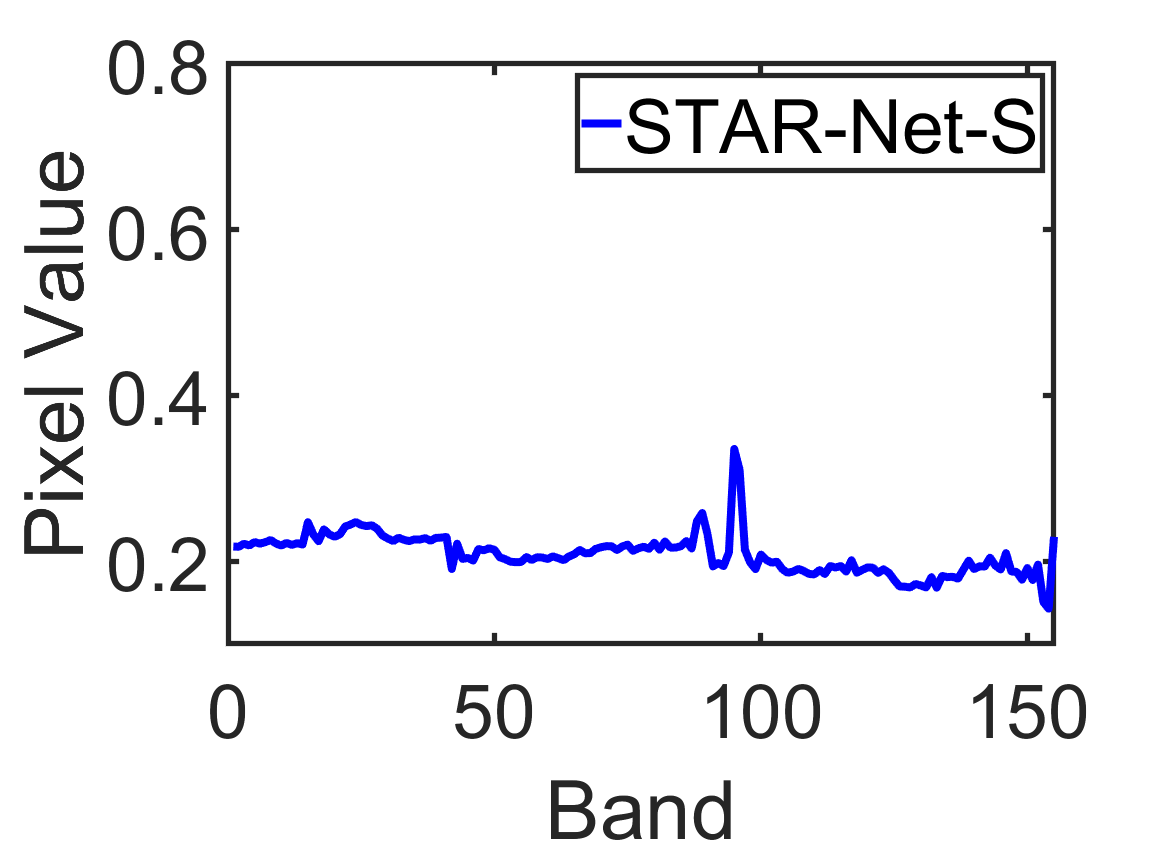}}
\end{minipage}%
\vskip-0.2cm
\caption{Denoising results of pixel (160, 185) on Beijing Capital Airport. (a) Noisy, (b) BM4D, (c) LLRT, (d) LRTDTV, (e) NGMeet, (f) NLSSR, (g) FastHyMix, (h) HSI-SDeCNN, (i) SMDS-Net, (j) Eigen-CNN, (k) RCILD, (l) STAR-Net, (m) STAR-Net-S.}
\label{captialpixel}
\end{figure*}

\subsection{Comparison on Real-World Datasets}
\subsubsection{Experiments on Beijing Capital Airport}

To further evaluate the reliability and adaptability of STAR-Net and STAR-Net-S, we conduct experiments on the Beijing Capital Airport dataset containing real-world noise.
Since there is no clean RSI for comparison, the denoising effect can only be evaluated by analyzing the denoised RSI and the corresponding spectral curve.
As shown in Figure \ref{captical}, BM4D and LLRT exhibit significant noise residuals.
Other methods can remove the noise, but Eigen-CNN, STAR-Net, and STAR-Net-S not only achieve noise removal effectively but also retain the original information of RSI.
In addition, Figure \ref{captialpixel}  shows  the spectral curves for pixel (160, 185).
Compared to Eigen-CNN, STAR-Net, and STAR-Net-S not only preserve the smoothness of the spectral curves but also retain the distinct characteristics of key spectral points.
Now we can conclude that STAR-Net and STAR-Net-S achieve excellent denoising performance.

\begin{figure*}[t]
\makeatletter
\renewcommand{\@thesubfigure}{\hskip\subfiglabelskip}
\makeatother
\centering
\raisebox{0.5\height}{
\subfigure[(a)]{
\includegraphics[width=0.13\linewidth]{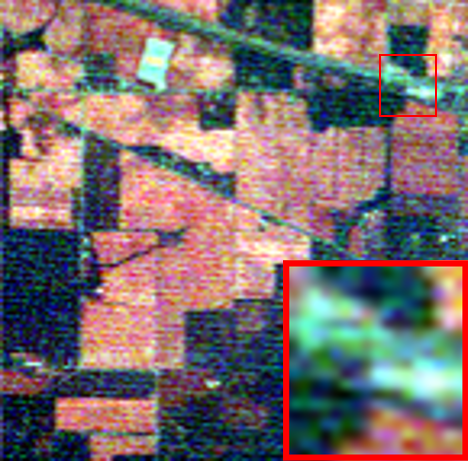}}}
\hspace{0.5mm}
\begin{minipage}[b]{0.13\textwidth}
    \centering
    \subfigure[(b)]{
    \includegraphics[width=\textwidth]{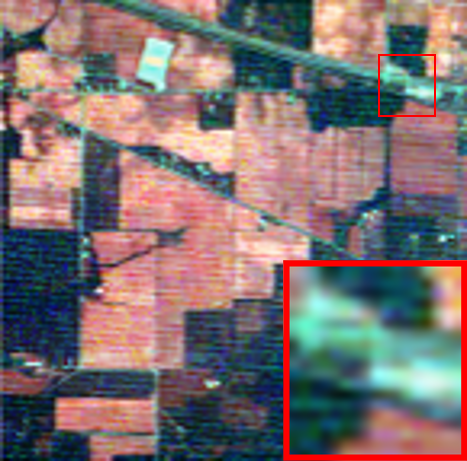}}\\
    \subfigure[(h)]{
    \includegraphics[width=\textwidth]{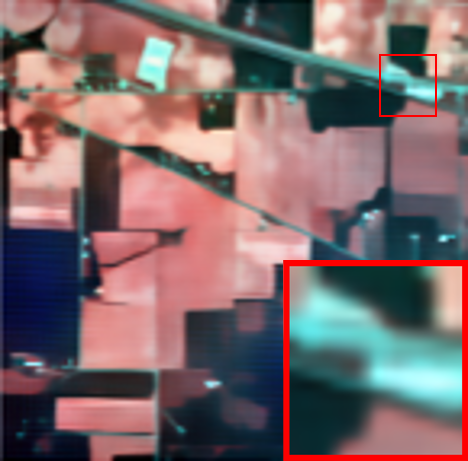}}
\end{minipage}%
\hspace{0.5mm}
\begin{minipage}[b]{0.13\textwidth}
    \centering
    \subfigure[(c)]{
    \includegraphics[width=\textwidth]{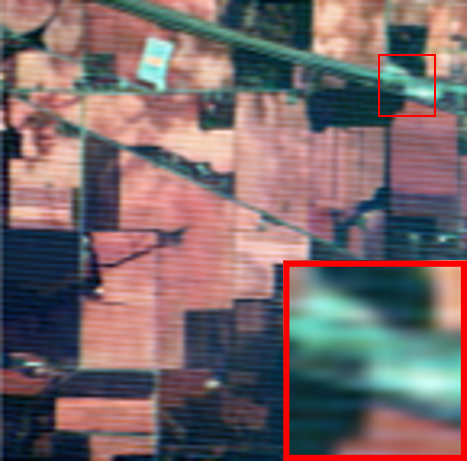}}\\
    \subfigure[(i)]{
    \includegraphics[width=\textwidth]{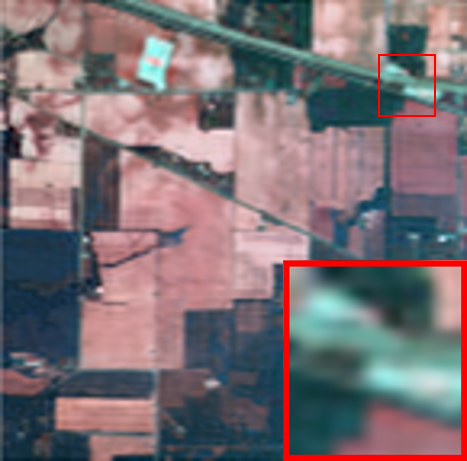}}
\end{minipage}%
\hspace{0.5mm}
\begin{minipage}[b]{0.13\textwidth}
    \centering
    \subfigure[(d)]{
    \includegraphics[width=\textwidth]{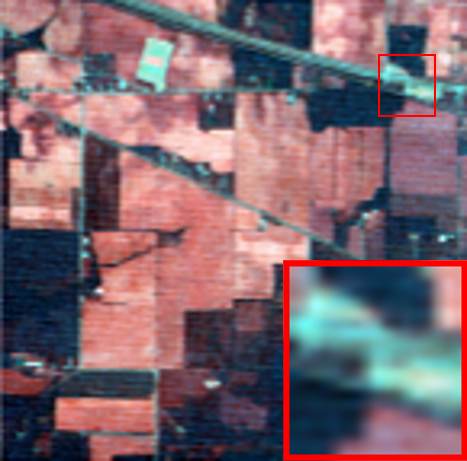}}\\
    \subfigure[(j)]{
    \includegraphics[width=\textwidth]{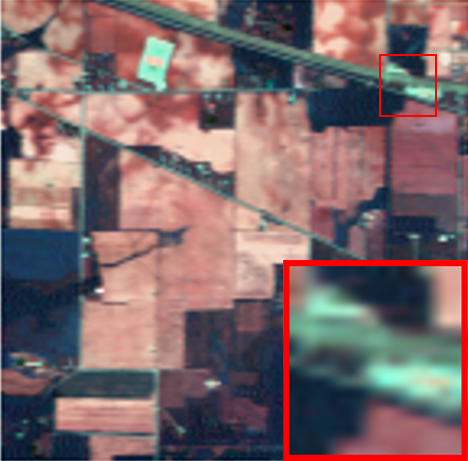}}
\end{minipage}%
\hspace{0.5mm}
\begin{minipage}[b]{0.13\textwidth}
    \centering
    \subfigure[(e)]{
    \includegraphics[width=\textwidth]{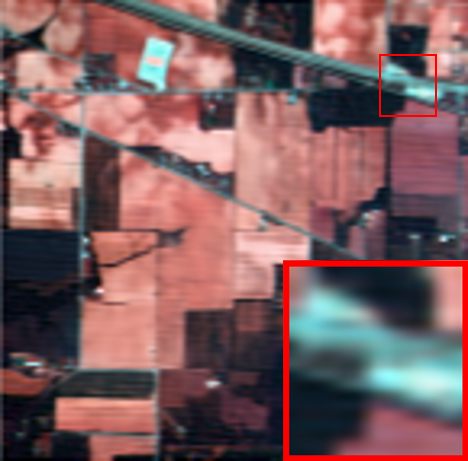}}\\
    \subfigure[(k)]{
    \includegraphics[width=\textwidth]{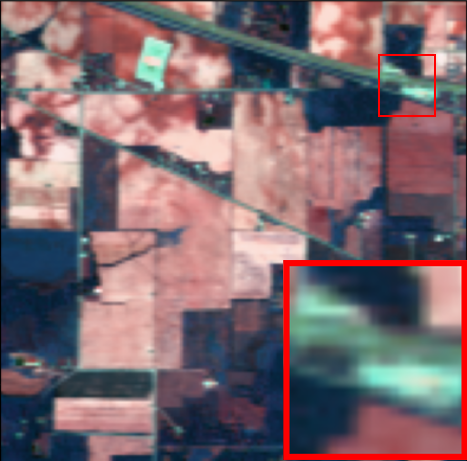}}
\end{minipage}%
\hspace{0.5mm}
\begin{minipage}[b]{0.13\textwidth}
    \centering
    \subfigure[(f)]{
    \includegraphics[width=\textwidth]{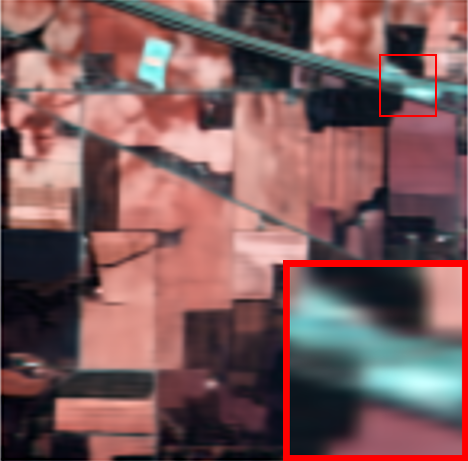}}\\
    \subfigure[(l)]{
    \includegraphics[width=\textwidth]{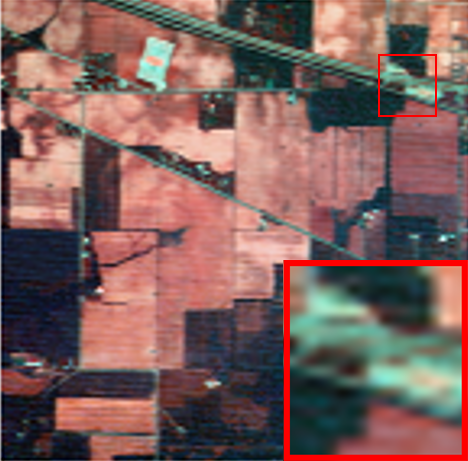}}
\end{minipage}%
\hspace{0.5mm}
\begin{minipage}[b]{0.13\textwidth}
    \centering
    \subfigure[(g)]{
    \includegraphics[width=\textwidth]{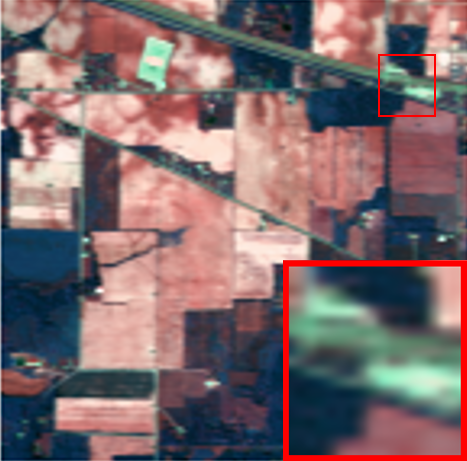}}\\
    \subfigure[(m)]{
    \includegraphics[width=\textwidth]{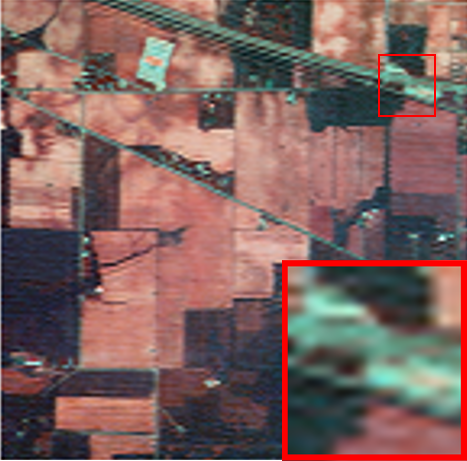}}
\end{minipage}%
\vskip-0.2cm
\caption{Denoising results on Indian Pines. The false-color images are generated by combining bands 1, 2, and 113. (a) Noisy, (b) BM4D, (c) LLRT, (d) LRTDTV, (e) NGMeet, (f) NLSSR, (g) FastHyMix, (h) HSI-SDeCNN, (i) SMDS-Net, (j) Eigen-CNN, (k) RCILD, (l) STAR-Net, (m) STAR-Net-S.}
  \label{indian}
\end{figure*}

\begin{figure*}[t]
\makeatletter
\renewcommand{\@thesubfigure}{\hskip\subfiglabelskip}
\makeatother
\centering
\raisebox{0.5\height}{
\subfigure[(a)]{
\includegraphics[width=0.132\linewidth]{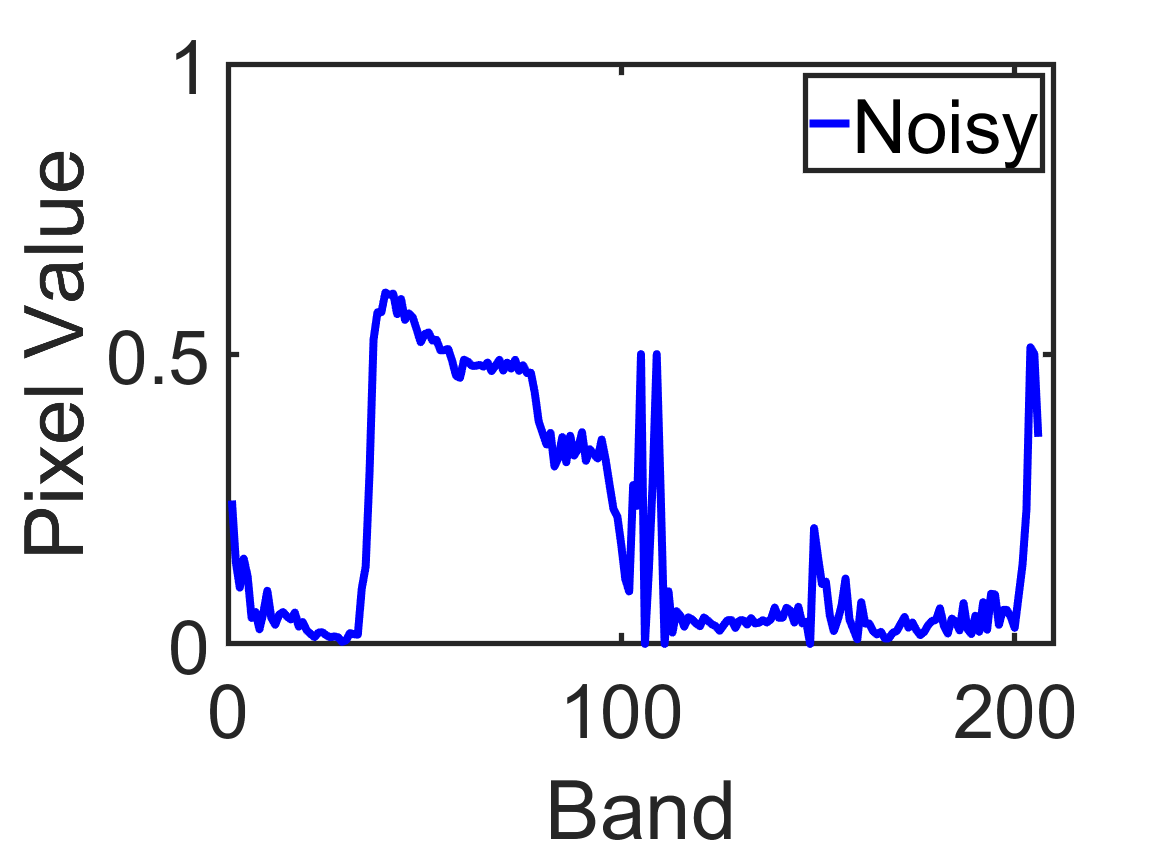}}}
\hspace{0.1mm}
\begin{minipage}[b]{0.132\textwidth}
    \centering
    \subfigure[(b)]{
    \includegraphics[width=\textwidth,height=0.75\textwidth]{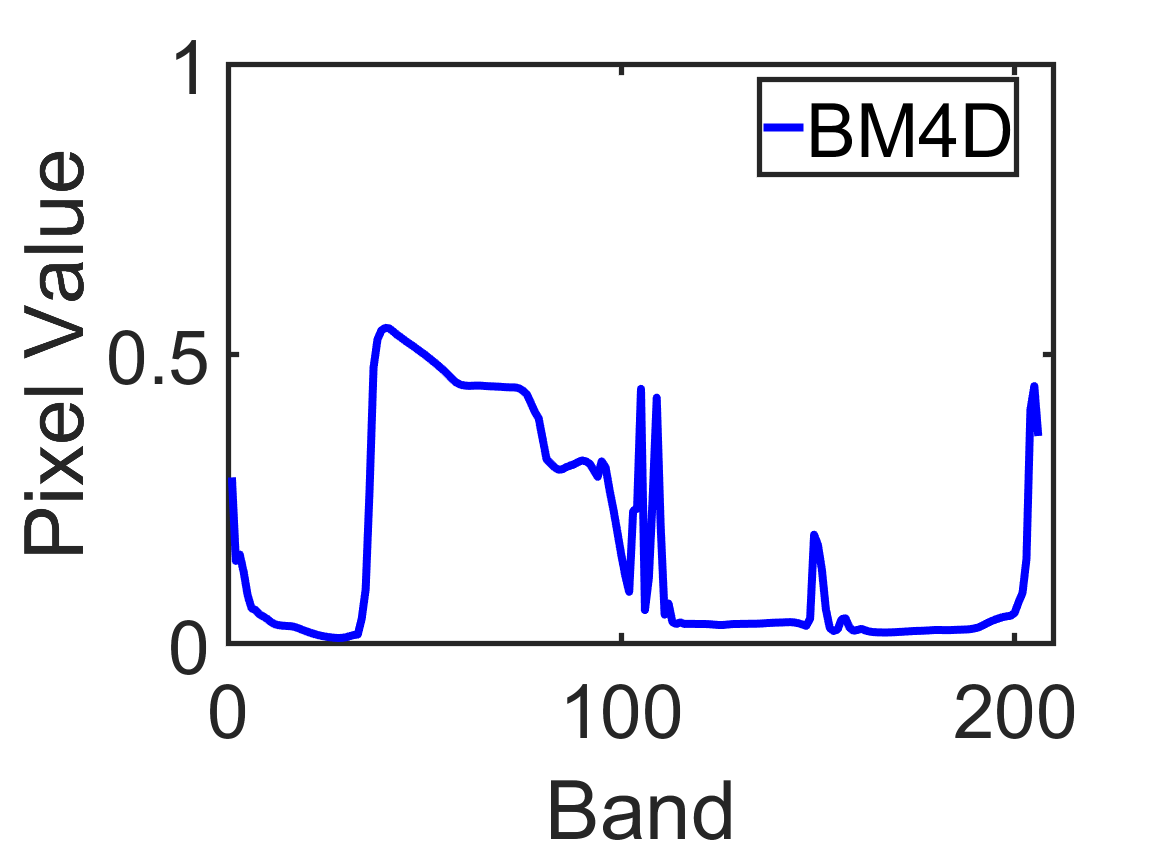}}\\
    \subfigure[(h)]{
    \includegraphics[width=\textwidth,height=0.75\textwidth]{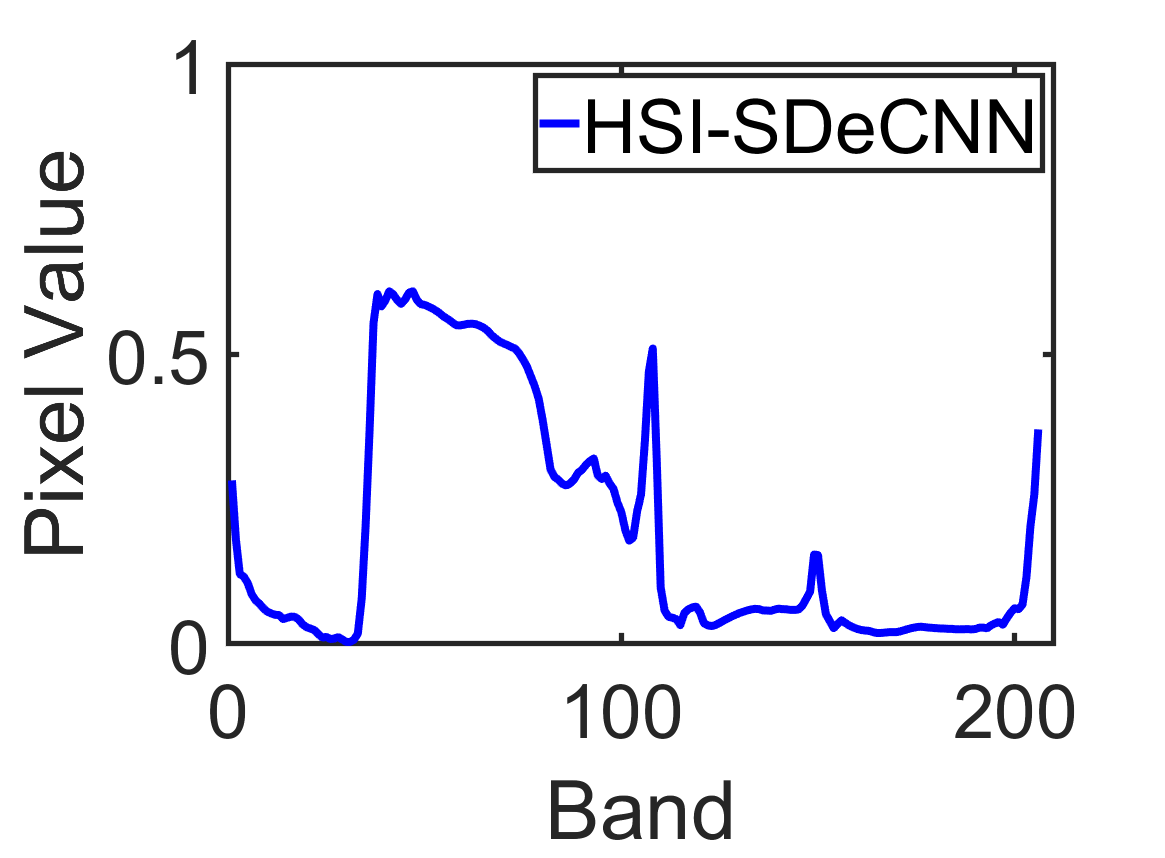}}
\end{minipage}%
\hspace{0.1mm}
\begin{minipage}[b]{0.132\textwidth}
    \centering
    \subfigure[(c)]{
    \includegraphics[width=\textwidth,height=0.75\textwidth]{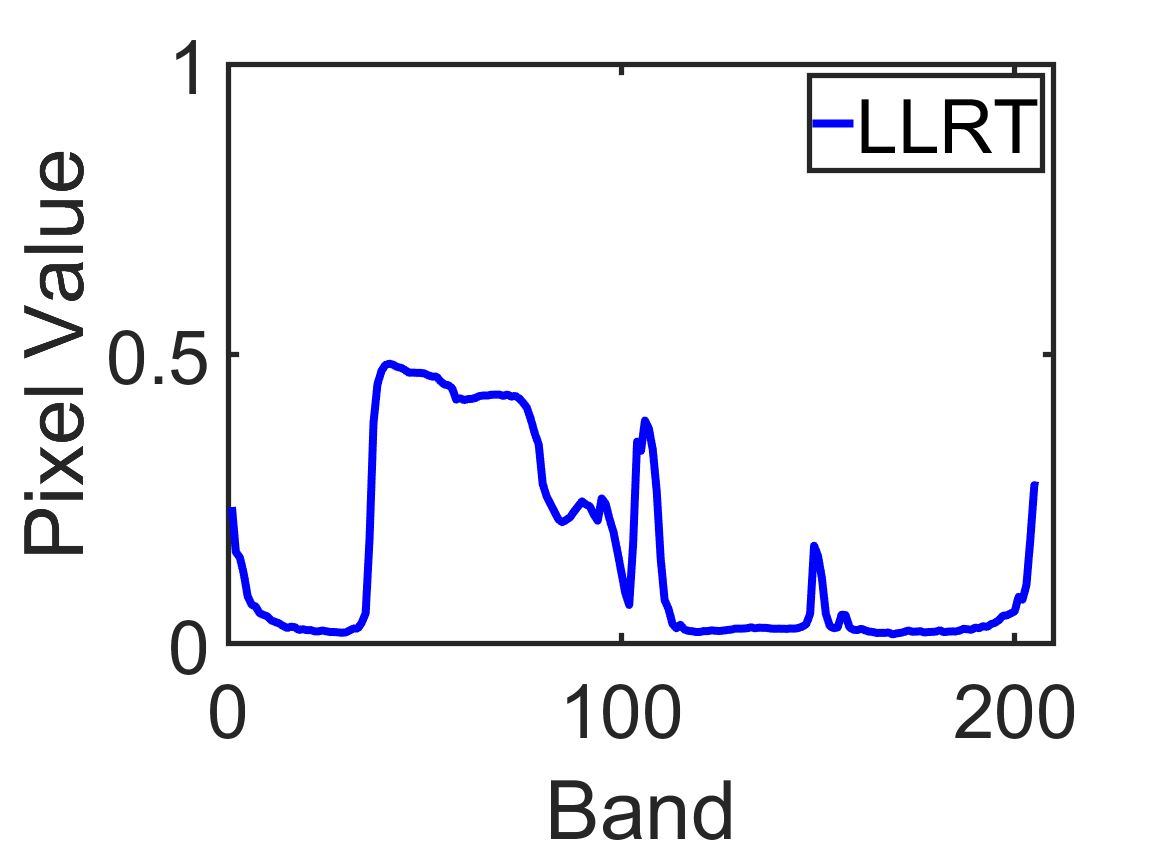}}\\
    \subfigure[(i)]{
    \includegraphics[width=\textwidth,height=0.75\textwidth]{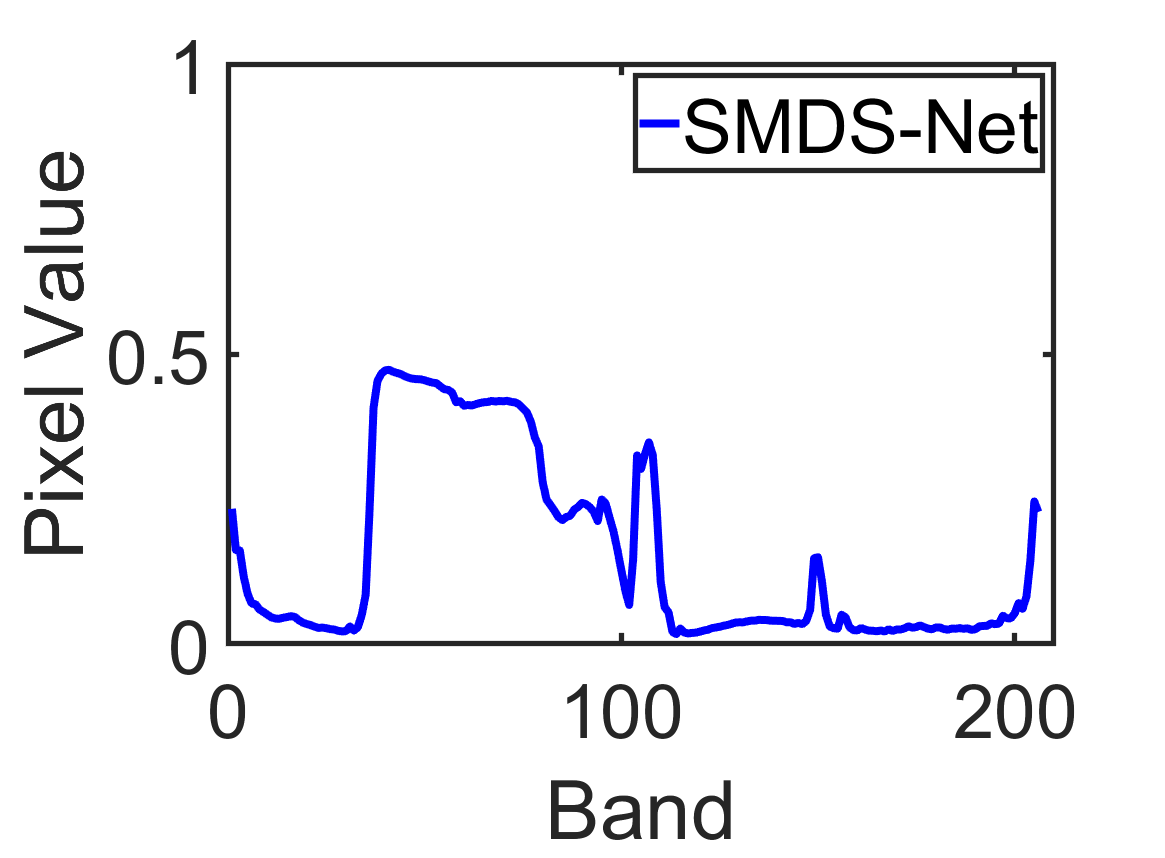}}
\end{minipage}%
\hspace{0.1mm}
\begin{minipage}[b]{0.132\textwidth}
    \centering
    \subfigure[(d)]{
    \includegraphics[width=\textwidth,height=0.75\textwidth]{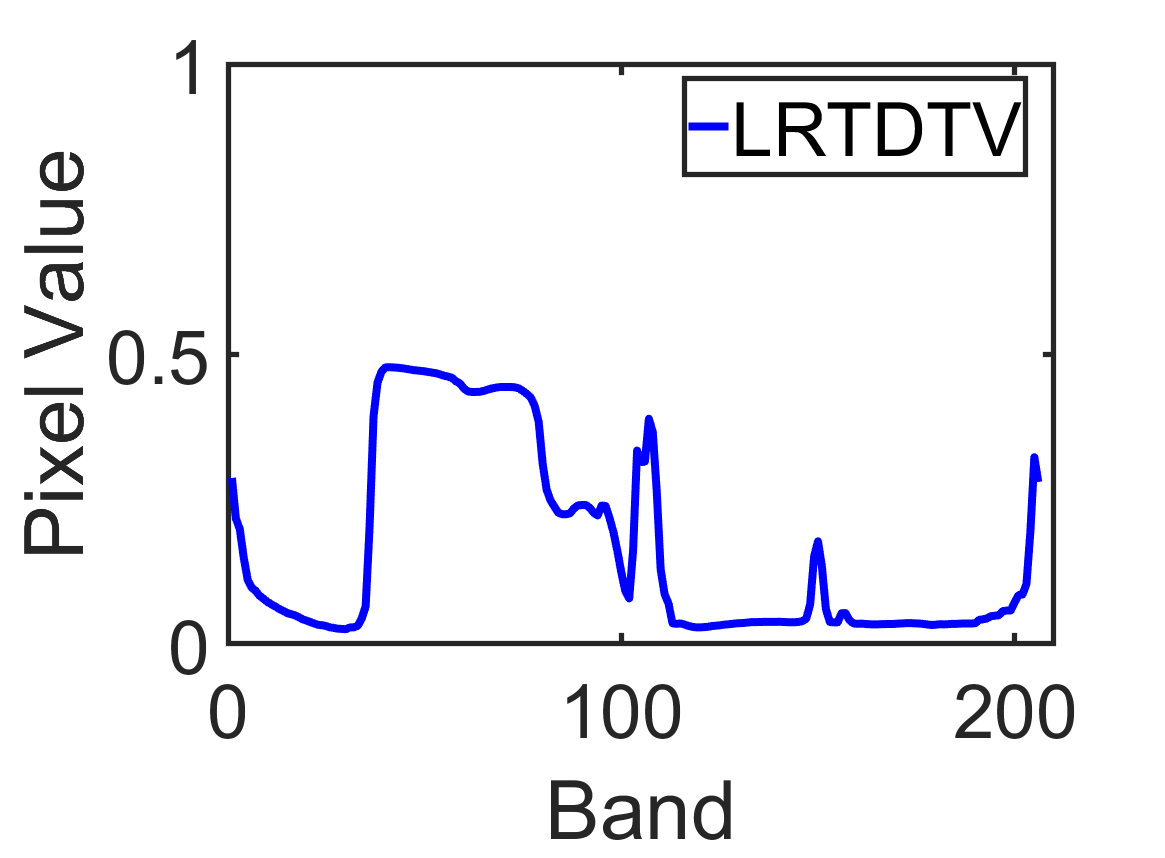}}\\
    \subfigure[(j)]{
    \includegraphics[width=\textwidth,height=0.75\textwidth]{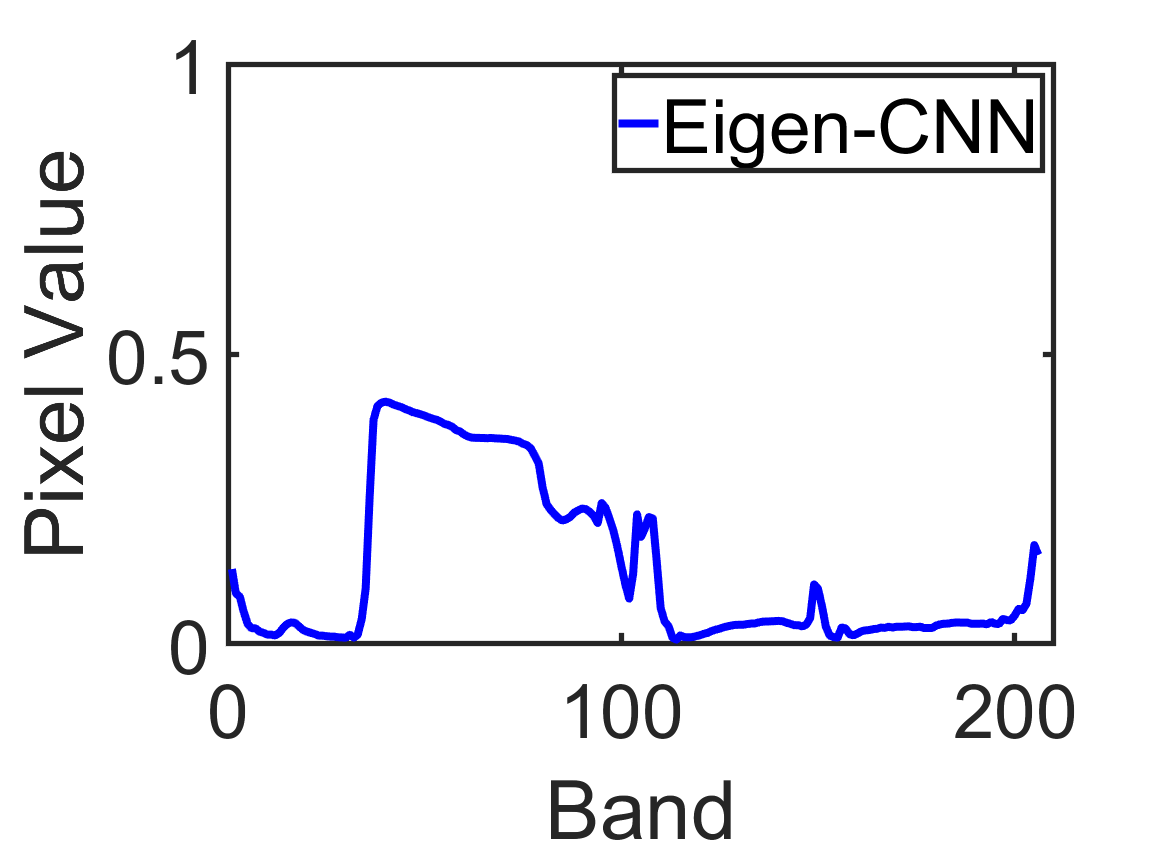}}
\end{minipage}%
\hspace{0.1mm}
\begin{minipage}[b]{0.132\textwidth}
    \centering
    \subfigure[(e)]{
    \includegraphics[width=\textwidth,height=0.75\textwidth]{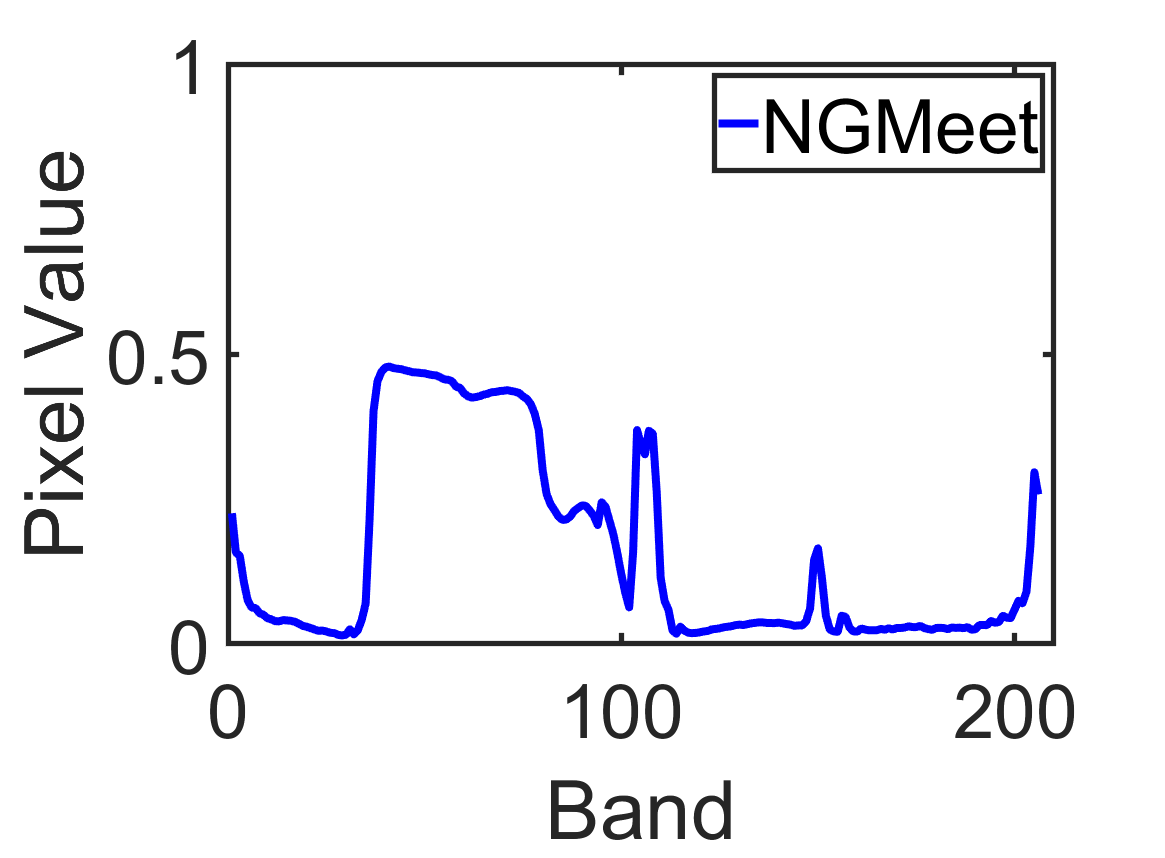}}\\
    \subfigure[(k)]{
    \includegraphics[width=\textwidth,height=0.75\textwidth]{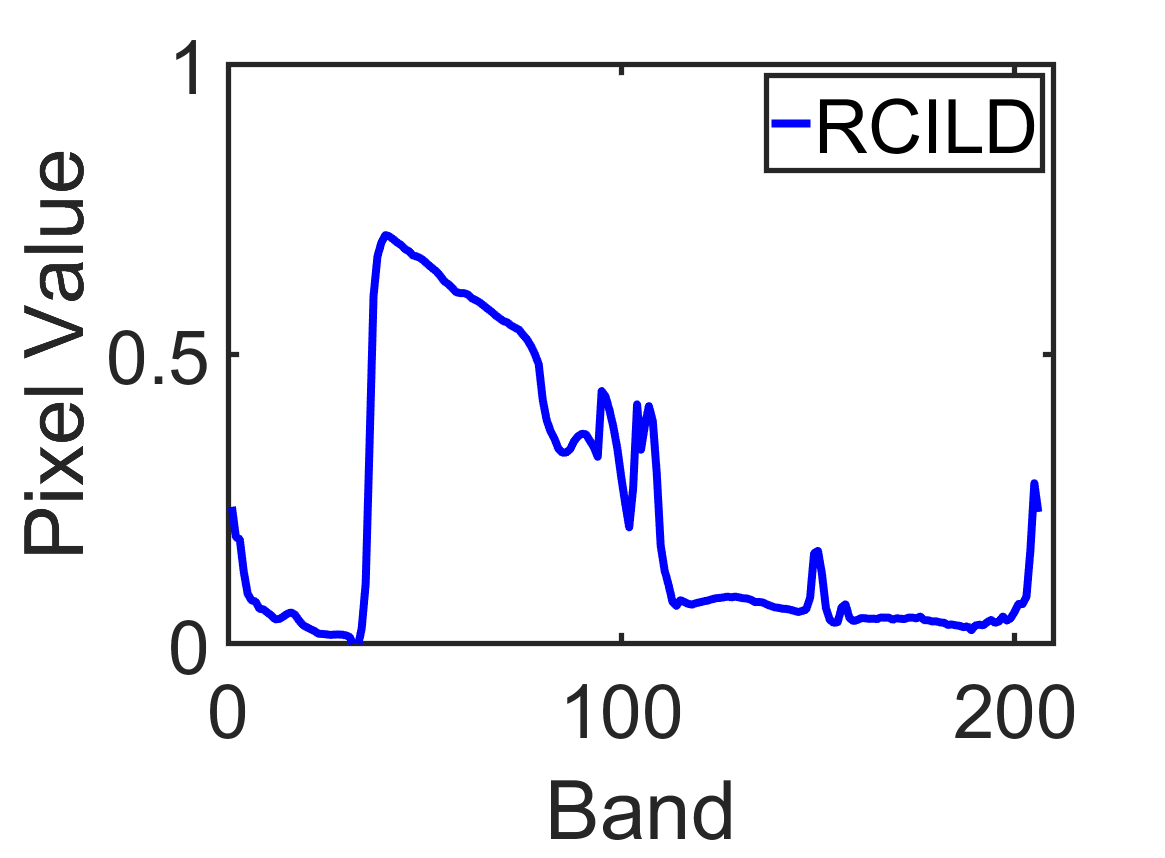}}
\end{minipage}%
\hspace{0.1mm}
\begin{minipage}[b]{0.132\textwidth}
    \centering
    \subfigure[(f)]{
    \includegraphics[width=\textwidth,height=0.75\textwidth]{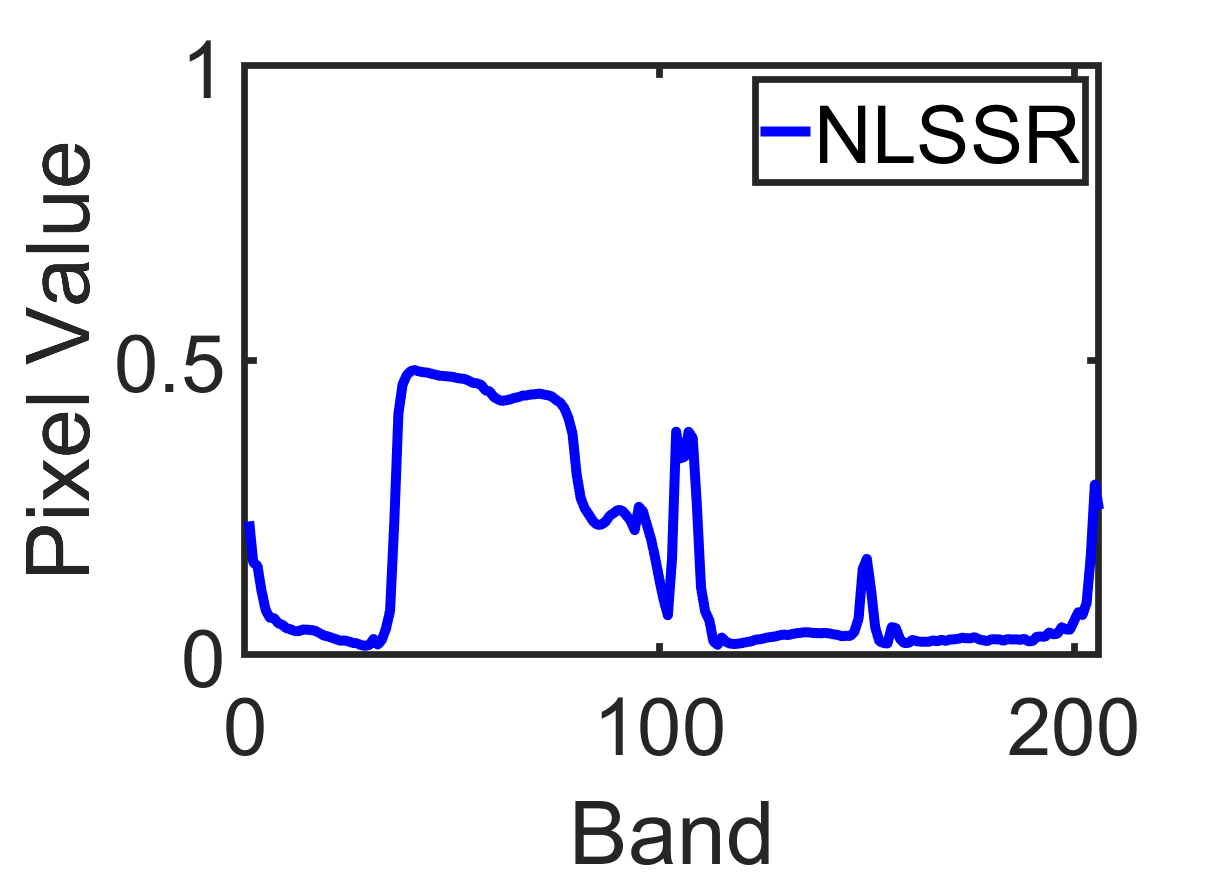}}\\
    \subfigure[(l)]{
    \includegraphics[width=\textwidth,height=0.75\textwidth]{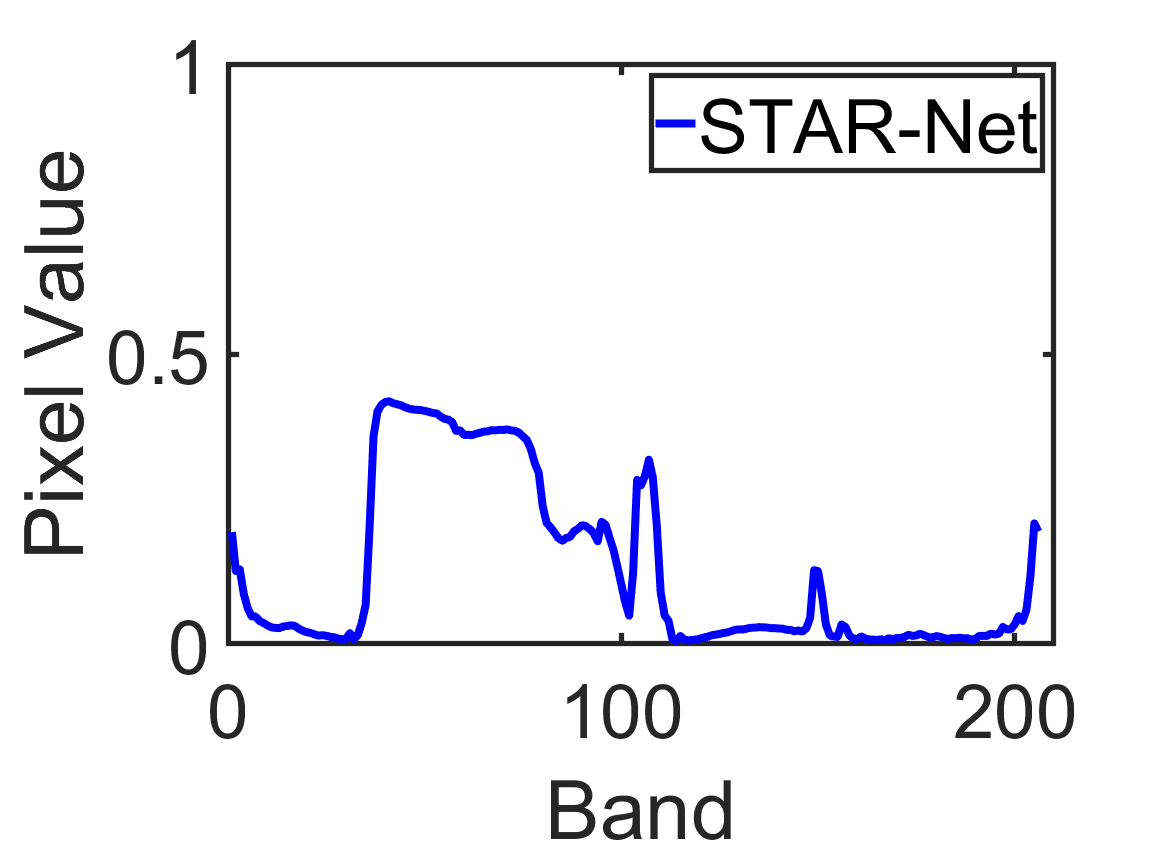}}
\end{minipage}%
\hspace{0.1mm}
\begin{minipage}[b]{0.132\textwidth}
    \centering
    \subfigure[(g)]{
    \includegraphics[width=\textwidth,height=0.75\textwidth]{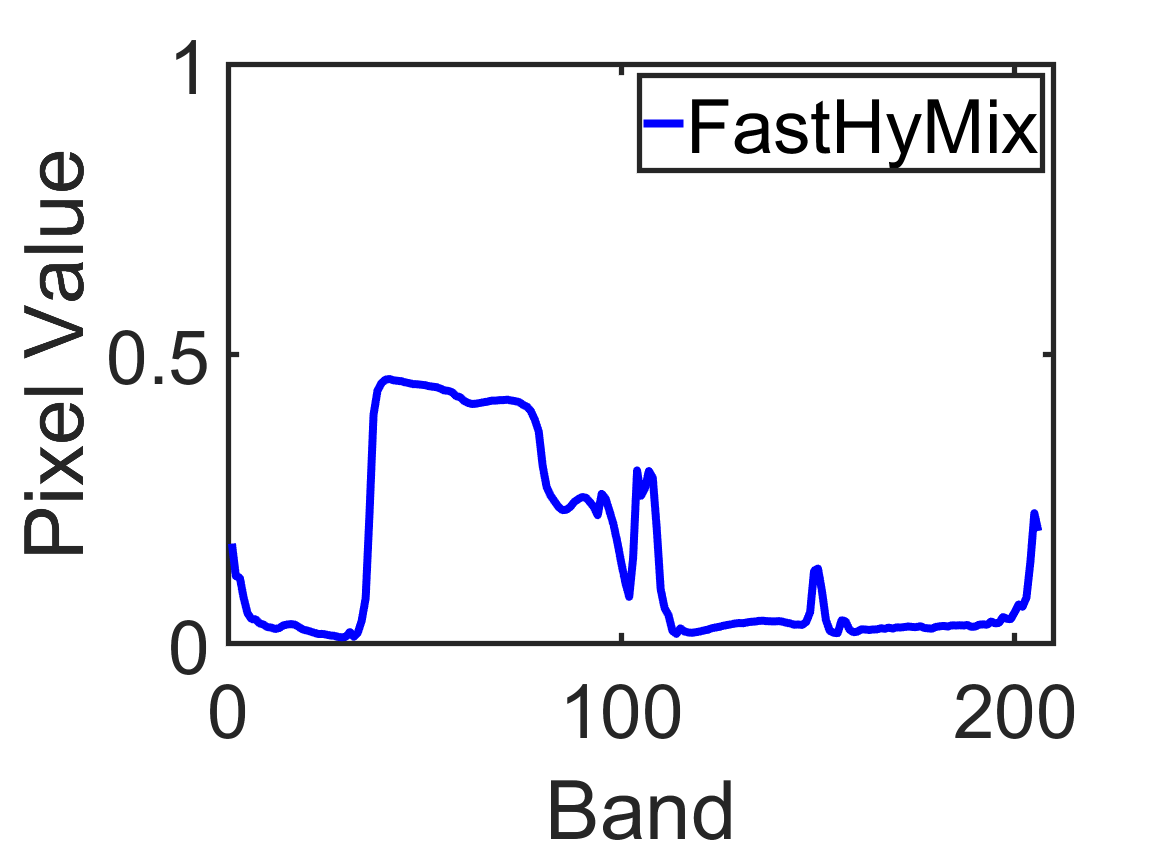}}\\
    \subfigure[(m)]{
    \includegraphics[width=\textwidth,height=0.75\textwidth]{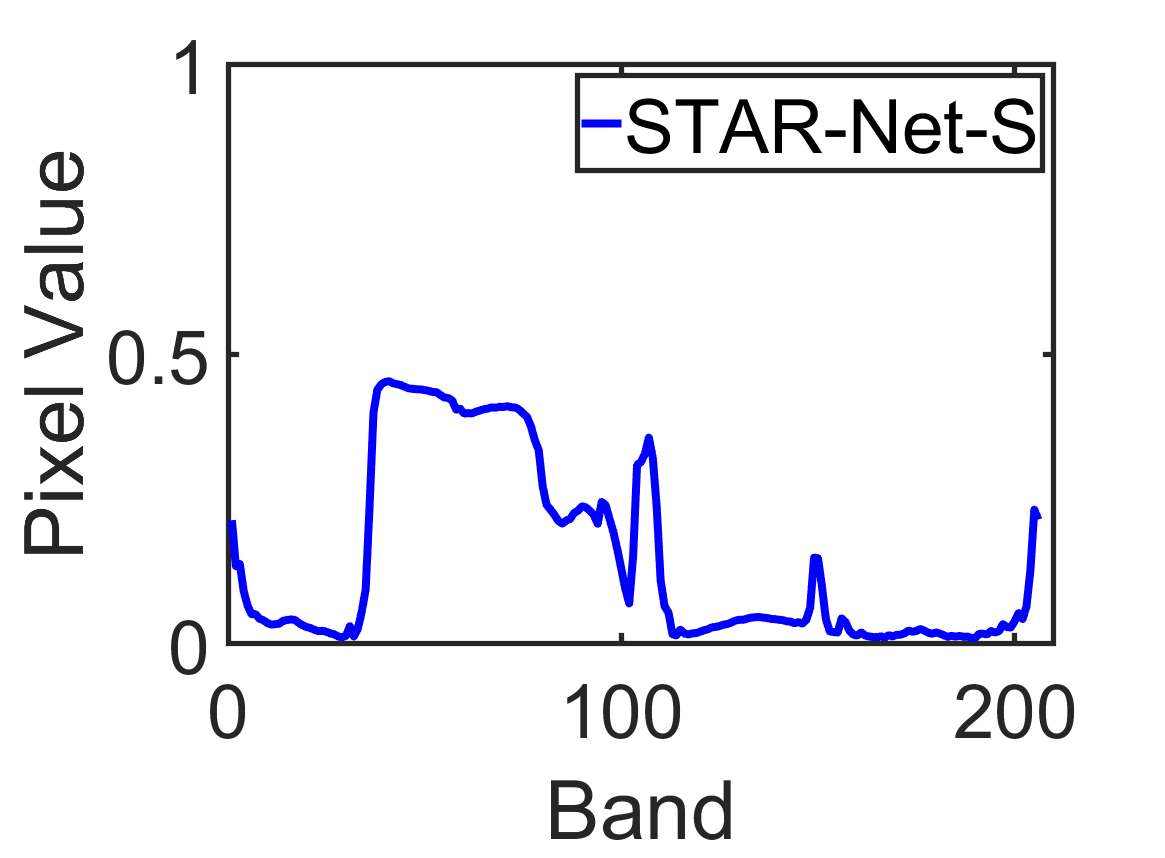}}
\end{minipage}%
\vskip-0.2cm
\caption{The denoising results of pixel (100, 100) on Indian Pines. (a) Noisy, (b) BM4D, (c) LLRT, (d) LRTDTV, (e) NGMeet, (f) NLSSR, (g) FastHyMix, (h) HSI-SDeCNN, (i) SMDS-Net, (j) Eigen-CNN, (k) RCILD, (l) STAR-Net, (m) STAR-Net-S.}
 \label{indianpixel}
\end{figure*}

\subsubsection{Experiments on Indian Pines}

Similarly, denoising experiments are conducted on the real-world Indian Pines dataset, with the denoising results presented in Figure \ref{indian}.
Although NLSSR, FastHyMix, and HSI-SDeCNN can remove noise, they cause distortion and loss of the original features. 
While SMDS-Net, Eigen-CNN, and RCILD preserve certain details, the bottom-right close-up reveals that their effectiveness is still inferior to that of STAR-Net and STAR-Net-S.
In contrast, STAR-Net and STAR-Net-S exhibit superior denoising performance, effectively preserving finer details after the denoising process.
It can be seen from Figure \ref{indianpixel} that although all methods produce relatively smooth curves, SMDS-Net, STAR-Net, and STAR-Net-S show better spectral continuity.


\begin{table*}[t]
\centering
\renewcommand\arraystretch{1.2} 
\caption{Classification results of all methods on Indian Pines. The top two values are marked as \textcolor[rgb]{1.00,0.00,0.00}{red} and \textcolor[rgb]{0.00,0.00,1.00}{blue}.}\label{classtable}
\vskip-0.2cm
\setlength{\tabcolsep}{3.2 pt}
\resizebox{\textwidth}{!}{
\begin{tabular}{cccccccccccccc}
\toprule
{Class}       & {Noisy}            & \makecell[c]{BM4D\\}   & \makecell[c]{LLRT\\}                       & \makecell[c]{LRTDTV\\} & \makecell[c]{NGMeet\\}& \makecell[c]{NLSSR\\}&\makecell[c]{FastHy\\Mix} & \makecell[c]{HSI-SDe\\CNN}  & \makecell[c]{SMDS-\\Net}    &\makecell[c]{Eigen-\\CNN}   & \makecell[c]{RCILD\\}         & \makecell[c]{STAR-\\Net}        & \makecell[c]{STAR-\\Net-S}   \\ \hline 
Alfalfa                      & 0.889                  & 1.000 & 1.000       & 0.875  & 1.000 & 1.000 & 1.000 & 0.889      & 0.889  & 1.000 & 1.000 & 0.889    & 1.000          \\
Corn-notill                  & 0.877                  & 0.949 & 0.966       & 0.776  & 0.858 & 0.857 & 0.859 & 0.957      & 0.922  & 0.806 & 0.905 & 0.948    & 0.978          \\
Corn-mintill                 & 0.897                  & 0.926 & 0.953       & 0.878  & 0.920 & 0.850 & 0.889 & 0.982      & 0.903  & 0.855 & 0.967 & 0.964    & 0.994          \\
Corn                         & 0.796                  & 0.913 & 1.000       & 0.756  & 0.891 & 0.816 & 0.860 & 0.978      & 0.911  & 0.857 & 0.917 & 0.917    & 1.000          \\
Grass-pasture                & 0.922                  & 0.942 & 1.000       & 0.939  & 0.940 & 0.957 & 0.893 & 0.980      & 0.970  & 0.900 & 0.959 & 0.960    & 0.960          \\
Grass-trees                  & 0.973                  & 1.000 & 1.000       & 0.906  & 0.924 & 0.935 & 0.953 & 1.000      & 0.993  & 0.940 & 0.973 & 0.993    & 0.993          \\
Grass-pasture-mowed          & 1.000                  & 0.833 & 1.000       & 1.000  & 0.833 & 1.000 & 1.000 & 1.000      & 1.000  & 0.800 & 0.800 & 1.000    & 1.000          \\
Hay-windrowed                & 0.990                  & 1.000 & 1.000       & 0.979  & 1.000 & 1.000 & 0.990 & 1.000      & 1.000  & 0.990 & 0.990 & 1.000    & 1.000          \\
Oats                         & 0.800                  & 1.000 & 1.000       & 0.800  & 0.800 & 1.000 & 0.667 & 0.800      & 0.800  & 0.600 & 0.800 & 1.000    & 1.000          \\
Soybean-notill               & 0.886                  & 0.925 & 0.964       & 0.805  & 0.841 & 0.860 & 0.847 & 0.944      & 0.918  & 0.827 & 0.894 & 0.972    & 0.949          \\
Soybean-mintill              & 0.869                  & 0.913 & 0.945       & 0.783  & 0.834 & 0.813 & 0.860 & 0.953      & 0.935  & 0.823 & 0.896 & 0.937    & 0.957          \\
Soybean-clean                & 0.921                  & 0.950 & 0.967       & 0.735  & 0.742 & 0.762 & 0.908 & 0.952      & 0.946  & 0.882 & 0.943 & 0.906    & 0.991          \\
Wheat                        & 0.976                  & 0.976 & 1.000       & 0.976  & 1.000 & 1.000 & 0.976 & 1.000      & 1.000  & 0.976 & 0.976 & 1.000    & 1.000          \\
Woods                        & 0.950                  & 0.969 & 1.000       & 0.939  & 0.977 & 0.977 & 0.961 & 0.984      & 0.996  & 0.922 & 0.969 & 0.996    & 0.996          \\
Buildings-Grass-Trees-Drives & 0.882                  & 0.931 & 0.987       & 0.881  & 0.953 & 0.932 & 0.896 & 0.960      & 0.987  & 0.857 & 0.913 & 0.987    & 1.000          \\
Stone-Steel-Towers           & 1.000                  & 1.000 & 1.000       & 1.000  & 1.000 & 1.000 & 1.000 & 1.000      & 1.000  & 1.000 & 1.000 & 1.000    & 1.000          \\ \hline
OA                           & 0.883                  & 0.944 & \textcolor[rgb]{0.00,0.00,1.00}{0.972} & 0.843  & 0.886 & 0.923 & 0.896 & 0.967      & 0.949  & 0.866 & 0.931 & 0.961    & \textcolor[rgb]{1.00,0.00,0.00}{0.978} \\
AA                           & 0.895                  & 0.952 & \textcolor[rgb]{0.00,0.00,1.00}{0.986} & 0.877  & 0.907 & 0.880 & 0.910 & 0.961      & 0.948  & 0.877 & 0.931 & 0.967    & \textcolor[rgb]{1.00,0.00,0.00}{0.989} \\
Kappa                        & 0.867                  & 0.936 & \textcolor[rgb]{0.00,0.00,1.00}{0.968} & 0.820  & 0.869 & 0.861 & 0.881 & 0.963      & 0.942  & 0.847 & 0.921 & 0.955    & \textcolor[rgb]{1.00,0.00,0.00}{0.974} \\ \bottomrule
\end{tabular}}
\end{table*}

\begin{figure*}[t]
  \centering
    \subfigure[]{
    \includegraphics[width=0.619 in,height=0.619 in]{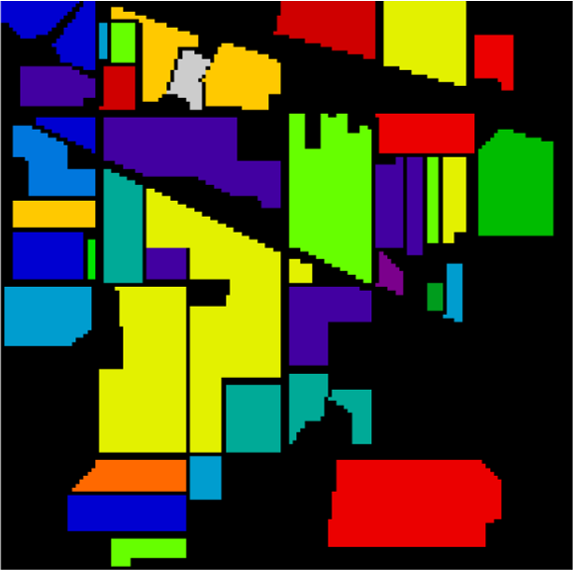}\label{ROC-b21}}
    \subfigure[]{
    \includegraphics[width=0.619 in,height=0.619 in]{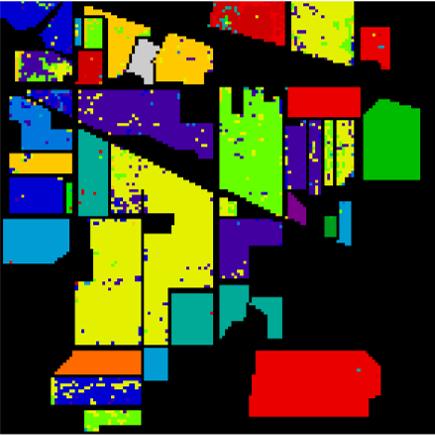}\label{ROC-b22}}
    \subfigure[]{
    \includegraphics[width=0.619 in,height=0.619 in]{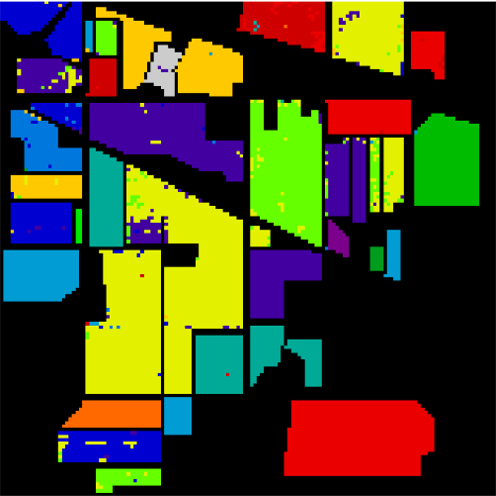}\label{ROC-b22}}
    \subfigure[]{
    \includegraphics[width=0.619 in,height=0.619 in]{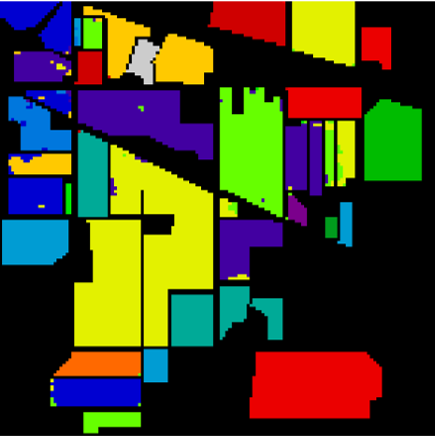}\label{ROC-b22}}
    \subfigure[]{
    \includegraphics[width=0.619 in,height=0.619 in]{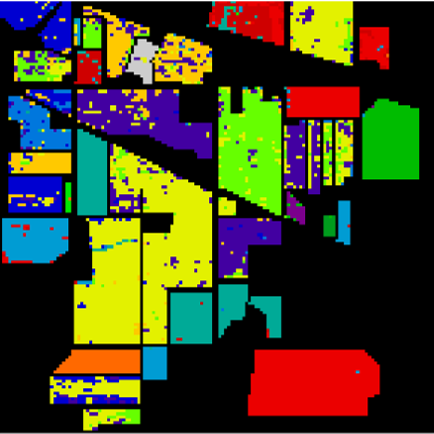}\label{ROC-b22}}
    \subfigure[]{
    \includegraphics[width=0.619 in,height=0.619 in]{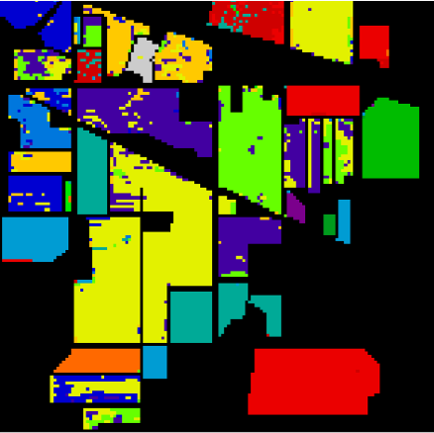}\label{ROC-b22}}
    \subfigure[]{
    \includegraphics[width=0.619 in,height=0.619 in]{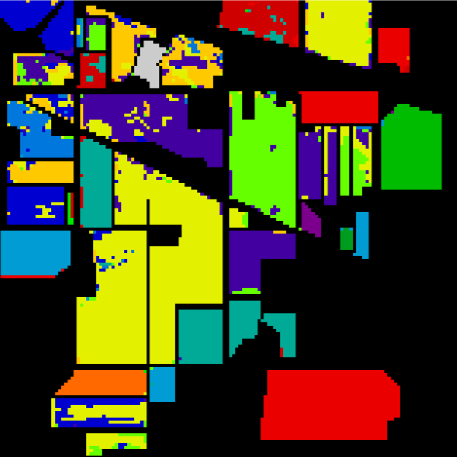}\label{ROC-b22}}
    \subfigure[]{
    \includegraphics[width=0.619 in,height=0.619 in]{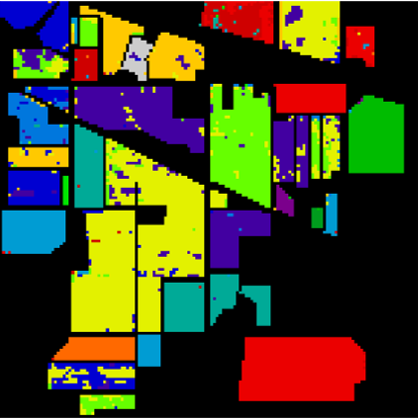}\label{ROC-b22}}
    \subfigure[]{
    \includegraphics[width=0.619 in,height=0.619 in]{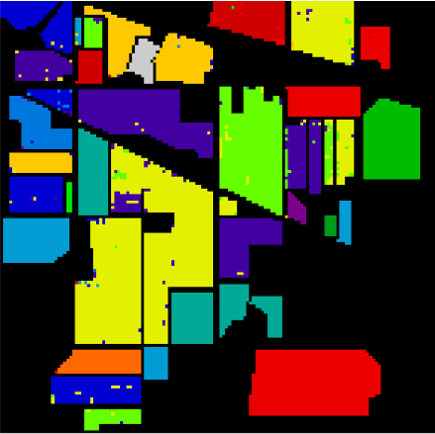}\label{ROC-b22}}
    \subfigure[]{
    \includegraphics[width=0.619 in,height=0.619 in]{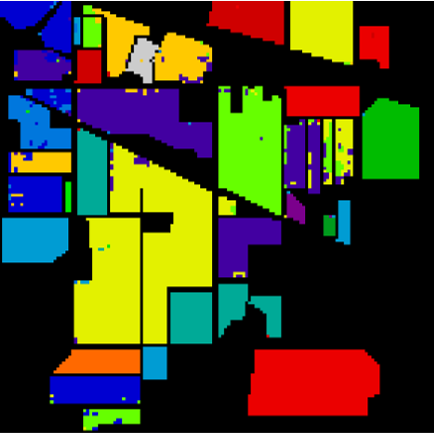}\label{ROC-b23}}
    \subfigure[]{
    \includegraphics[width=0.619 in,height=0.619 in]{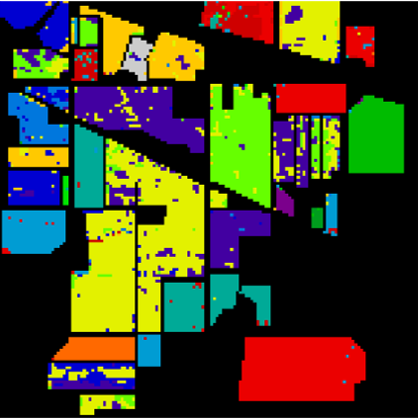}\label{ROC-b22}}
    \subfigure[]{
    \includegraphics[width=0.619 in,height=0.619 in]{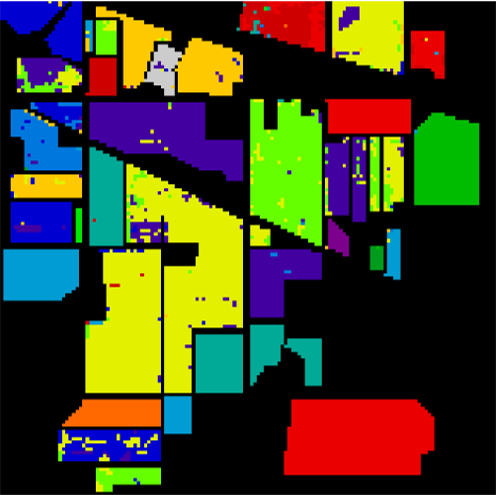}\label{ROC-b22}}
    \subfigure[]{
    \includegraphics[width=0.619 in,height=0.619 in]{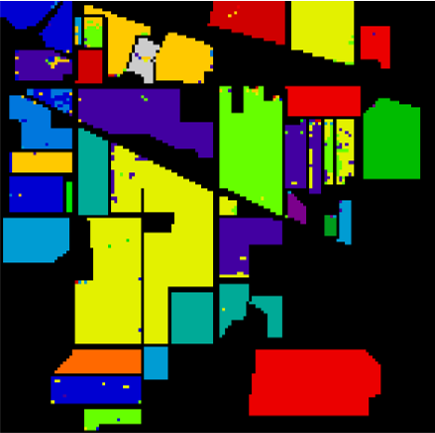}\label{ROC-b24}}
    \subfigure[]{
    \includegraphics[width=0.619 in,height=0.619 in]{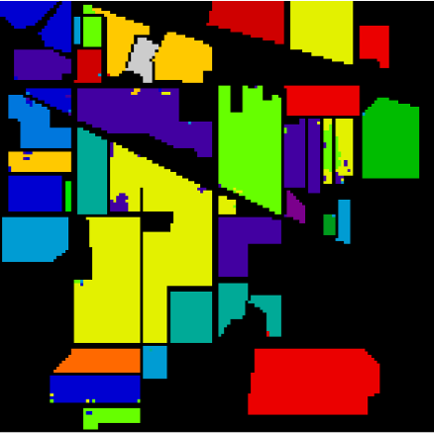}\label{ROC-b24}}
         \vskip-0.2cm
  \caption{Classification results obtained by all methods on Indian Pines. (a) Ground truth, (b) Noisy, (c) BM4D, (d) LLRT, (e) LRTDTV, (f) NGMeet, (g) NLSSR, (h) FastHyMix, (i) HSI-SDeCNN, (j) SMDS-Net, (k) Eigen-CNN, (l) RCILD, (m) STAR-Net, (n) STAR-Net-S.}\label{class}
\end{figure*}

To further compare the influence of the denoising performance of various methods on downstream tasks in the real dataset, we conduct classification experiments using the support vector machine (SVM).
Table \ref{classtable} summarizes the classification accuracy of 16 categories, along with the overall accuracy (OA), average accuracy (AA), and kappa coefficient (Kappa).
Additionally, Figure \ref{class} shows the corresponding classification visualization results.
Obviously, STAR-Net-S is closest to the ground truth, followed by LLRT and STAR-Net.
This further validates the effectiveness of our proposed STAR-Net and STAR-Net-S on downstream tasks.

\subsection{Discussion}

\subsubsection{Number of Parameters}

The number of parameters for each deep learning-based methods discussed in this paper is listed in Table \ref{Parameter}.
Among them, HSI-SDeCNN and RCILD have more parameters, while our proposed STAR-Net and STAR-Net-S have relatively fewer parameters, which may be attributed to the utilization of the model to assist in representing the network.


\begin{table}[t]
\centering
\renewcommand\arraystretch{1.2}
\caption{Number of parameters. The top two values are marked as \textcolor[rgb]{1.00,0.00,0.00}{red} and \textcolor[rgb]{0.00,0.00,1.00}{blue}.}\label{Parameter}
\vskip-0.2cm
\setlength{\tabcolsep}{4 pt}
\resizebox{\textwidth}{!}{
\begin{tabular}{cccccccc}
\toprule
Method   &  \makecell[c]{FastHyMix}  & \makecell[c]{HSI-SDeCNN} & \makecell[c]{SMDS-Net}& \makecell[c]{Eigen-CNN}& \makecell[c]{RCILD}     & \makecell[c]{STAR-Net} & \makecell[c]{STAR-Net-S} \\ \hline
$\#$Parameters &$/$& 1892100    & \textcolor[rgb]{1.00,0.00,0.00}{5103} & $/$  & 2892288 &  \textcolor[rgb]{0.00,0.00,1.00}{27702}      & 28487      \\ \bottomrule
\end{tabular}}
\end{table}

\begin{figure}[t]
  \centering
    \subfigure[PSNR]{
    \includegraphics[width=1.1 in,height=0.825 in]{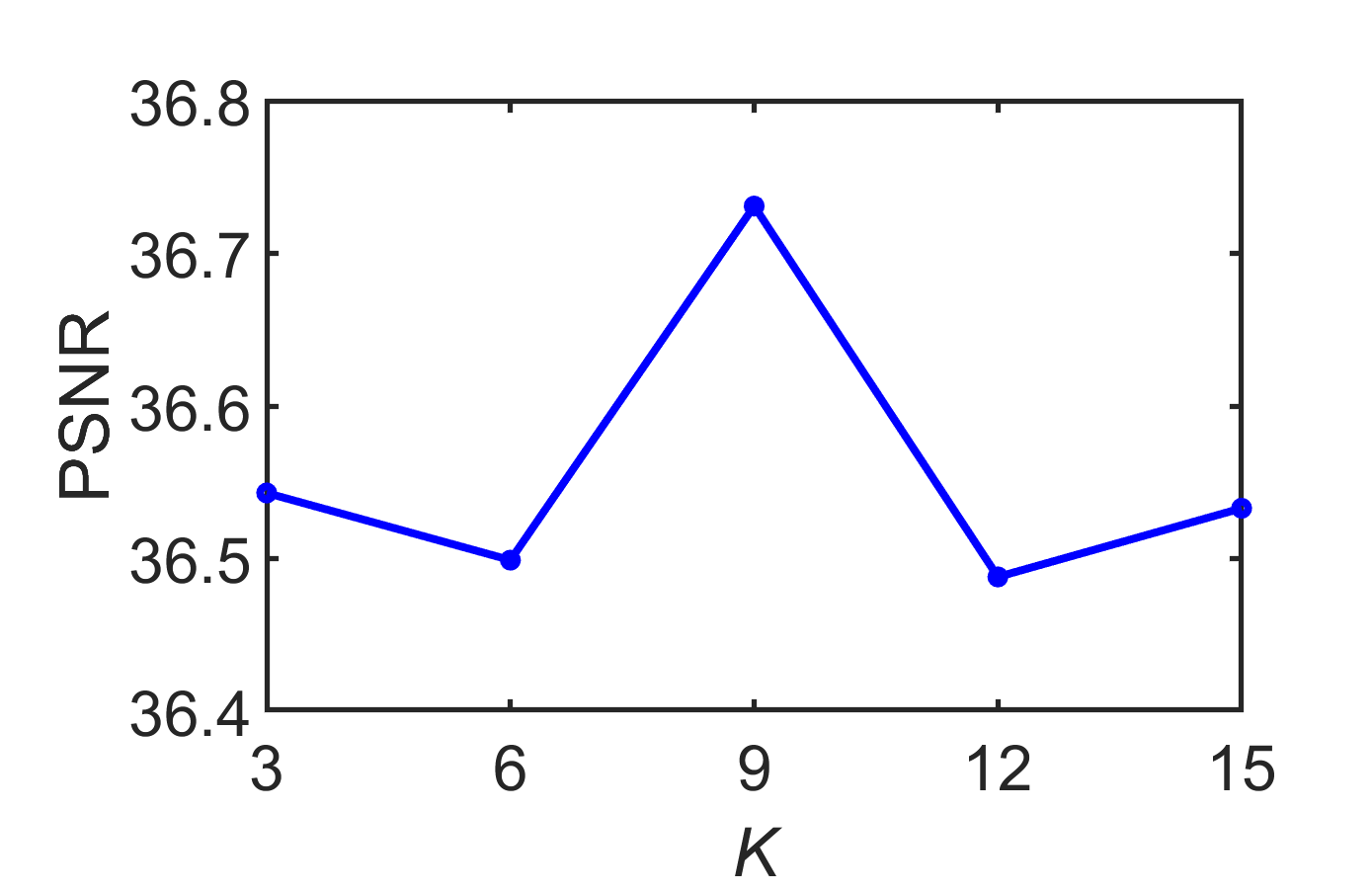}\label{ROC-b21}}
    \subfigure[SSIM]{
    \includegraphics[width=1.1 in,height=0.825 in]{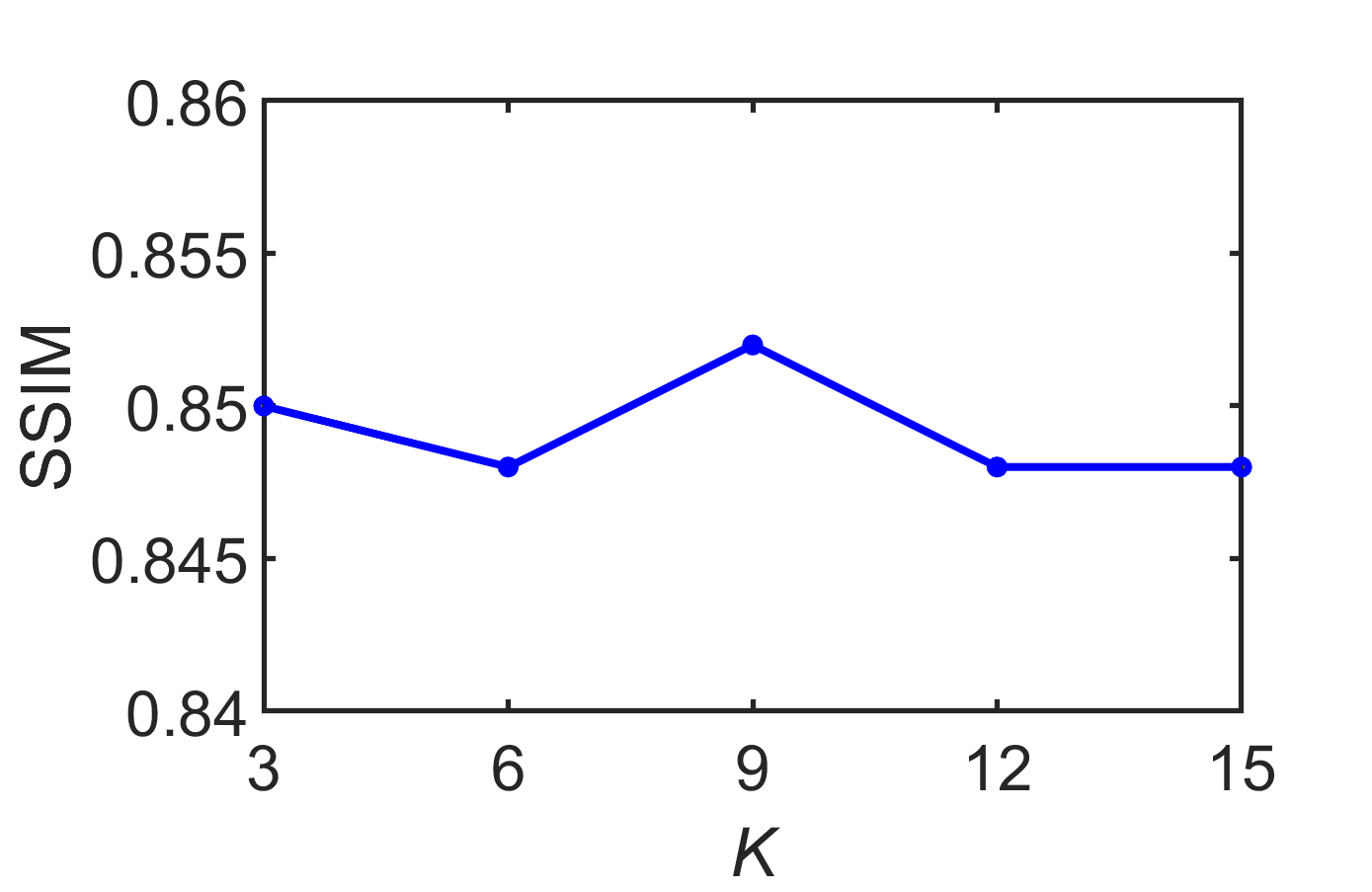}\label{ROC-b22}}
    \subfigure[SAM]{
    \includegraphics[width=1.1 in,height=0.825 in]{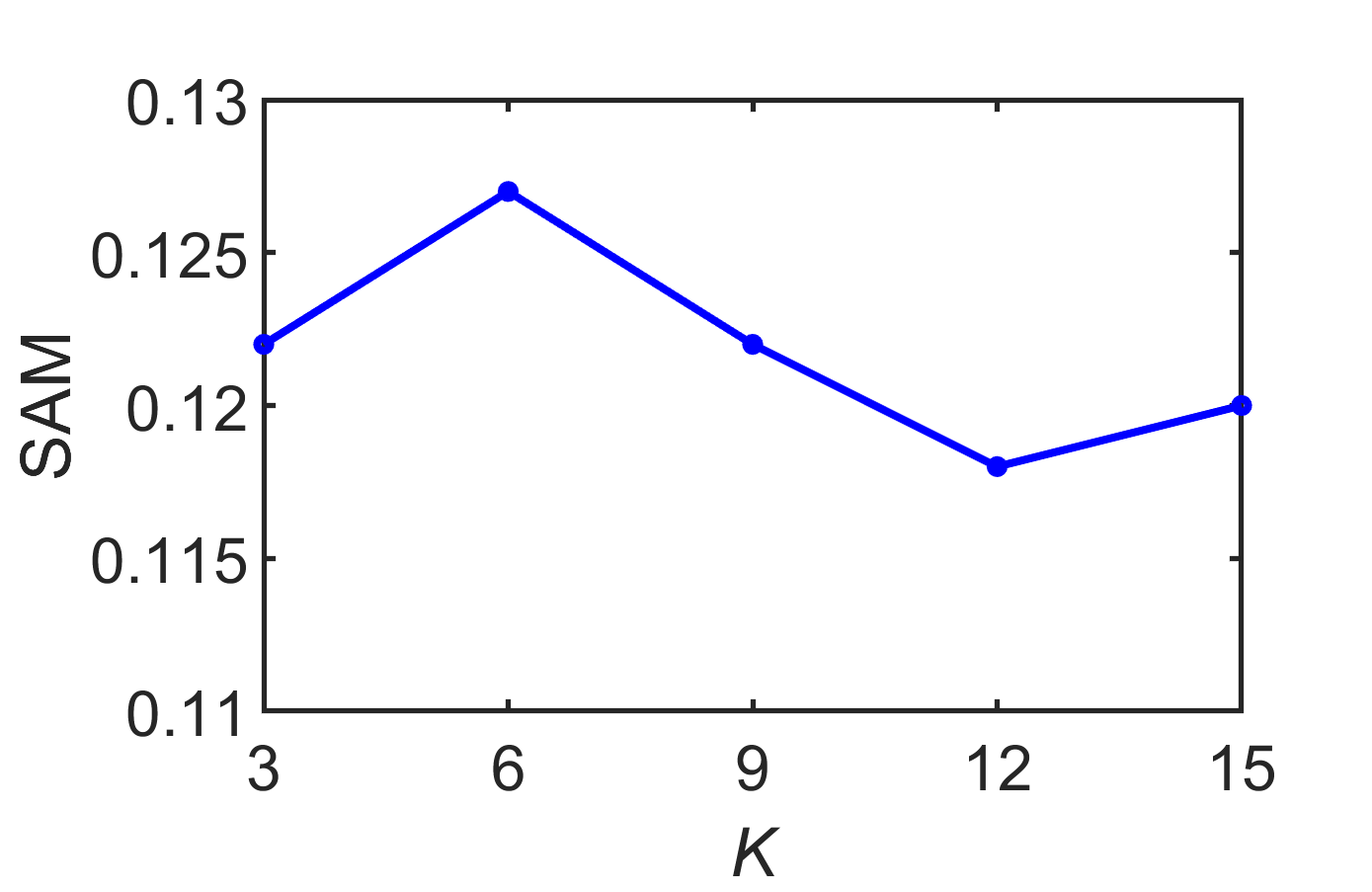}\label{ROC-b22}}
    \subfigure[$\#$Parameters]{
    \includegraphics[width=1.1 in,height=0.825 in]{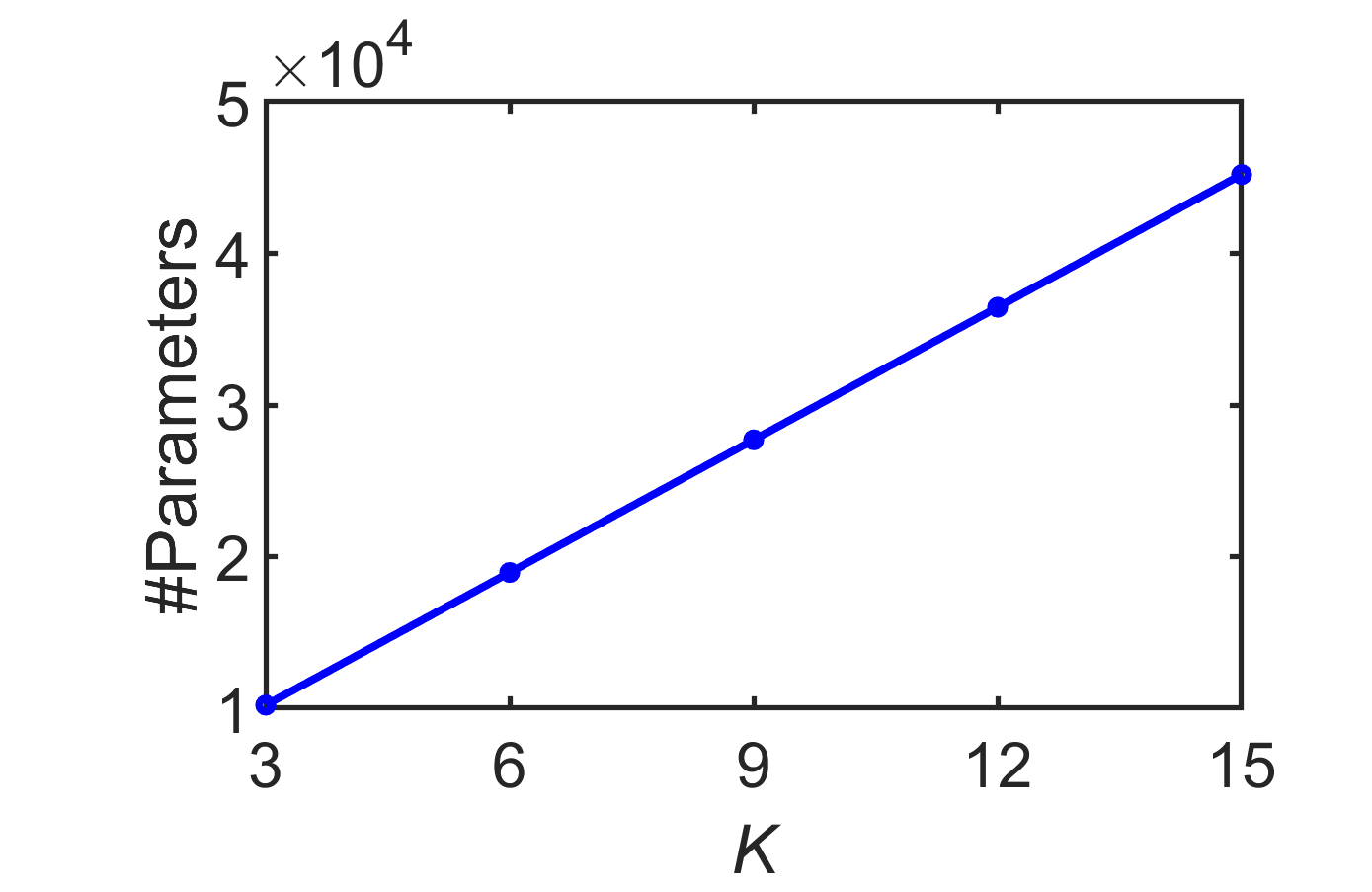}\label{ROC-b22}}
         \vskip-0.2cm
  \caption{Impact  of unrolling iteration $K$ of STAR-Net.}\label{k1}
\end{figure}

\begin{figure}[t]
  \centering
    \subfigure[PSNR]{
    \includegraphics[width=1.1 in,height=0.825 in]{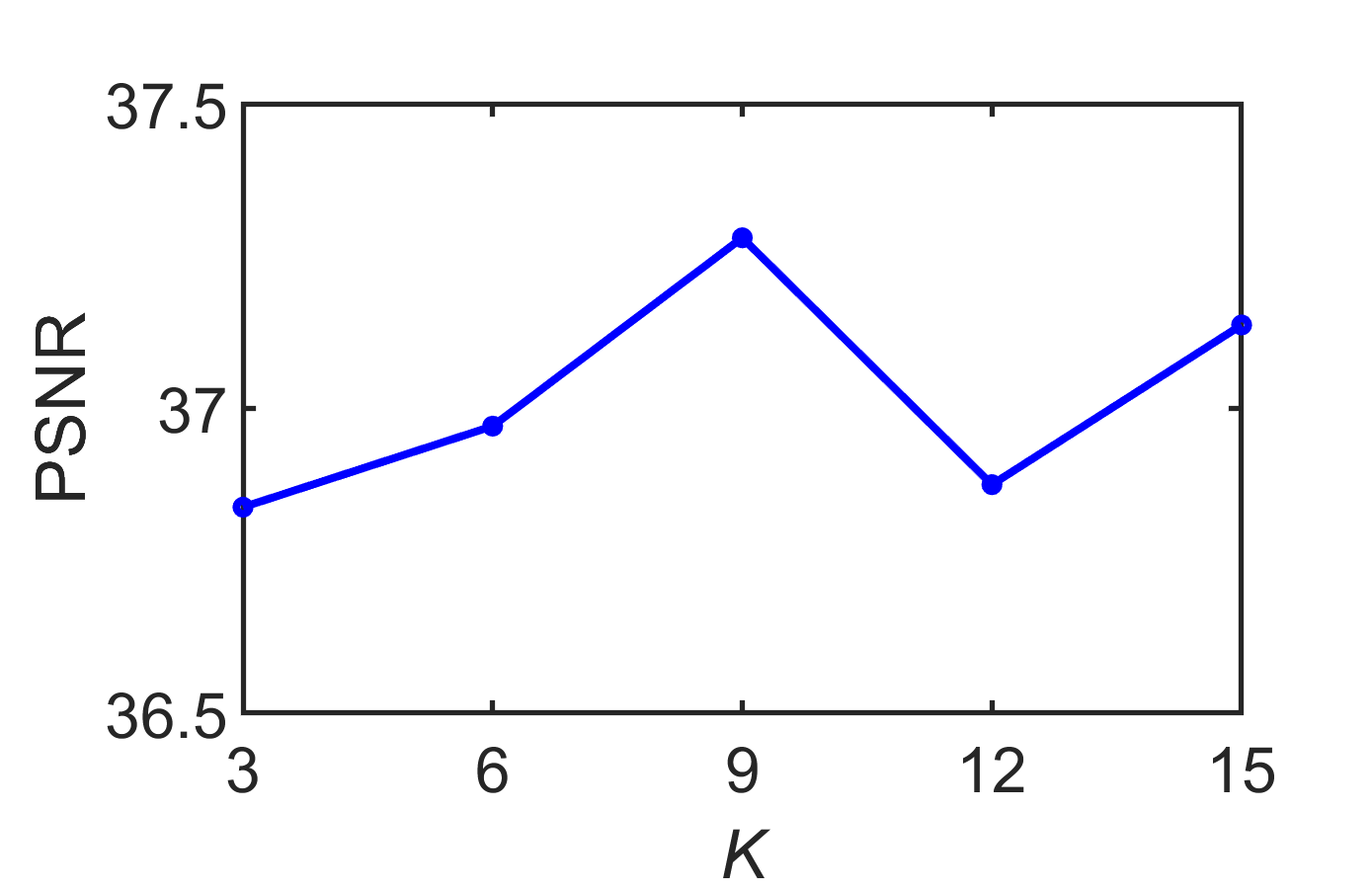}\label{ROC-b21}}
    \subfigure[SSIM]{
    \includegraphics[width=1.1 in,height=0.825 in]{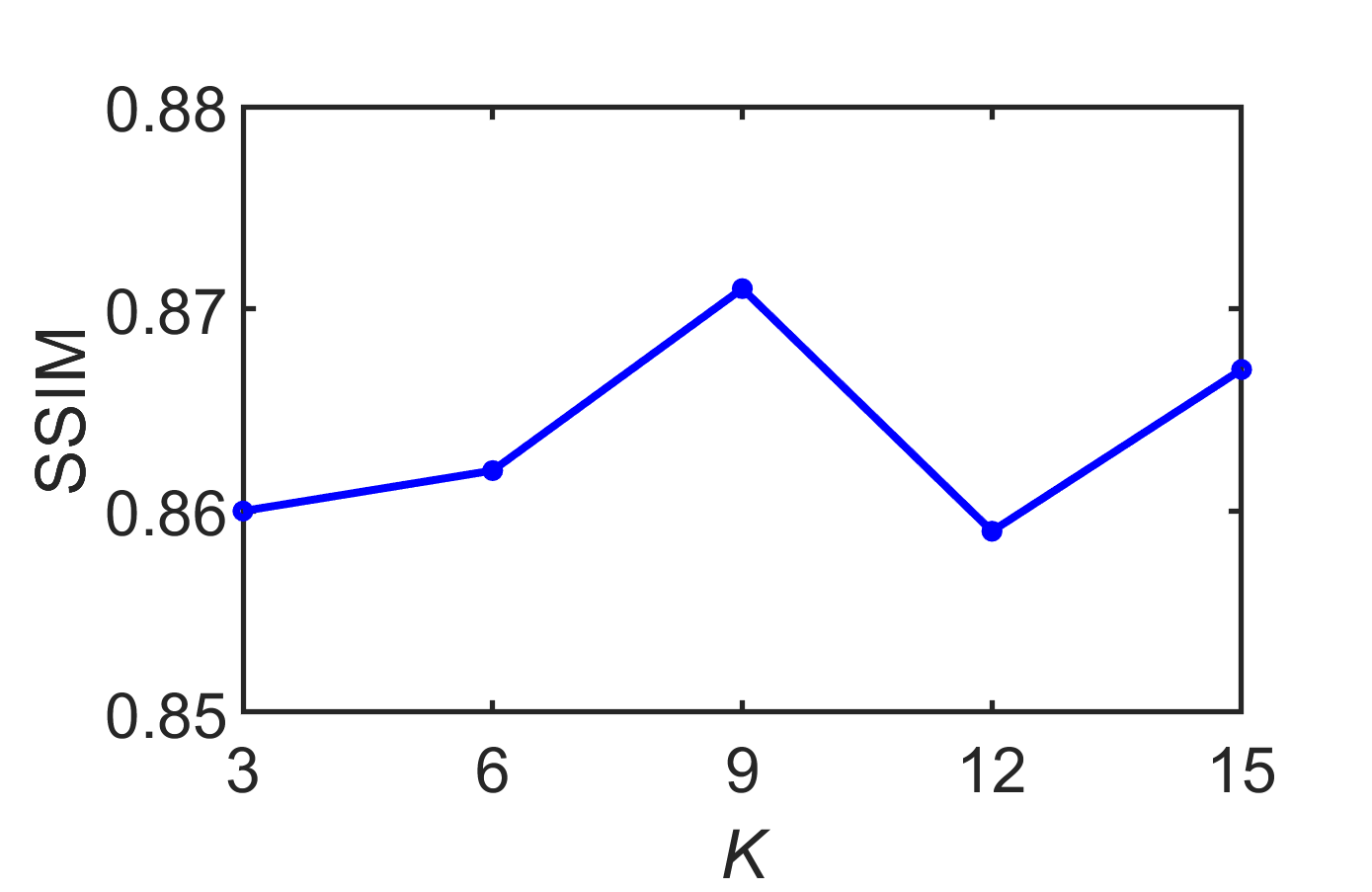}\label{ROC-b22}}
    \subfigure[SAM]{
    \includegraphics[width=1.1 in,height=0.825 in]{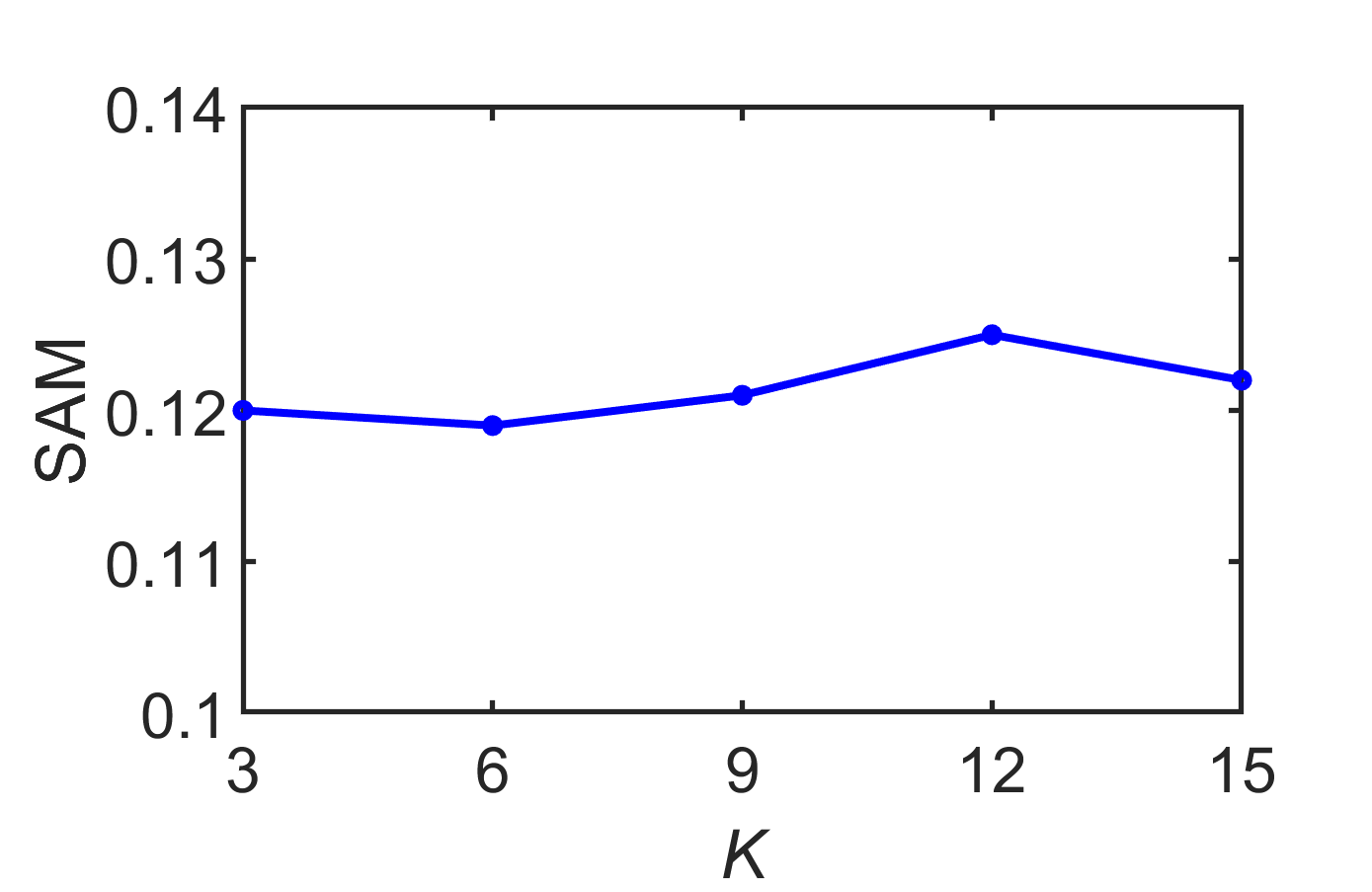}\label{ROC-b22}}
    \subfigure[$\#$Parameters]{
    \includegraphics[width=1.1 in,height=0.825 in]{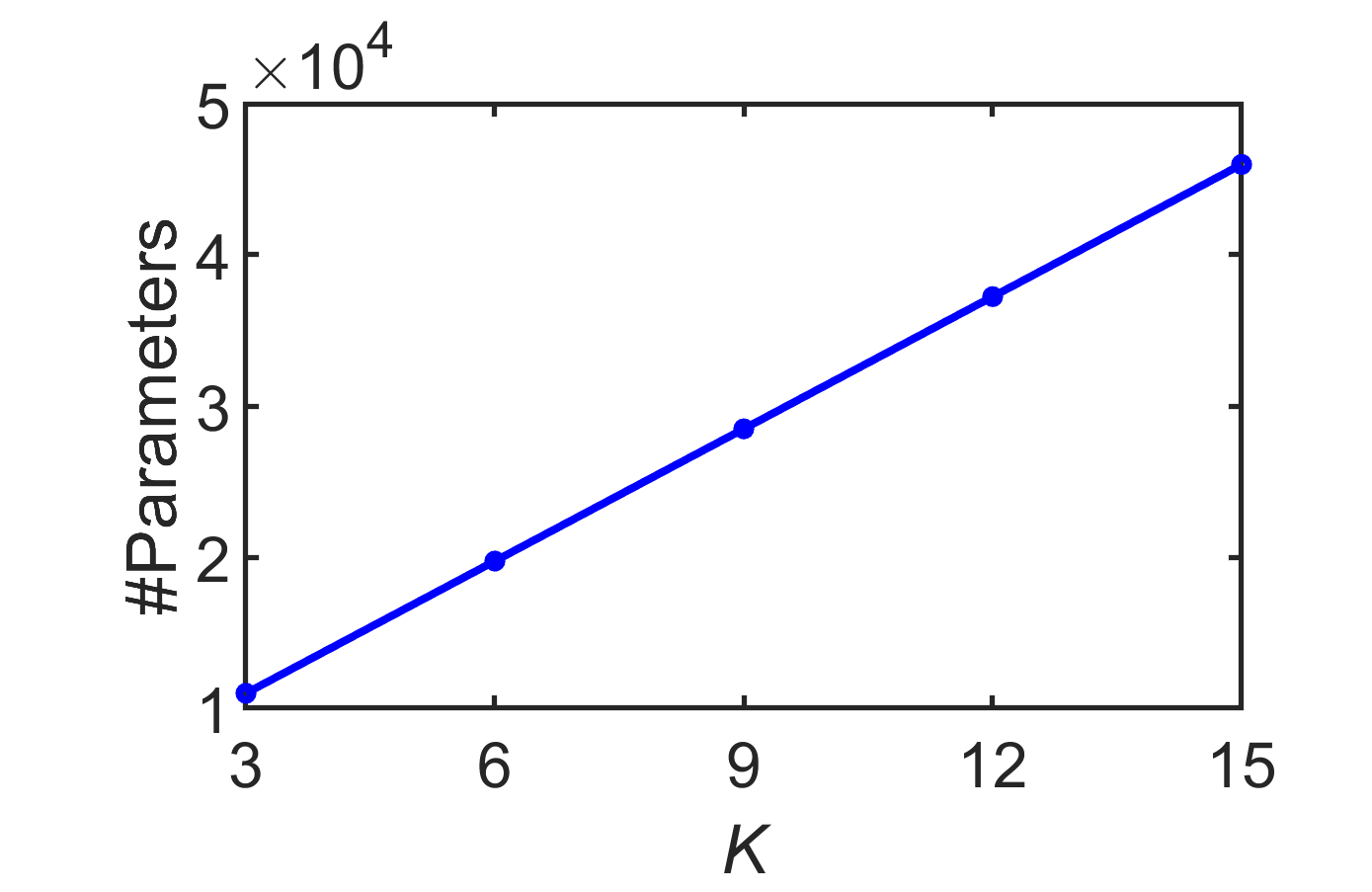}\label{ROC-b22}}
         \vskip-0.2cm
  \caption{Impact  of unrolling iteration $K$ of STAR-Net-S.}\label{k2}
\end{figure}

\begin{figure}[t]
  \centering
    \subfigure[PSNR]{
    \includegraphics[width=1.1 in,height=0.825 in]{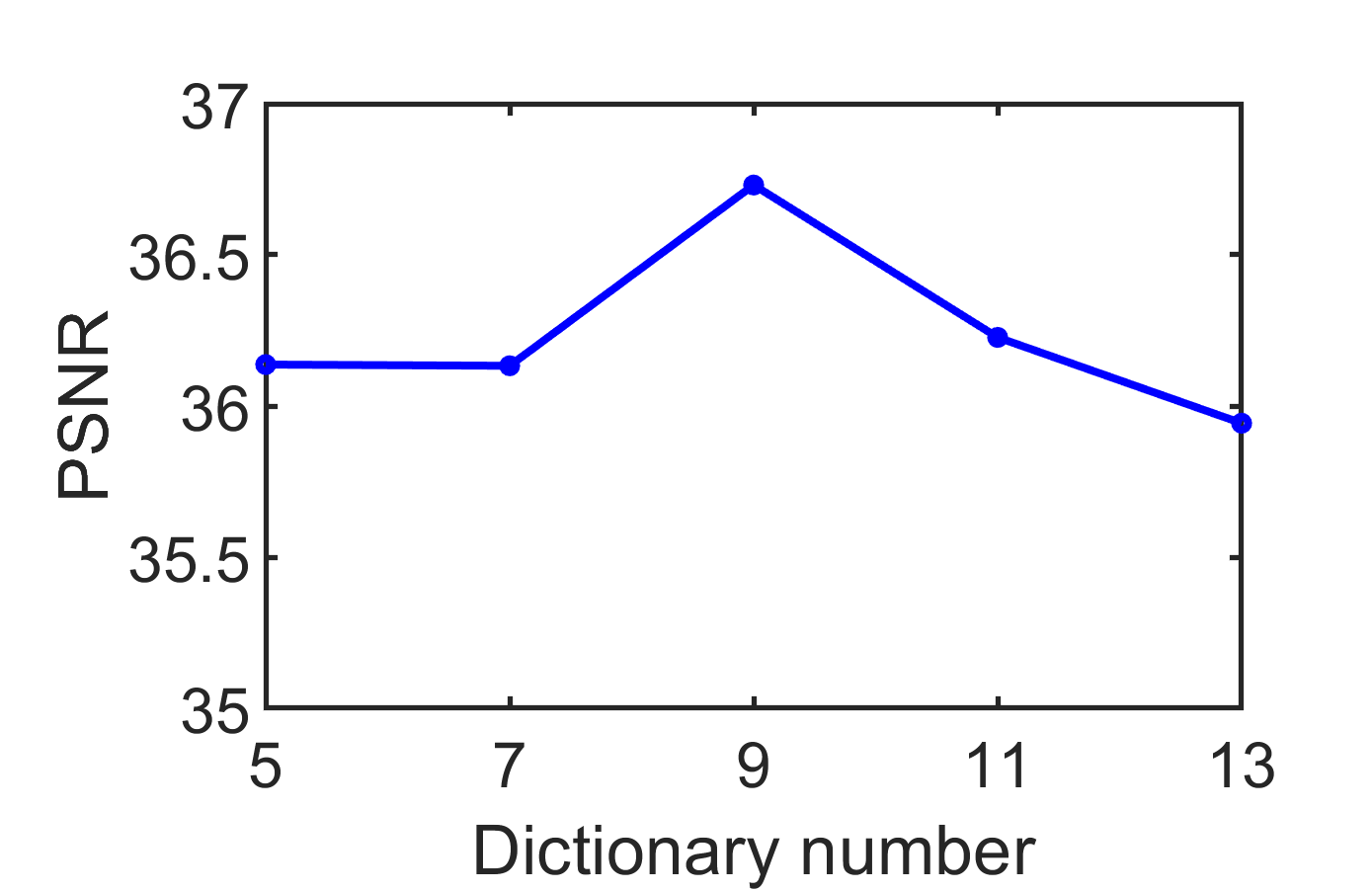}\label{ROC-b21}}
    \subfigure[SSIM]{
    \includegraphics[width=1.1 in,height=0.825 in]{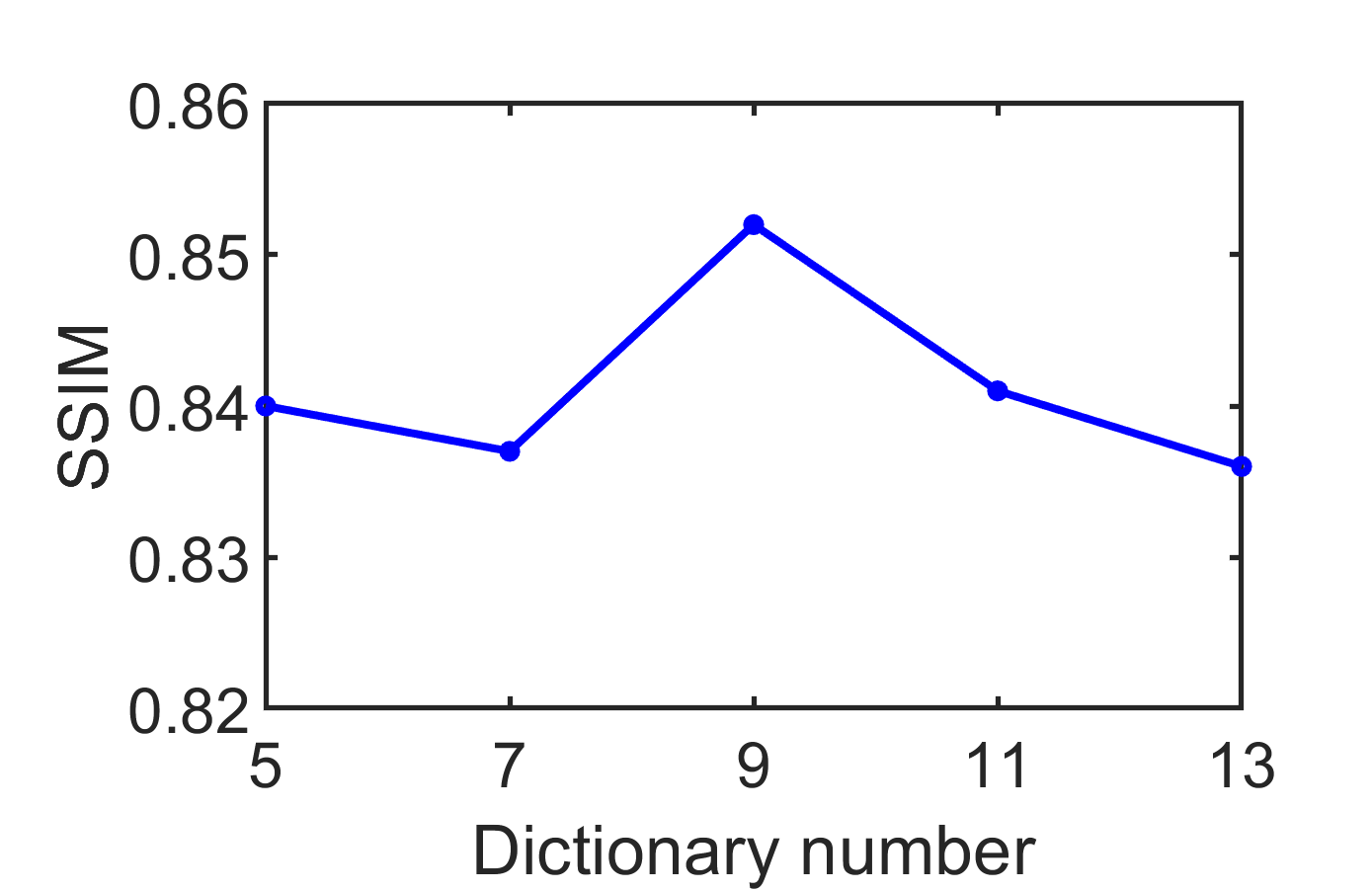}\label{ROC-b22}}
    \subfigure[SAM]{
    \includegraphics[width=1.1 in,height=0.825 in]{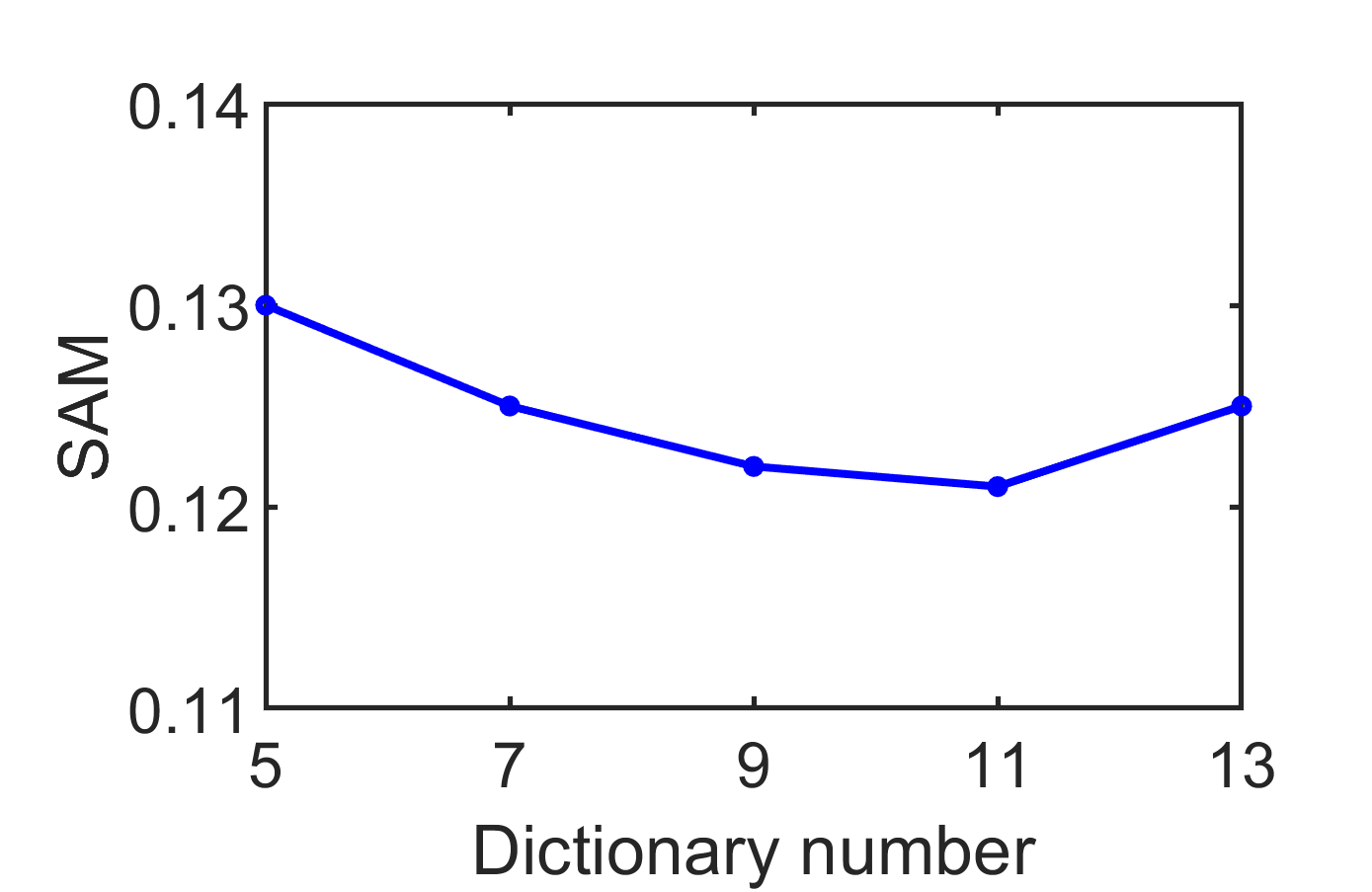}\label{ROC-b22}}
    \subfigure[$\#$Parameters]{
    \includegraphics[width=1.1 in,height=0.825 in]{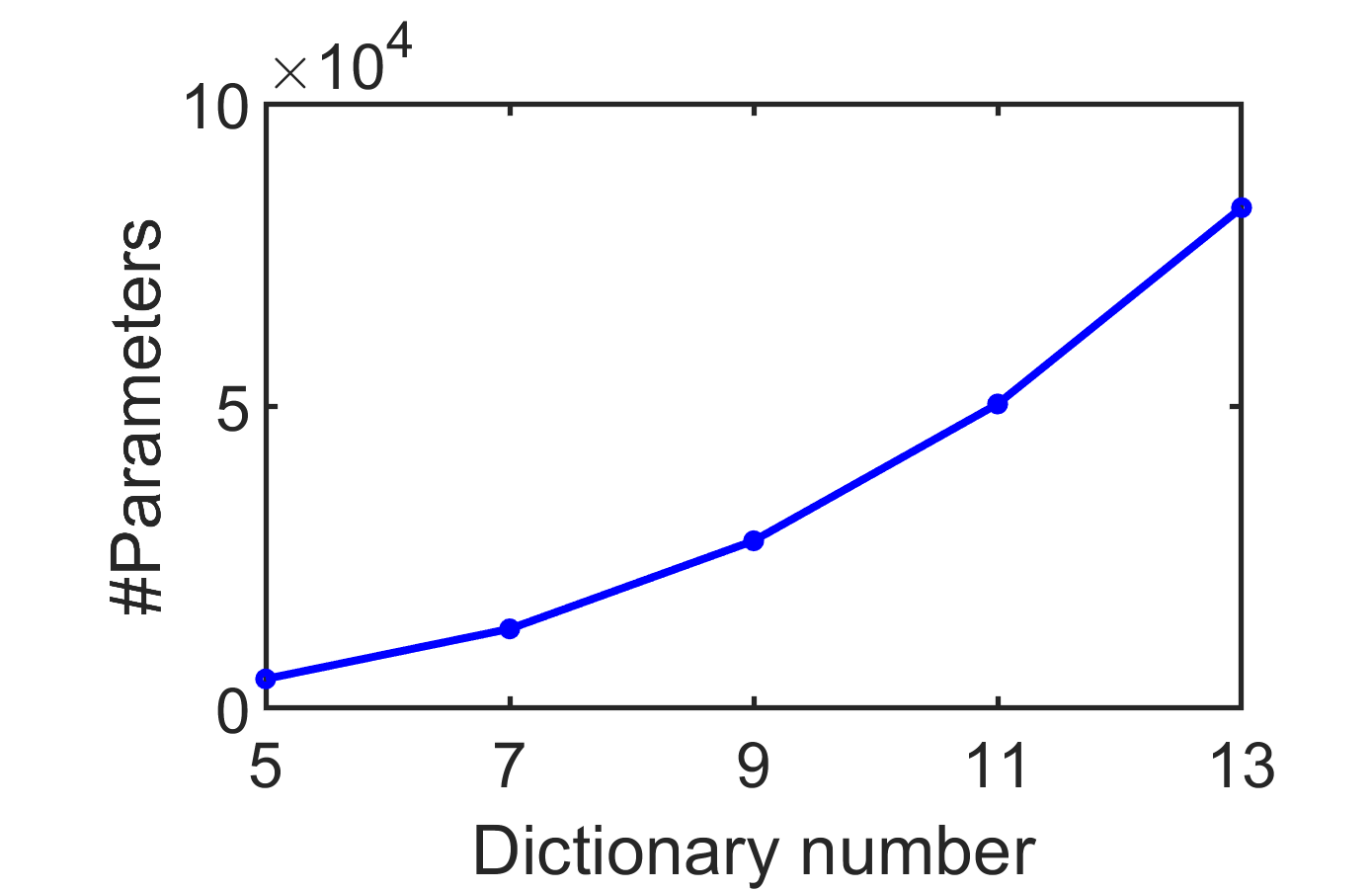}\label{ROC-b22}}
         \vskip-0.2cm
  \caption{Impact  of the dictionary number of STAR-Net.}\label{dic1}
\end{figure}

\begin{figure}[h!]
  \centering
    \subfigure[PSNR]{
    \includegraphics[width=1.1 in,height=0.825 in]{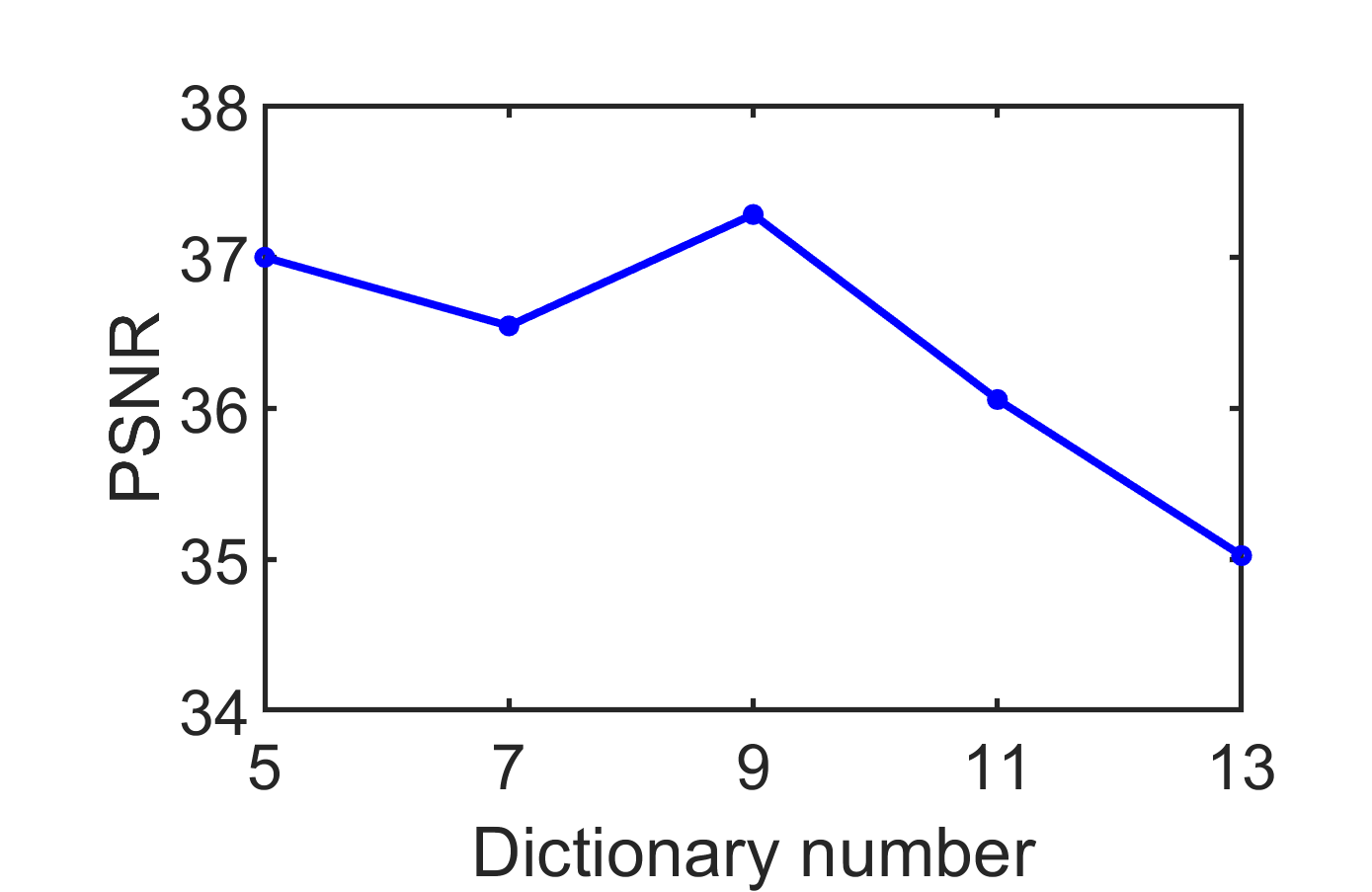}\label{ROC-b21}}
    \subfigure[SSIM]{
    \includegraphics[width=1.1 in,height=0.825 in]{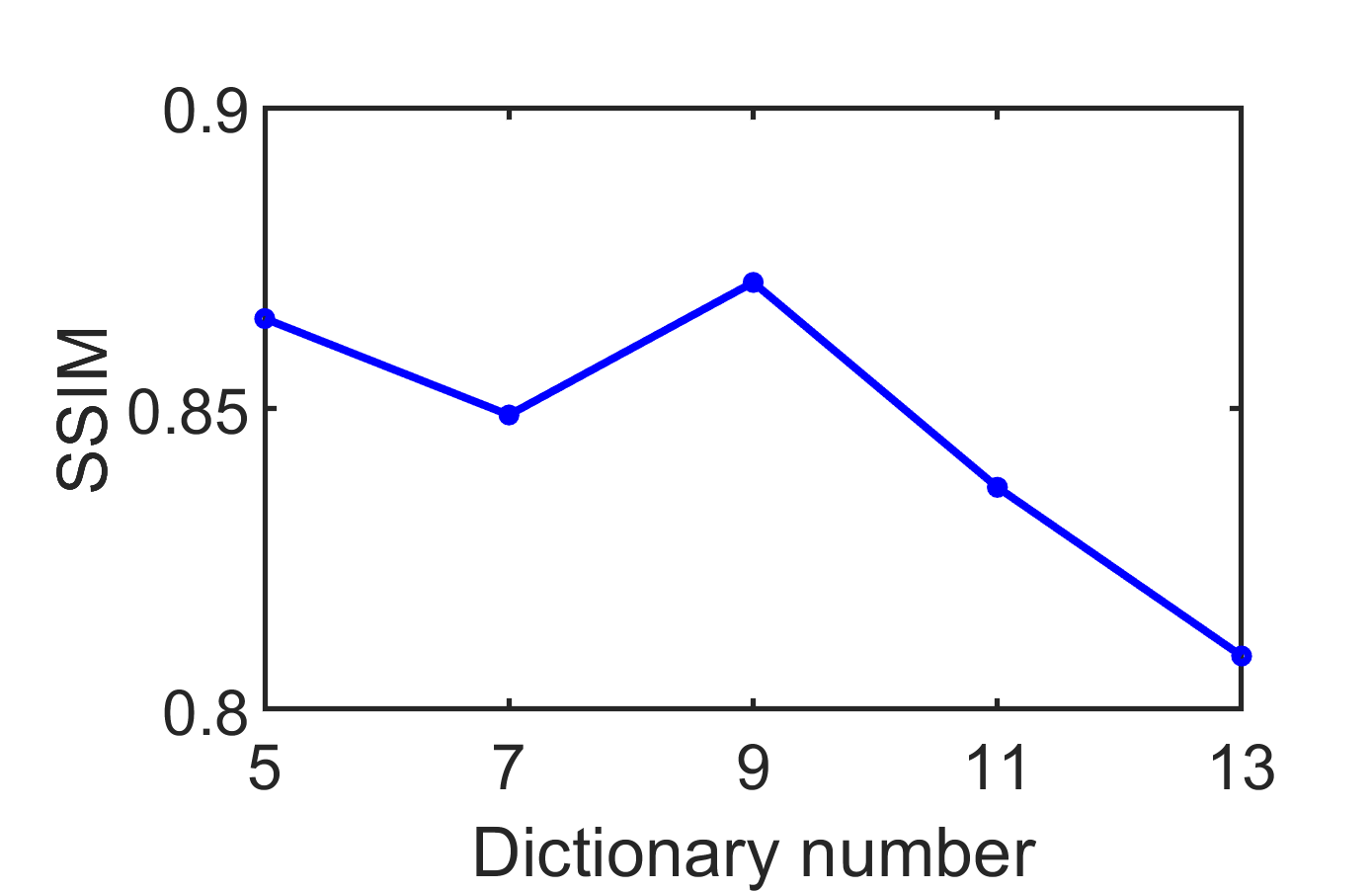}\label{ROC-b22}}
    \subfigure[SAM]{
    \includegraphics[width=1.1 in,height=0.825 in]{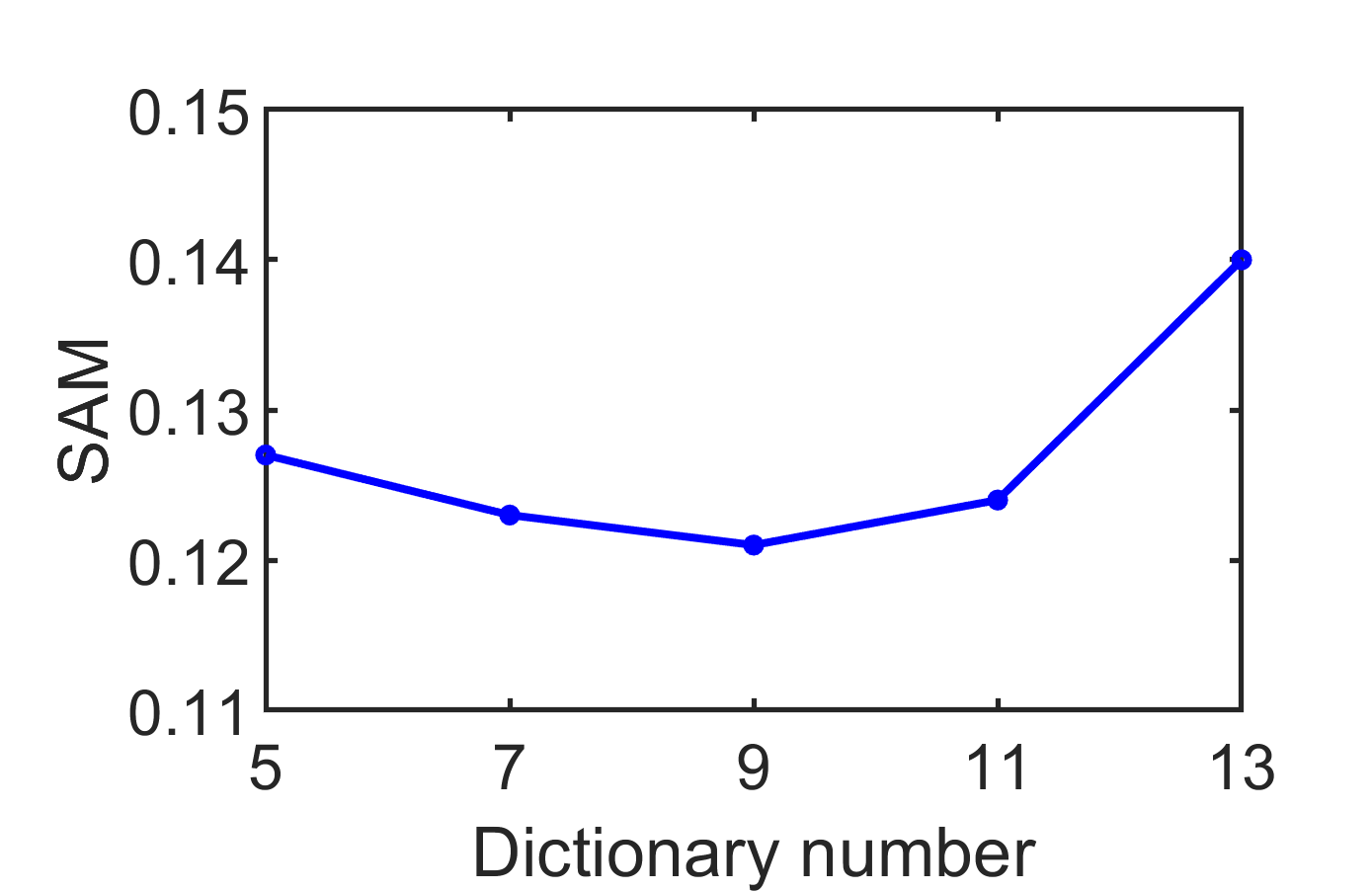}\label{ROC-b22}}
    \subfigure[$\#$Parameters]{
    \includegraphics[width=1.1 in,height=0.825 in]{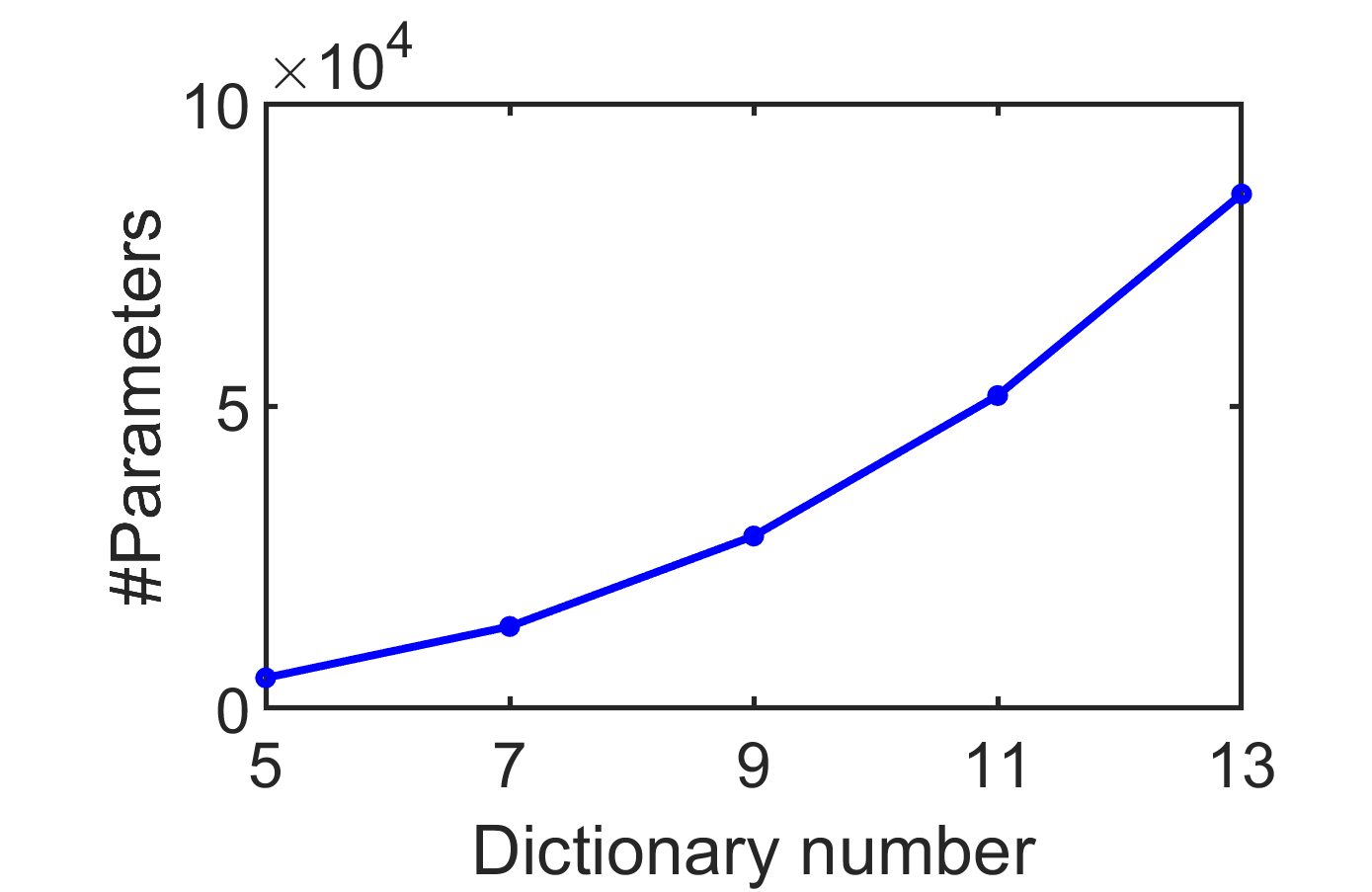}\label{ROC-b22}}
         \vskip-0.2cm
  \caption{Impact of the dictionary number of STAR-Net-S.}\label{dic2}
\end{figure}

\begin{figure}[t]
	\centering
	\includegraphics[width=5.0 in]{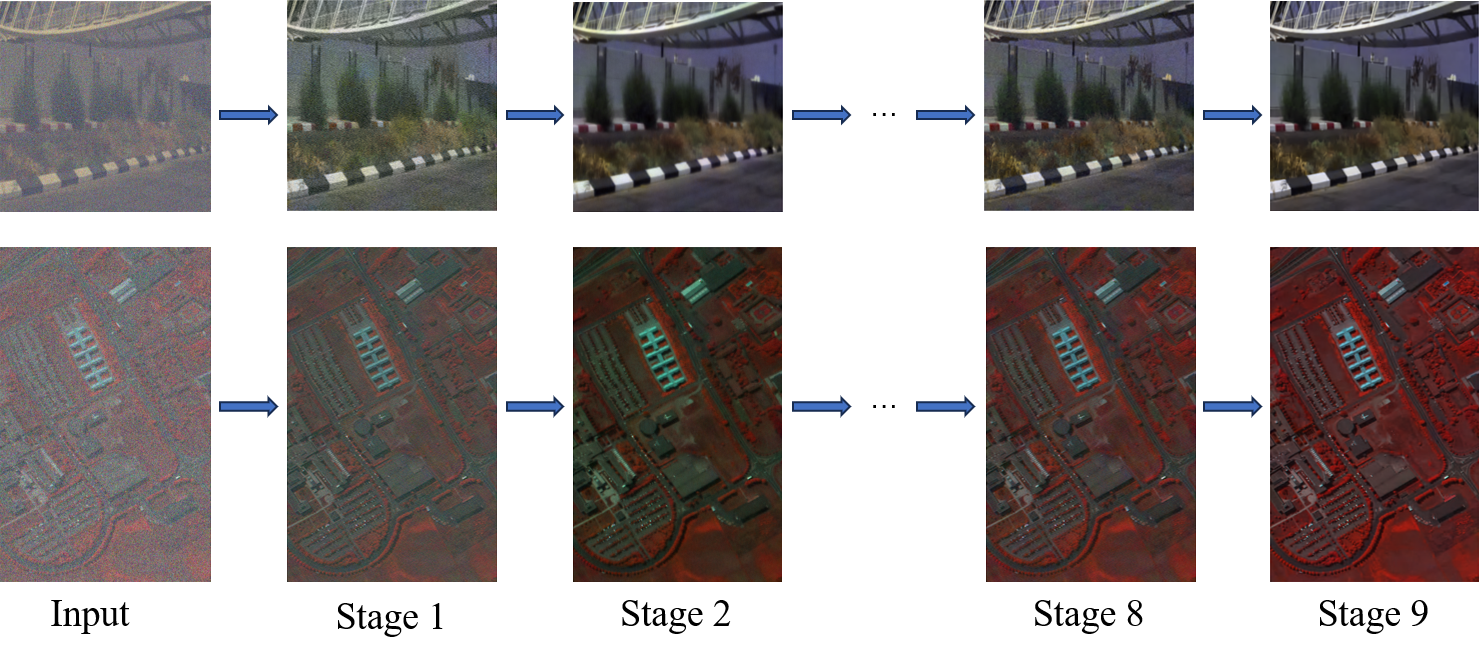}
	\caption{Stage-by-stage visualization process of STAR-Net-S.}
	\label{learningpng}
\end{figure}

\begin{table}[t]
\centering
\small
\renewcommand\arraystretch{1.2}
\caption{Effect of initialization values. The top two values are marked as \textcolor[rgb]{1.00,0.00,0.00}{red} and \textcolor[rgb]{0.00,0.00,1.00}{blue}.}\label{initi1}
\vskip-0.2cm
\setlength{\tabcolsep}{5 pt}
\begin{tabular}{cccccc}
\toprule
Index & 0       & 0.01    & 0.02    & 0.03    & 0.04    \\ \hline
 PSNR $\uparrow$  & 36.730  & 37.455  & \textcolor[rgb]{1.00,0.00,0.00}{37.548}  & \textcolor[rgb]{0.00,0.00,1.00}{37.542}  & 37.346  \\
 SSIM $\uparrow$  & 0.854   & 0.876   & \textcolor[rgb]{1.00,0.00,0.00}{0.879}   & \textcolor[rgb]{0.00,0.00,1.00}{0.878}   & 0.873   \\
 SAM $\downarrow$   & 0.134   & 0.117   & \textcolor[rgb]{1.00,0.00,0.00}{0.115}   & \textcolor[rgb]{0.00,0.00,1.00}{0.116}   & 0.117   \\
 ERGAS $\downarrow$ & 126.123 & 122.737 & \textcolor[rgb]{1.00,0.00,0.00}{120.349} & \textcolor[rgb]{0.00,0.00,1.00}{121.118} & 124.545 \\\bottomrule
\end{tabular}
\end{table}

\begin{table}[t]
\centering
\small
\renewcommand\arraystretch{1.2}
\caption{Friedman test in terms of PSNR. The top two values are marked as \textcolor[rgb]{1.00,0.00,0.00}{red} and \textcolor[rgb]{0.00,0.00,1.00}{blue}.}\label{friedman}
\vskip-0.2cm
\setlength{\tabcolsep}{2 pt}
\begin{tabular}{cccc}
\toprule
~~~Method ~~~    & ~~~Ranking~~~ & ~~~$\varrho$-value~~~                  & ~~~Hypothesis ~~~              \\ \hline
BM4D        & 9.667  & \multirow{12}{*}{0.0001} & \multirow{12}{*}{Reject} \\ 
LLRT        & 10.933   &                          &                          \\ 
LRTDTV       & 8.000   &                          &                          \\ 
NGMeet      & 7.933   &                          &                          \\ 
NLSSR       & 5.333   &                          &                          \\ 
FastHyMix    & 6.800   &                          &                          \\ 
HSI-SDeCNN   & 8.600   &                          &                          \\ 
SMDS-Net    & 5.200   &                          &                          \\ 
Eigen-CNN   & 4.333   &                          &                          \\ 
RCILD      & 7.467   &                          &                          \\ 
STAR-Net   & \textcolor[rgb]{0.00,0.00,1.00}{2.600}   &                          &                          \\ 
STAR-Net-S & \textcolor[rgb]{1.00,0.00,0.00}{1.133}   &                          &                          \\ \bottomrule
\end{tabular}
\end{table}

\begin{figure}[t]
	\centering
	\includegraphics[width=4.0 in]{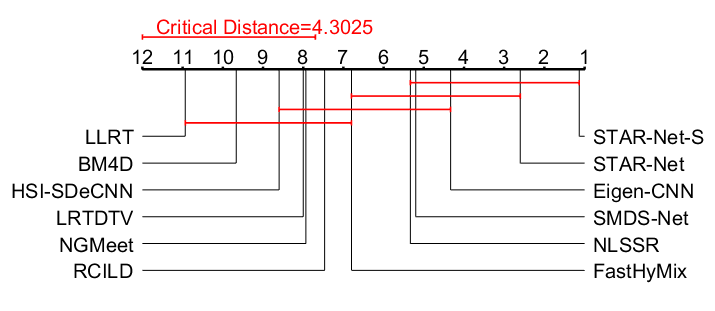}
	\caption{Post-hoc Nemenyi test in temns of PSNR.}
	\label{Nemenyi}
\end{figure}

\subsubsection{Number of Unrolling Iterations} \label{num}
Note that $K$ represents the number of layers in the network and also the number of unrolling iteration. 
Below, we analyze its impact on STAR-Net and STAR-Net-S; see Figure \ref{k1} and Figure \ref{k2}.
As $K$  increases, the number of network parameters also increases. 
For STAR-Net, when $K$ is 9, PSNR and SSIM are the best.
When $K$ is 12, SAM achieves its optimal value. 
Due to the different focus of the four indexes, their variation trends are also different.
Therefore, considering the balance of these four indexes, the unrolling number of STAR-Net is set to 9.
Similarly, Figure \ref{k2} shows that 9 is the optimal option for STAR-Net-S.

\subsubsection{Number of Dictionaries}\label{dic}
For convenience, we set the three dictionaries $\mathrm{D}_{1} ,\mathrm{D}_{2} ,\mathrm{D}_{3}$ to the same size.
The results presented in Figure \ref{dic1} and Figure \ref{dic2} demonstrate that the dictionary size has a direct impact on the number of parameters.
Considering the balance between denoising ability and the number of parameters, setting the number of dictionaries to 9 is the optimal choice for both STAR-Net and STAR-Net-S.

\subsubsection{Analysis of Feature Visualization}\label{learning}
Figure \ref{learningpng} illustrates the intermediate update process of STAR-Net-S on the ICVL and PaviaU datasets, highlighting the stage-by-stage denoising behavior of the deep unrolling network.
In the initial stage, such as Stage1, high-frequency noise is primarily removed, allowing the overall structure of the image to become more distinct. As the iterations progress, structural details and textures are gradually recovered, enhancing image contrast, eliminating residual noise, and ultimately approaching a high-quality clean image.  This progression demonstrates the specific role of each unrolling stage in the denoising process, highlighting the interpretable and observable optimization behavior of the model.

\subsubsection{Analysis of Initialization}\label{Initialization22}

Table \ref{initi1} presents the impact of different initialization values of learnable parameters $\gamma_1$, $\gamma_2$, $l$, $\lambda$, $\mu$, $\beta$ on the model's performance.
For convenience, we set all parameter initial values to be same.
When the initialization value is set to 0, all performance indexes of the model degrade significantly, indicating that the learning capability of the model is greatly impaired, leading to a decline in overall denoising effectiveness.
For other initialization values, the performance differences are relatively minor. When the initialization value is set to 0.02, the model achieves the best performance across all four indexes. Therefore, we initialize the learnable parameters $\gamma_1$, $\gamma_2$, $l$, $\lambda$, $\mu$, $\beta$ to 0.02 in this paper.

\subsubsection{Statistical Tests}

The Friedman test analyzes the rankings of measurements across multiple conditions to determine if there are significant differences in their average performance.
We ranked all 12 methods from best to worst based on PSNR for each of the four noise scenarios across two synthetic datasets, with rankings assigned from 1 to 12.
The performance rankings of the different approaches are shown in Table \ref{friedman}.
In this experiment, the null hypothesis $\mathcal{H}_0$ of the Friedman test assumes that there are no significant differences in performance across all the compared methods.
With the significance level of $\alpha$ = 0.05, Table \ref{friedman} indicates that $\varrho$ = 0.0001, which results in the rejection of the null hypothesis $\mathcal{H}_0$. 
This demonstrates that there are significant differences among the various comparison methods.

However, the Friedman test is unable to indicate which methods differ from each other. 
Therefore, a subsequent Nemenyi test is required to further analyze and identify the significant differences between the methods.
The Nemenyi test uses the critical difference (CD) value to evaluate whether the ranking difference between two methods is statistically significant.
Figure \ref{Nemenyi} presents the results of the post-hoc Nemenyi test, illustrating the statistical significance of ranking differences between methods.
Figure \ref{Nemenyi} shows that the performance differences among STAR-Net, STAR-Net-S, SMDS-Net, Eigen-CNN, and NLSSR are minimal. 
However, STAR-Net and STAR-Net-S exhibit significant differences when compared to the other seven methods.
When considering both Table \ref{friedman} and Figure \ref{Nemenyi}, it is evident that STAR-Net and STAR-Net-S demonstrate the strongest performance among all methods, further confirming the effectiveness of the proposed methods.

\begin{table}[t]
\centering
\renewcommand\arraystretch{1.2}
\caption{Testing runtime (seconds) of all methods. The top two values are marked as \textcolor[rgb]{1.00,0.00,0.00}{red} and \textcolor[rgb]{0.00,0.00,1.00}{blue}.}\label{time}
\vskip-0.2cm
\setlength{\tabcolsep}{2.5 pt}
\resizebox{\textwidth}{!}{
\begin{tabular}{ccccccccccccc}
\toprule
Dataset    & \makecell[c]{BM4D\\}   & \makecell[c]{LLRT\\}                       & \makecell[c]{LRTDTV\\} & \makecell[c]{NGMeet\\}& \makecell[c]{NLSSR\\}&\makecell[c]{FastHy\\Mix} & \makecell[c]{HSI-SDe\\CNN}  & \makecell[c]{SMDS-\\Net}    &\makecell[c]{Eigen-\\CNN}   & \makecell[c]{RCILD\\}         & \makecell[c]{STAR-\\Net}        & \makecell[c]{STAR-\\Net-S}    \\ \hline
ICVL & 561.307  &888.305  &136.796  &335.265  &206.964  & \textcolor[rgb]{0.00,0.00,1.00}{8.555}  & 11.093      & 105.109 & \textcolor[rgb]{1.00,0.00,0.00}{6.478  } &  24.743  & 107.552   & 107.958     \\
PaviaU & 1123.815 &2581.424 &354.911 &716.722 &491.568 &82.029 & 60.522     & 349.292  & \textcolor[rgb]{1.00,0.00,0.00}{11.361 } &  \textcolor[rgb]{0.00,0.00,1.00}{30.706} & 337.133    & 341.220     \\ \bottomrule
\end{tabular}}
\end{table}

\begin{figure}[t]
	\centering
	\includegraphics[width=2.5 in]{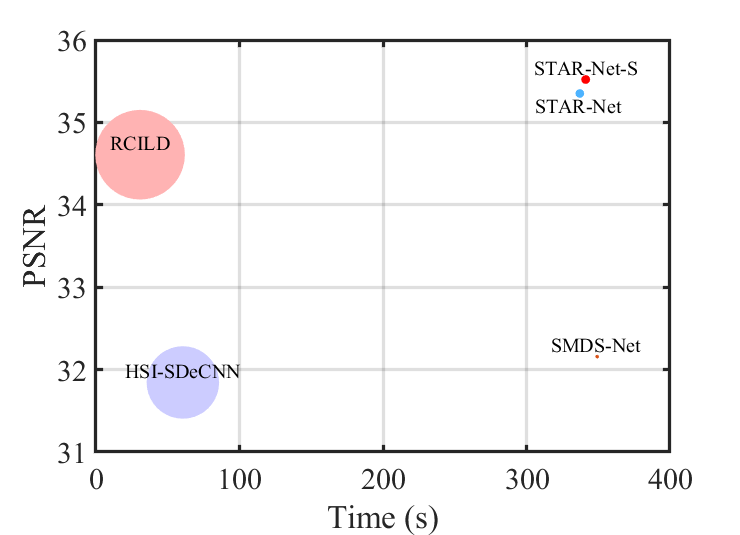}
	\caption{Performance and runtime comparison on the PaviaU dataset. The size of each circle represents the number of model parameters.}
	\label{parapng}
\end{figure}

\subsubsection{Runtime}

This section discusses the computation time of all methods on the two datasets, with the results summarized in Table \ref{time}.
As shown in Table \ref{time}, the model-based methods exhibit relatively slow runtime, particularly LLRT.
Since STAR-Net and STAR-Net-S unroll the whole model into network components, their runtime is slower compared to deep learning-based methods like HSI-SDeCNN, Eigen-CNN, and RCILD, but still improves compared to SMDS-Net.

Meanwhile, Figure \ref{parapng} illustrates the performance and runtime comparison of several deep learning methods on the PaviaU dataset, with circle sizes representing the number of parameters. Considering the three metrics of PSNR, runtime, and number of parameters, STAR-Net and STAR-Net-S demonstrate the most balanced and superior overall performance.

\begin{figure}[t]
  \centering
    \subfigure[STAR-Net]{
    \includegraphics[width=1.8 in,height=1.4 in]{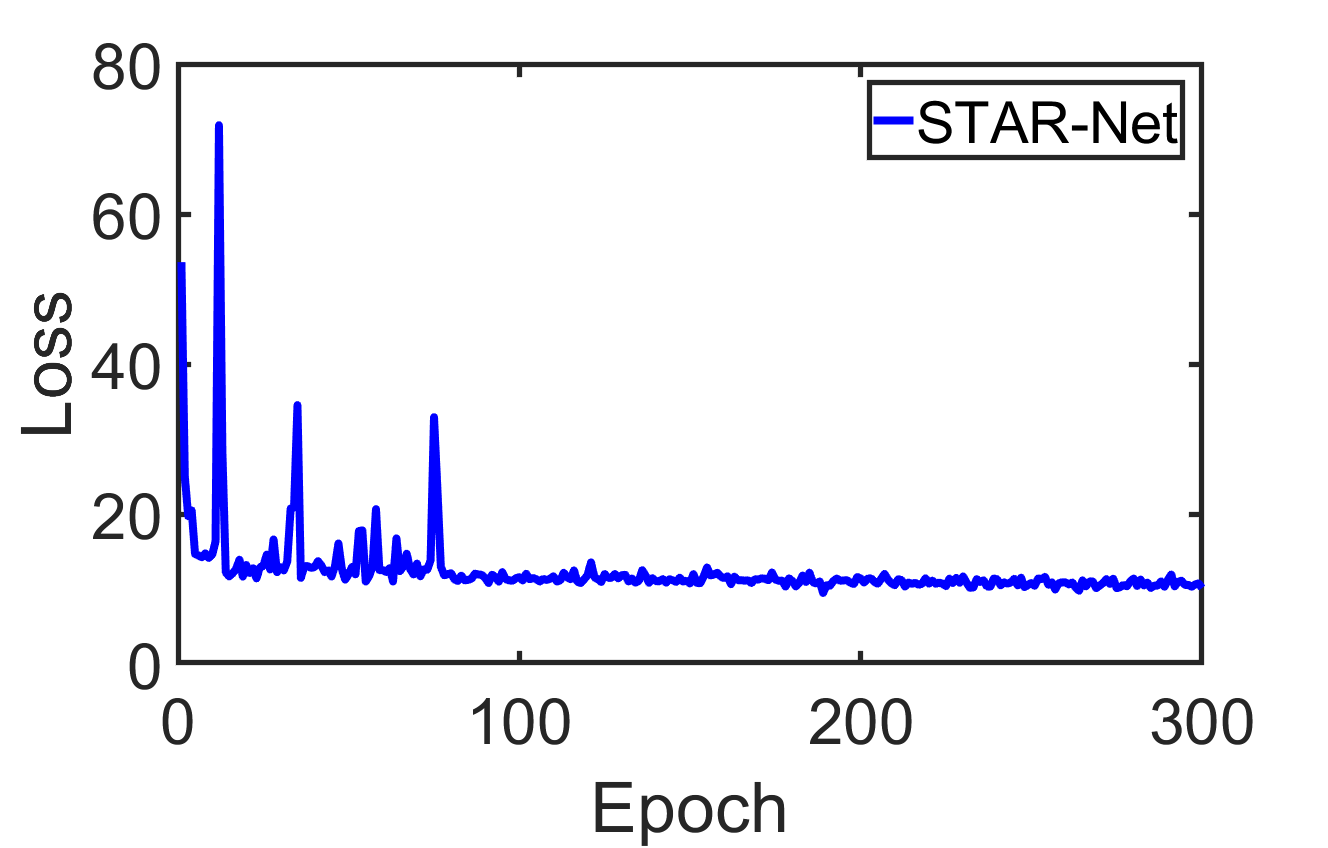}\label{ROC-b21}}
    \subfigure[STAR-Net-S]{
    \includegraphics[width=1.8 in,height=1.4 in]{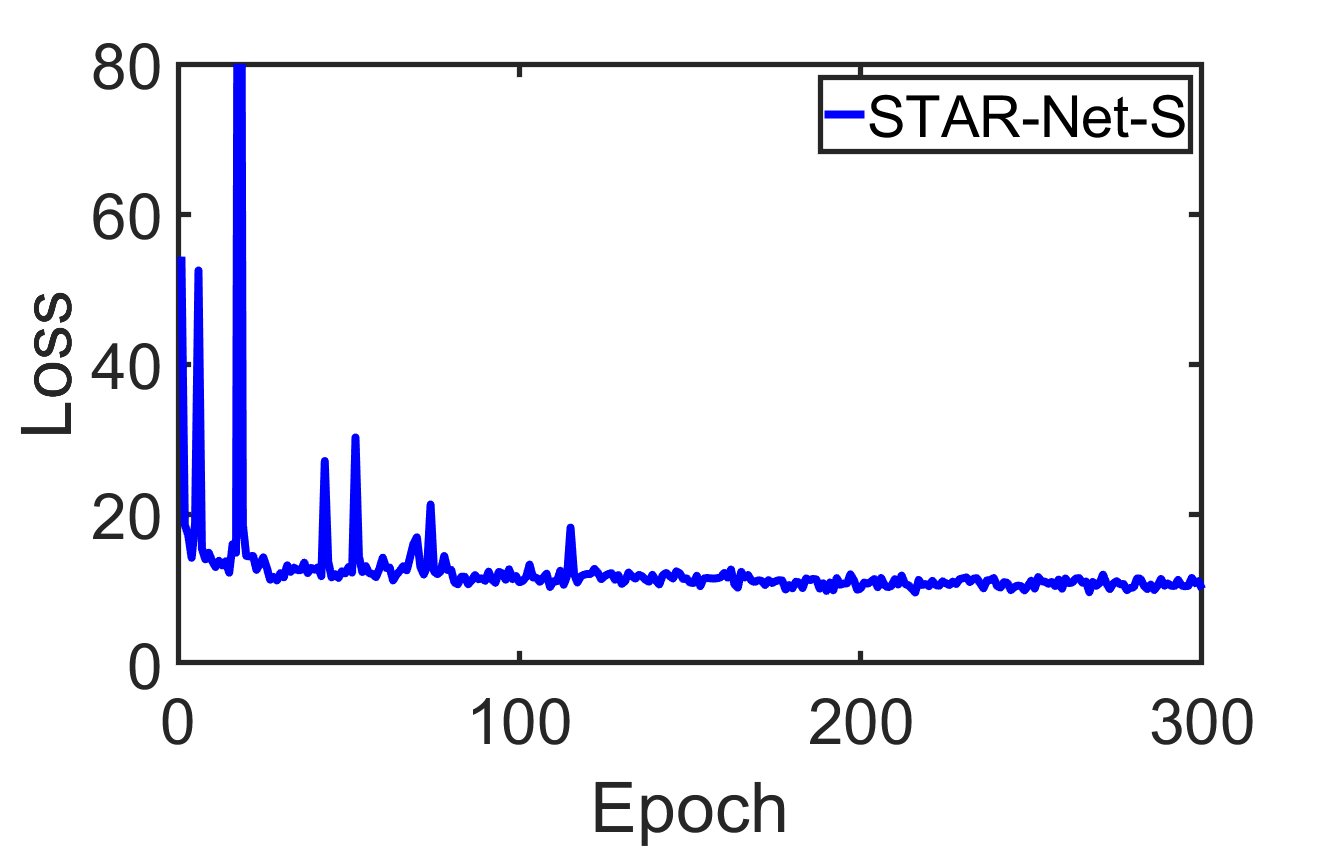}\label{ROC-b22}}
         \vskip-0.2cm
  \caption{Convergence analysis of the loss function.}\label{loss1}
\end{figure}


\subsubsection{Convergence Analysis}
Regarding the convergence, Figure \ref{loss1} depicts the loss curves of STAR-Net and STAR-Net-S after 300 epochs of training. 
It can be clearly seen that the loss curves of STAR-Net and STAR-Net-S during training tend to be stable after 100 epochs, which also suggests the convergence of STAR-Net and STAR-Net-S.

\section{Conclusion}\label{sec5}

In this paper, we propose two novel RSI denoising methods, i.e., STAR-Net and STAR-Net-S.
The core is to introduce a low-rank prior to preserve non-local self-similarity and a sparse prior to improve robustness for non-Gaussian noise.
Subsequently, the classical ADMM framework is integrated with deep unrolling networks, transforming the iterative optimization process into a trainable network. 
This design enables the model to learn parameters in an end-to-end manner, thereby eliminating the need for tedious manual parameter tuning that is typically required in conventional model-based approaches. 
Therefore, our proposed STAR-Net and STAR-Net-S inherit the advantages of model-based and deep learning-based approaches, and have strong interpretability and learnability. 
Comprehensive experiments on both synthetic and real-world remote sensing datasets demonstrate the effectiveness of the proposed STAR-Net and STAR-Net-S methods.
The PSNR values for STAR-Net and STAR-Net-S increased by 2.16\% and 2.85\% on the ICVL dataset.
In addition, the number of parameters, unrolling iterations, dictionaries, feature visualization, initialization setting of the network are discussed in detail.

Although STAR-Net and STAR-Net-S have demonstrated encouraging denoising performance, their testing runtime are still not optimal.
In the future, we are interested in exploring the development of faster optimization algorithms or more efficient physics-informed networks, with the goal of reducing computational complexity and runtime while maintaining or even enhancing denoising effectiveness.

%

\section*{Acknowledgement}
This work was supported in part by the National Natural Science Foundation of China under Grant 62204044 and 12371306, and in part by the State Key Laboratory of Integrated Chips and Systems under Grant SKLICS-K202302.




%
%
%

\bibliographystyle{elsarticle-num-names}  

\biboptions{sort&compress}
\bibliography{mybibfile1}
\end{document}